\documentclass[prx,aps,reprint,twocolumn,superscriptaddress,nofootinbib,amsmath,amssymb]{revtex4-1}

\usepackage{graphicx}
\usepackage{dcolumn}
\usepackage{bm}
\usepackage{hyperref}
\hypersetup{colorlinks,linkcolor=blue, citecolor=red}

\usepackage{amsthm} 
\makeatletter
\newtheorem*{rep@theorem}{\rep@title}
\newcommand{\newreptheorem}[2]{%
\newenvironment{rep#1}[1]{%
 \def\rep@title{#2 \ref{##1}}%
 \begin{rep@theorem}}%
 {\end{rep@theorem}}}
\makeatother
\newtheorem{theorem}{Theorem} 
\newreptheorem{theorem}{Theorem} 
\newreptheorem{lemma}{Lemma} 
\newreptheorem{property}{Property} 

\usepackage{float}
\usepackage{subfigure}
\usepackage{algorithmic}

\usepackage{xcolor}

\def\Cmag{\overline{C}}
\newcommand{\Cbar}{\text{Cbar}}
\newcommand{\R}{\mathbb{R}}

\allowdisplaybreaks

\begin{document}

\preprint{APS/123-QED}

\title{Marvels and pitfalls of the Langevin algorithm \\ in noisy high-dimensional inference}

\author{Stefano Sarao Mannelli}
\affiliation{Universit\'e Paris-Saclay, CNRS, CEA, Institut de physique th\'eorique, 91191, Gif-sur-Yvette, France.}
\author{Giulio Biroli}
\affiliation{Laboratoire de Physique de l'Ecole normale sup\'erieure ENS,  Universit\'e PSL, CNRS, Sorbonne Universit\'e, Universit\'e Paris-Diderot, Sorbonne Paris Cit\'e Paris, France}
\author{Chiara Cammarota}
\affiliation{Department of Mathematics, King's College London, Strand London WC2R 2LS, UK}
\author{Florent Krzakala}
\affiliation{Laboratoire de Physique Statistique, CNRS \& Universit\'e Pierre \& Marie Curie \& Ecole Normale Sup\'erieure \& PSL Universit\'e, 75005 Paris, France}
\author{Pierfrancesco Urbani}
\affiliation{Universit\'e Paris-Saclay, CNRS, CEA, Institut de physique th\'eorique, 91191, Gif-sur-Yvette, France.}
\author{Lenka Zdeborov\'a}
\affiliation{Universit\'e Paris-Saclay, CNRS, CEA, Institut de physique th\'eorique, 91191, Gif-sur-Yvette, France.}

\date{\today}

\begin{abstract}
        Gradient-descent-based algorithms and their stochastic versions
	have widespread applications in machine learning and statistical inference.
	In this work we carry out an analytic study of the performance
        of the one most commonly considered in physics, the Langevin algorithm,
	in the context of noisy high-dimensional inference.
	We employ the Langevin algorithm to sample the posterior
        probability measure for the spiked mixed matrix-tensor model. 
	The typical behaviour of this algorithm is described by a system of integro-differential equations that we call the Langevin state evolution, whose solution is compared with the one of the state evolution of approximate message passing (AMP).
	Our results show that, remarkably, the algorithmic threshold of the Langevin algorithm is sub-optimal with respect to
	the one given by AMP.
	This phenomenon is due to the residual glassiness
	present in that region of parameters. 
	We present also a simple heuristic expression of the transition line which appears to be in agreement with the numerical results.
\end{abstract}

\maketitle

\section{Motivation}

Algorithms based on noisy variants of gradients descent
\cite{bottou2010large,welling2011bayesian} stand at the roots of many
modern applications of data science, and are being used in a wide
range of high-dimensional non-convex optimization problems. The
widespread use of stochastic gradient descent in deep learning
\cite{lecun2015deep} is certainly one of the most prominent
examples. For such algorithms, the existing theoretical analysis
mostly concentrate on convex functions, convex relaxations or on
regimes where spurious local minima become irrelevant. For problems
with complicated landscapes where, instead, useful convex relaxations
are not known and spurious local minima cannot be ruled out, the
theoretical understanding of the behaviour of gradient-descent-based
algorithm remains poor and represents a major avenue of research.

The goal of this paper is to contribute to such an understanding in
the context of statistical learning, and to transfer
ideas and techniques developed for glassy
dynamics \cite{bouchaud1998out} to
the analysis of non-convex high-dimensional inference.
In statistical learning, the minimization of
a cost function is not the goal per se, but rather a way to uncover
an unknown structure in the data. One common way to model and analyze
this situation is to generate data with a hidden structure, and to see
if the structure can be recovered. This is easily set up as a {\it
		teacher-student} scenario
\cite{ReviewTishby,REVIEWFLOANDLENKA}: {\it
First} a teacher generates latent variables and uses them as input
of a prescribed model to generate a synthetic dataset.  {\it Then},
the student observes the dataset and tries to infer the values of the
latent variables. The analysis of this setting has been carried out
rigorously in a wide range of teacher-student models for
high-dimensional inference and learning tasks as diverse as planted
clique \cite{deshpande2015finding}, generalized linear models such as
compressed sensing or phase retrieval \cite{barbier2017phase},
factorization of matrices and tensors
\cite{XXT,MiolaneTensor}
or simple models of neural networks
\cite{aubin2018committee}. In these works, the
information theoretically optimal performances ---the one obtained by
an ideal Bayes-optimal estimator, not limited in time
and memory--- have been computed.

The main question is, of course, how {\it practical algorithms}
---operating in polynomial time with respect to the problem size---
compare to these ideal performances. The last decade brought
remarkable progress into our understanding of the performances
achievable computationally. 
In particular, many algorithms based on 
message passing \cite{DMM09,REVIEWFLOANDLENKA}, spectral methods
\cite{2013PNAS..11020935K}, and semidefinite programs (SDP)
\cite{hopkins2017bayesian} were analyzed. Depending on the
signal-to-noise ratio, these algorithms were shown to be very efficient in many of those task. Interestingly,
all these algorithm fail to reach good performance in the same region of the parameter
space, and this striking observation has led to the identification of
a well-defined {\it hard phase}. This is a regime of parameters in which the
underlying statistical problem can be information-theoretically
solved, but no efficient algorithms are
known, rendering the problem essentially unsolvable for large instances.  This stream of
ideas is currently gaining momentum and impacting research in
statistics, probability, and computer science.

The performance of the noisy-gradient descent algorithms 
remains an
entirely open question. Do they allow to reach the same performances
as message passing and SDPs? Can they enter
the hard phase, do they stop to be efficient at the same moment as
the other approaches, or are they worse? The ambition of the present paper is to
address these questions by analyzing the performance of the Langevin
algorithm in the high-dimensional limit of a particular spiked mixed matrix-tensor
model, defined in detail in the next section.

Similar models have played a
fundamental role in statistics and random matrix theory
\cite{baik2005phase,johnstone2009consistency}. Tensor
factorization is also an important topic in machine
learning and is widely used in data analysis \cite{anandkumar2014tensor,richard2014statistical,hopkins2015tensor,ge2017optimization,arous2018algorithmic,ros2018complex}. At variance
with the pure spiked tensor case \cite{richard2014statistical},
this mixed matrix-tensor model has the advantage that the algorithmic threshold
appears at the same scale as the
information-theoretic one, similarly to what is observed in simple
models of neural networks
\cite{barbier2017phase,aubin2018committee}. We view the spiked mixed
matrix-tensor model as a prototype for non-convex high-dimensional
landscape. The key virtue of the model is its tractability. 

We focus on the {\it Langevin
algorithm} for two main reasons: Firstly it is the gradient-based
algorithm that is most widely studied in physics. Secondly, at large time
(possibly growing exponentially with the system size) it is known to
sample the associated Boltzmann measure thus evaluating the
Bayes-optimal estimator for the inference problem. We evaluate 
performance of the algorithm at times that are large but not growing with the system
size. We explicitly
compare thus obtained performance to the one of the Bayes optimal estimator
and to the best known efficient algorithm so-far -- the approximate message passing algorithm \cite{DMM09, REVIEWFLOANDLENKA}. 
In particular, contrary to what
has been anticipated in
\cite{krzakala2009hiding,decelle2011asymptotic}, but as surmised in  \cite{antenucci2018glassy}, we observe that the
performance of the Langevin algorithm is hampered by the
many spurious metastable states still present in the AMP-easy
phase. 
In showing that, we shed light on a number of
  properties of the Langevin algorithm that may seem counterintuitive
  at a first sight (e.g. the performance getting worse as the noise
  decreases).

The possibility to describe analytically the behavior of the Langevin
algorithm in this model is enabled by the existence of the
Crisanti-Horner-Sommers-Cugliandolo-Kurchan (CHSCK) equations in spin
glass theory, describing the behavior of the Langevin dynamics in the
so-called spherical $p$-spin model \cite{CHS93, cugliandolo1993analytical},
where the method can be rigorously
justified~\cite{arous2006cugliandolo}. These equations were a key
development in the field of statistical physics of disordered systems
that lead to detailed understanding and predictions about the slow
dynamics of glasses \cite{bouchaud1998out}. In this paper, we bring
these powerful methods and ideas into the realm of statistical
learning.

\section{The spiked matrix-tensor model}
We now detail the spiked mixed matrix-tensor problem: a {\it teacher}
generates a $N$-dimensional vector $x^*$ by choosing each of its
components independently from a normal Gaussian distribution of zero
mean and unit variance.  In the large $N$ limit this is equivalent to
have a flat distribution over the $N$-dimensional hypersphere
${\cal S}_{N-1}$ defined by $|x^*|^2=N$. In the paper we will use
either of these two, as convenient.  The teacher then generates
a symmetric matrix $Y_{ij}$ and a symmetric order-$p$ tensor
$T_{i_1,\dots,i_p}$ as
\begin{equation}
	\begin{split}
		Y_{ij} &= \frac{1}{\sqrt{N}} x^*_i x^*_j + \xi_{ij}  \ \ \ \
		\forall i<j\, ,\\
		T_{i_1 \dots i_p} &=   \frac{\sqrt{(p-1)!}}{N^{(p-1)/2}}  x^*_{i_1}
		\dots x^*_{i_p} + \xi_{i_1 \dots i_p} \ \ \ \forall i_1<\ldots <i_p\, ,
		\label{theModel}
	\end{split}
\end{equation}
where $\xi_{ij}$ and $\xi_{i_1,\dots,i_p}$  are iid Gaussian components of a
symmetric random matrix and tensor of
zero mean and variance $\Delta_2$ and $\Delta_p$, respectively;
$\xi_{ij}$ and $\xi_{i_1,\dots,i_p}$ correspond
to noises corrupting the signal of the teacher.
In the limit $\Delta_2 \to 0$, and
$\Delta_p\to 0$, the above model reduces to the canonical spiked
Wigner model \cite{deshpande2014information}, and spiked tensor model \cite{richard2014statistical}, respectively.
The goal of the student is to infer the vector $x^*$ from the
knowledge of the matrix $Y$, of the tensor $T$, of the values
$\Delta_2$ and $\Delta_p$, and the knowledge of the spherical prior.  The
scaling with $N$ as specified in Eq.~(\ref{theModel}) is chosen in
such a way that the information-theoretically best achievable error varies between perfectly reconstructed spike $x^*$ and random guess
from the flat measure on ${\cal S}_{N-1}$.  Here, and in the rest of the
paper we denote $x \in {\cal S}_{N-1}$ the $N$-dimensional vector, and
$x_i$ with $i=1,\dots,N$ its components.

This model belongs to the generic direction of study of
Gaussian functions on the $N$-dimensional sphere, known as $p$-spin
spherical spin glass models in the physics literature, and as isotropic models in the Gaussian process literature~\cite{gross1984simplest,fyodorov2004complexity,auffinger2013random,sagun2014explorations,arous2017landscape}. In statistics and machine learning, these models have appeared following the studies of spiked matrix and tensor models \cite{johnstone2009consistency,deshpande2014information,richard2014statistical}.
Analogous mixed matrix-tensor models, where next to a order-$p$ tensor one
observes a matrix created from the same spike are studied e.g. in
\cite{anandkumar2014tensor} in the context of topic modeling, or in \cite{richard2014statistical}.
From the optimization-theory point of view, this model is highly
non-trivial being high-dimensional and non-convex.
For the purpose of the present paper
this model is chosen with the hypothesis that its energy landscape
presents properties that will generalize to other non-convex
high-dimensional problems. The following three ingredients are key to
the analysis: (a) It
is in the class of models for which the Langevin algorithm can be
analyzed exactly in large $N$ limit.
(b) The different phase transitions, both
algorithmic and information theoretic, discussed hereafter,
all happen at $\Delta_2={\cal O}(1)$, $\Delta_p={\cal O}(1)$. This
means that when the problem becomes algorithmically tractable it is
still in the noisy regime, where the optimal mean squared error is bounded away from zero. 
(c) The AMP algorithm is in this model conjectured to be optimal among polynomial
algorithms.  It is
this second and third  ingredient that are not present in the pure spiked tensor
model \cite{richard2014statistical}, making it unsuitable for our
present study. 
We note that the Langevin algorithm was recently
analyzed for the pure spiked tensor model in
\cite{arous2018algorithmic} in a regime where the noise variance is
very small $\Delta \sim N^{-p/2}$, but we also note that in that model algorithms
such as tensor unfolding, semidefinite programming, homotopy methods,
or improved message passing schemes work better,
roughly up to $\Delta \sim N^{-p/4}$
\cite{richard2014statistical,anandkumar2014tensor,hopkins2015tensor,anandkumar2016homotopy,wein2019kikuchi}.



\section{Bayes-optimal estimation and message-passing}
In this section we present the performance of the Bayes-optimal
estimator and of the approximate message passing
algorithm.
This theory is based on a straightforward
adaptation of analogous results known for the pure spiked matrix model
\cite{deshpande2014information,LKZ17,XXT} and for the pure spiked
tensor model \cite{richard2014statistical,MiolaneTensor}.

The Bayes-optimal estimator $\hat x$ is defined as the one that among all estimators
minimizes the mean-squared error (MSE) with the spike $x^*$.  Starting
from the posterior probability distribution
\begin{equation}
	\begin{split}
		P(x|Y,T) & = \frac{1}{Z(Y,T)} \left[\prod_{i=1}^N e^{-x_i^2/2} \right] \prod_{i<j}e^{-\frac1{2\Delta_2}\left(Y_{ij}-\frac{x_ix_j}{\sqrt{N}}\right)^2}\\
		& \prod_{i_1<\dots<i_p} e^{-\frac1{2\Delta_p}\left(T_{i_1\dots i_p}-\frac{\sqrt{(p-1)!}}{N^{(p-1)/2}}x_{i_1}\dots x_{i_p}\right)^2} \, ,
	\end{split}
	\label{posterior}
\end{equation}
the Bayes-optimal estimator reads
\begin{equation}
	\hat x_i = {\mathbb E}_{P(x|Y,T)}(x_i) \, .  \label{BO_estimator}
\end{equation}
To simplify notation, and to make contact with the energy landscape
and the statistical physics notations, it is convenient to introduce
the energy cost function, or Hamiltonian, as
\begin{equation}
	\begin{split}
		\mathcal{H}(x) & = \mathcal{H}_2 + \mathcal{H}_p  =  - \frac1{\Delta_2 \sqrt{N}} \sum_{i<
			j} Y_{ij}  x_ix_j   \\ & -  \frac{\sqrt{(p-1)!}}{\Delta_p
			N^{(p-1)/2}} \!\!\! \sum_{ i_1<\dots<i_p} \! \! \! T_{i_1\dots
				i_p}x_{i_1}\dots  x_{i_p}
	\end{split}
	\label{Hamiltonian}
\end{equation}
so that keeping in mind that for $N\to \infty$ the spherical
constraint is satisfied $|x|^2=N$, the posterior is written as
$P(x|Y,T) = \exp[ -\mathcal{H}(x) ] /\tilde Z(Y,T)$, where
$\tilde Z$ is the normalizing partition function.

With the use of the replica theory and its recent proofs from
\cite{XXT,MiolaneXX,MiolaneTensor} one can establish rigorously that
the mean squared error achieved by the Bayes-optimal estimator (\ref{posterior}) is
given as ${\rm MMSE} = 1- m^*$ where $m^* \in {\mathbb R}$ is the
global maximizer of the so-called free entropy of the problem
\begin{eqnarray}
	\Phi_{\rm RS}(m) = \frac12\log(1-m) + \frac{m}2 + \frac{m^2}{4\Delta_2} + \frac{m^p}{2p\Delta_p} \, .  \label{RS_free_entropy}
\end{eqnarray}
This expression is derived, and proven, in the Appendix
Sec.~\ref{SI:AMP_proof_RS}. We note that the proof
  applies to the posterior distribution  (\ref{posterior}) with the
  Gaussian prior.

We now turn to the approximate message-passing (AMP) \cite{richard2014statistical,
	MiolaneTensor}, that is the best algorithm known so far for this problem.
AMP is an iterative algorithm inspired from the work of
Thouless-Anderson and Palmer in statistical physics
\cite{TAP77}. We explicit its form in the Appendix Sec.~\ref{SI:AMP}. Most
remarkably performance of AMP can be evaluated by
tracking its evolution with the iteration time and it is given in
terms of the (possibly local) maximum of the above free entropy that
is reached as a fixed point of the following iterative process
\begin{eqnarray}
	m^{t+1} = 1- \frac1{1+m^t/\Delta_2+(m^t)^{p-1}/\Delta_p}  \label{eq_SE}
\end{eqnarray}
with initial condition $m^{t=0}=\epsilon$ with $0 <\epsilon \ll
	1$. Eq.~(\ref{eq_SE}) is called the {\it State Evolution} of AMP and
its validity is proven for closely related models in
\cite{javanmard2013state}. We
denote the corresponding fixed point $m_{\rm AMP}$ and the
corresponding estimation error ${\rm MSE}_{\rm AMP}= 1-m_{\rm AMP}$.

\begin{figure}
	\centering
	\includegraphics[scale=.48]{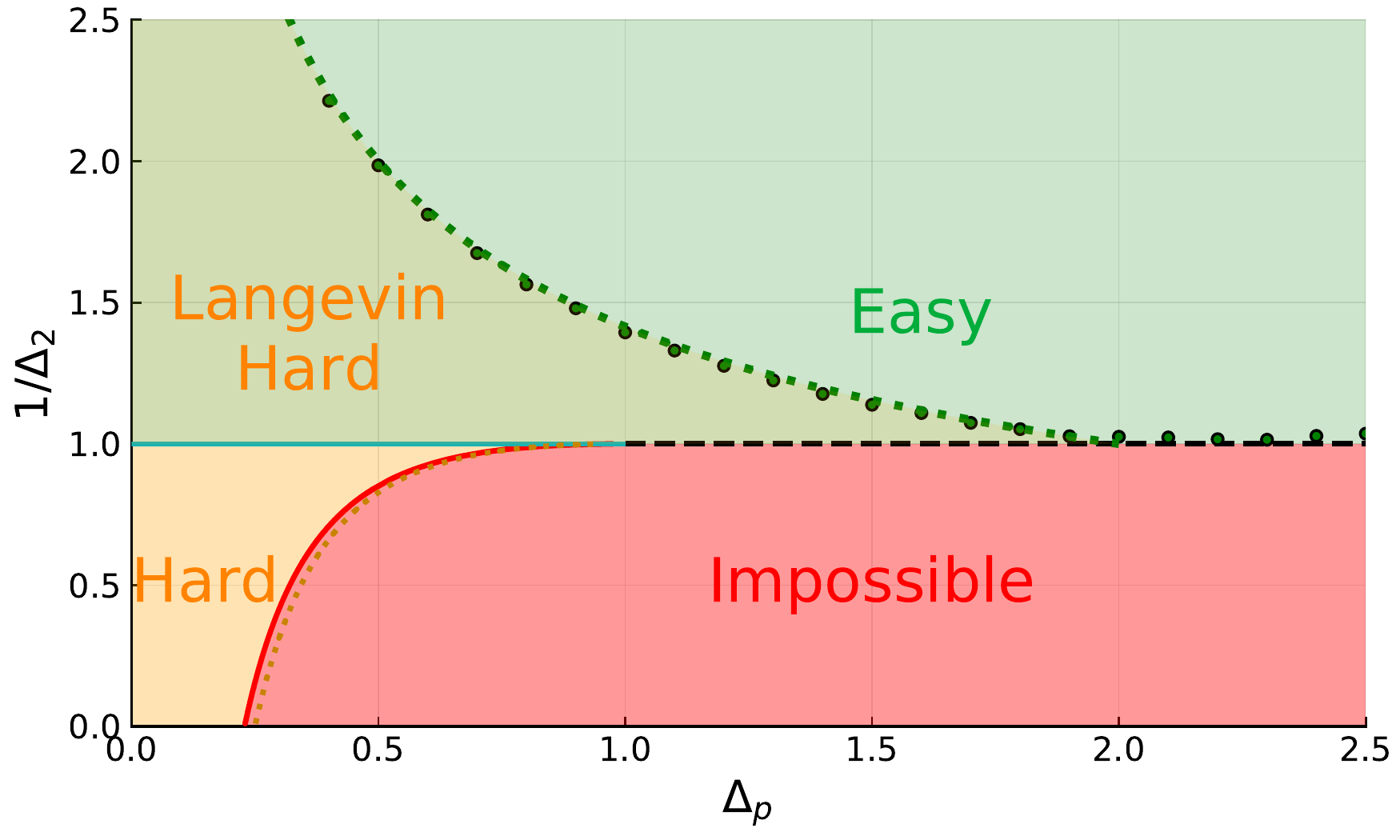}
	\caption{Phase
		diagram of the spiked $2+3$-spin model (matrix plus order 3
		tensor are observed). In the easy (green)
		region AMP achieves the optimal error
		smaller than random pick from the prior. In the impossible
		region (red) the optimal error is as bad as random pick
		from the prior, and AMP achieves it as well. In the hard
		region (orange) the optimal error is low, but AMP does not
		find an estimator better than random pick
		from the prior.
		In the case of Langevin algorithm the performance is strictly worse than that
		for AMP in the sense that
		the hard region increases up to line $1/\Delta_2^* =\max(1 , \sqrt{\Delta_3/2})$, depicted in green
              dots. The green circles are obtained by numerical
              extrapolation of the Langevin state evolution
              equations. 
		%
		%
		\label{fig:phase_diagram}}
\end{figure}

The phase diagram presented in Fig.~\ref{fig:phase_diagram} summarizes
this theory for the spiked $2+3$-spin model. It is deduced by
investigating the local maxima of the scalar function
(\ref{RS_free_entropy}). Notably we observe that the phase diagram in
terms of $\Delta_2$ and $\Delta_p$ splits into three phases
\begin{itemize}
	\item{{\bf Easy} in green for $\Delta_2<1$ and any $\Delta_p$: The fixed point of the state evolution
	      (\ref{eq_SE}) is the global maximizer of the free entropy
	      (\ref{RS_free_entropy}), and $ m^*= m_{\rm AMP} >0$.}
	\item{{\bf Hard} in orange for $\Delta_2>1$ and low $\Delta_p < \Delta_p^{\rm IT}(\Delta_2)$: The fixed point of the state evolution
	      (\ref{eq_SE}) is not the global maximizer of the free entropy
	      (\ref{RS_free_entropy}), and $m^* > m_{\rm AMP}  = 0$.}
	\item{{\bf Impossible} in red for $\Delta_2>1$ and high
	      $\Delta_p>\Delta_p^{\rm IT}(\Delta_2)$: The fixed point of the state evolution
	      (\ref{eq_SE}) is the global maximizer of the free entropy
	      (\ref{RS_free_entropy}), and $ m^* = m_{\rm AMP} = 0$.}
\end{itemize}

For the $2+p$-spin model with $p > 3$ the phase diagram is slightly
richer and is presented in the Appendix Sec.~\ref{SI:numerical_extrapolation}.

\section{Langevin Algorithm and its Analysis}
We now turn to the core of the paper and the analysis of the Langevin
algorithm. In statistics, the most commonly used way to compute the
Bayes-optimal estimator (\ref{BO_estimator}) is to attempt to sample
the posterior distribution (\ref{posterior}) and use several
independent samples to compute the expectation in
(\ref{BO_estimator}).  In order to do that one needs to set up a
stochastic dynamics on $x$ that has a stationary measure at long times
given by the posterior measure (\ref{posterior}).  The Langevin
algorithm is one of the possibilities (others include notably Monte
Carlo Markov chain).
The common bottleneck is that the time needed to achieve stationarity can be in general exponential in the system size.
In which case the algorithm is practically useless. However, this is not always the case and there are regions in parameter space where one can expect that the
relaxation to the posterior measure happens on tractable timescales.
Therefore it is crucial to understand where this happens and what are
the associated relaxation timescales.

The Langevin algorithm on the hypersphere with Hamiltonian given by Eq.~(\ref{Hamiltonian}) reads
\begin{eqnarray}
	\dot{x}_i(t) =- \mu(t) x_i(t) -\frac{\partial
		\mathcal{H}}{\partial x_i} + \eta_i(t) \, , \label{Langevin}
\end{eqnarray}
where $\eta_i(t)$ is a zero mean noise term, with $\langle
	\eta_i(t)\eta_j(t') \rangle = 2\delta_{ij} \delta(t-t')$ where the average $\langle \cdot \rangle$ is
with respect to the realizations of the noise. The Lagrange multiplier $\mu(t)$ is chosen
in such a way that the dynamics remains on the hypersphere.
In the large $N$-limit one finds $\mu(t) =
	1 - 2 \mathcal{H}_2(t) - p \mathcal{H}_p(t)$ where the
$\mathcal{H}_2(t)$ is the 1st term from (\ref{Hamiltonian}) evaluated at $x(t)$, and $\mathcal{H}_p(t)$ is the value of the 2nd term from
(\ref{Hamiltonian}).


The presented spiked
matrix-tensor model falls into the particular class of spherical
$2+p$-spin glasses \cite{CL04,CL06} for which
the performance of the Langevin algorithm
can be tracked exactly in the large-$N$ limit via a set of integro-partial differential
equations \cite{CHS93,cugliandolo1993analytical}, beforehand dubbed CHSCK.
We call this generalised version of the CHSCK equations  \emph{Langevin State Evolution} (LSE) equations
in analogy with the state evolution
of AMP.

In order to write the LSE equations, we defined three
dynamical correlation functions
\begin{eqnarray}
	C_N(t,t') &\equiv \frac{1}{N} \sum_{i=1}^N x_i(t) x_i(t') \, ,\\
	\overline C_N(t) &\equiv \frac{1}{N} \sum_{i=1}^N x_i(t) x_i^*
	\, ,
	\\
	R_N(t,t') &\equiv \frac{1}{N} \sum_{i=1}^N   \partial
	x_i(t)/\partial h_i(t')   |_{h_i=0}  \label{RRR}\, ,
\end{eqnarray}
where $h_i$ is a pointwise external field
applied at time $t'$ to the Hamiltonian as $\mathcal{H} + \sum_i h_i x_i$.
We note that the correlation functions defined above depend on the
realization of the thermal history (i.e. of the
noise $\eta(t)$) and on the disorder (here the matrix $Y$ and tensor
$T$). However, in the large-$N$ limit they all concentrate around
their averages. We thus define $C(t,t') = \lim_{N\to \infty} {\mathbb
		E}_{Y,T} \langle C_N(t,t')  \rangle_\eta$ and analogously for
$\overline C(t)$ and $R(t,t')$.
Standard field theoretical methods \cite{MSR73} or dynamical cavity
method arguments \cite{MPV87} can then be used to
obtain a closed set of integro-differential equations for the averaged dynamical correlation
functions, describing the average \emph{global} evolution of the system under the Langevin algorithm.
The resulting LSE equations are (see the Appendix for a complete derivation)
\begin{widetext}
	\begin{equation}
		\begin{split}
			&\frac{\partial}{\partial t} C(t,t') =2R(t',t)-\mu(t)C(t,t')+
			Q'(\overline C(t)) \overline C(t') + \int_0^t dt''
			R(t,t'')Q''(C(t,t''))C(t',t'') + \int_0^{t'}dt'' R(t',t'')Q'(C(t,t''))
			\, ,\\
			&\frac{\partial}{\partial t}
			R(t,t')=\delta(t-t')-\mu(t)R(t,t')+\int_{t'}^tdt''
			R(t,t'')Q''(C(t,t''))R(t'',t')\, ,\\
			&\frac{\partial}{\partial t}  \Cmag(t) =-\mu(t)\Cmag(t)+Q'(\overline
			C(t)) + \int_{0}^tdt'' R(t,t'')\overline C(t'') Q''(C(t,t'')) \, ,
		\end{split}\label{LSE}
	\end{equation}
\end{widetext}
\noindent where we have defined $Q(x) = x^2/(2\Delta_2) + x^p/(p\Delta_p)$.
The Lagrange multiplier, $\mu(t)$, is fixed by the spherical constraint, through the condition  $C(t,t)=1\;\forall t$.
Furthermore causality implies that $R(t,t')=0$ if $t<t'$.
Finally the Ito convention on the stochastic equation (\ref{Langevin}) gives
$\forall t\; \lim_{t'\rightarrow t^-} R(t,t') = 1$.

\section{Behavior of the Langevin algorithm}
\begin{figure}
	\includegraphics[width=\columnwidth]{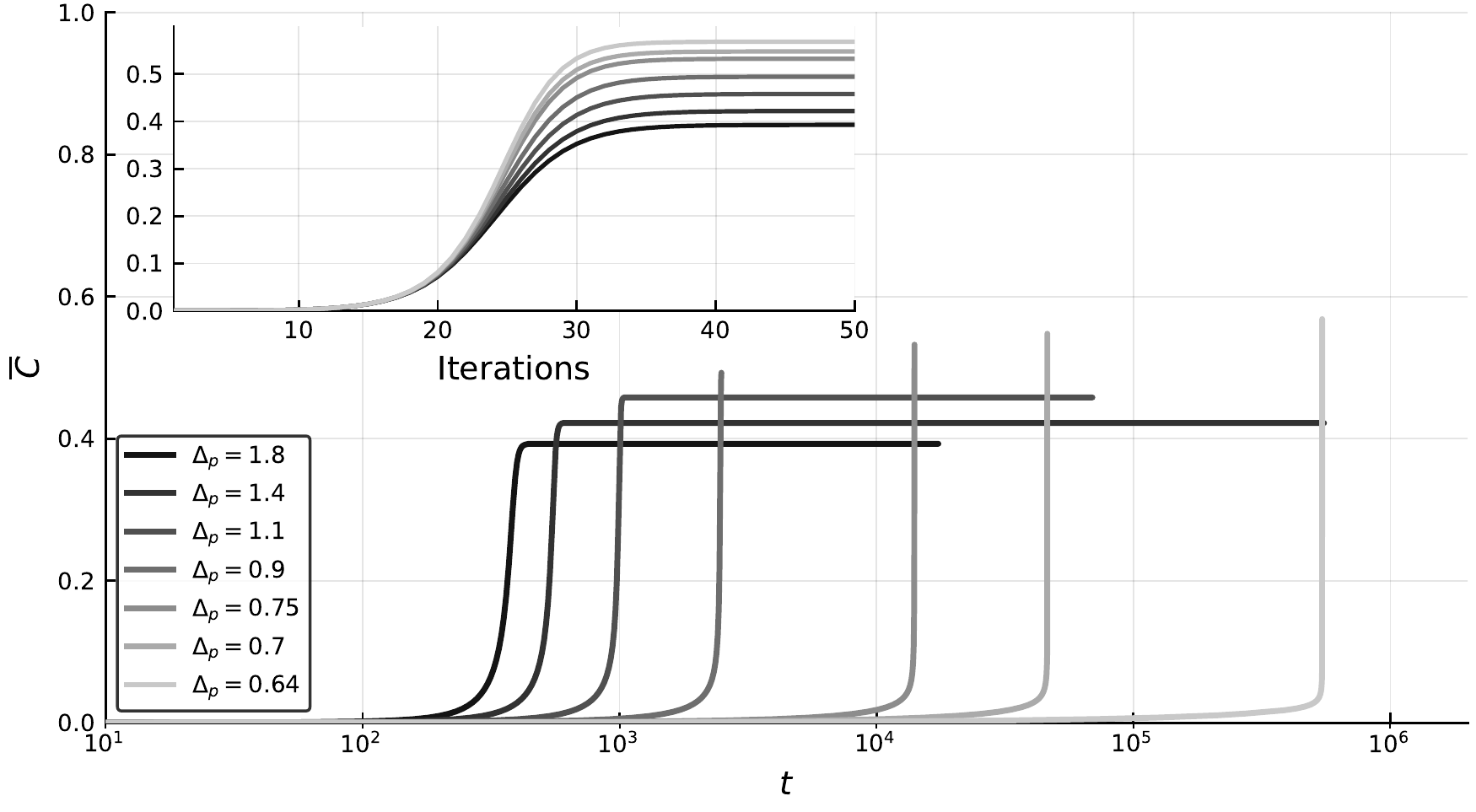}
	\caption{Evolution of the correlation with the signal $\overline C(t)$ starting from $\overline{C}(t=0) = 10^{-4}$
		in the Langevin algorithm at fixed noise on the matrix
		($\Delta_2=0.7$) and different noises on the tensor
		($\Delta_p$). As we decrease $\Delta_p$ 
                the time required to jump to the solution appears to diverge.
		Inset: the behavior of $\overline C(t)$ as a function of
		the iteration time for the AMP algorithm for the same
                values of $\Delta_p$ and the same initialization.
		\label{fig:m_t}}
\end{figure}

In order to assess the perfomances of the Langevin algorithm and
compare it with AMP, we notice that the correlation function
$\overline C(t)$ is directly related to accuracy of the algorithm. We
solve the differential equations (\ref{LSE}) numerically along the lines
of \cite{kim2001dynamics, kuni2}, for a detailed procedure see the
Appendix Sec.~\ref{SI:DMFT_derivation}, codes available online at~\cite{LSEcode}. In
Fig.~\ref{fig:m_t} we plot the correlation with the spike
$\overline C(t)$ as a function of the running time $t$ for $p=3$,
fixed $\Delta_2=0.7$ and several values of $\Delta_p$, we use as
initial condition $\overline C(t=0)=10^{-4}$. In the inset of
the plot we compare it to the same quantity obtained from the state
evolution of the AMP algorithm, with the same initial condition.
For the Langevin algorithm in Fig.~\ref{fig:m_t} we see a pattern that
is striking.  One would expect that as the noise $\Delta_p$ decreases
the inference problem is getting easier, the correlation with the
signal is larger and is reached sooner in the iteration. This is,
after all, exactly what we observe for the AMP algorithm in the inset
of Fig.~\ref{fig:m_t}.
Also for the Langevin algorithm the plateau reached for large
times $t$ becomes higher (better
accuracy) as the noise $\Delta_p$ is reduced. Furthermore the
height of the plateau coincides with that reached by AMP,
thus testifying the algorithm reached equilibrium.
However, contrary to AMP,
the relaxation time for the Langevin algorithm increases dramatically when diminishing $\Delta_p$
(notice the log scale on x-axes of Fig.~\ref{fig:m_t}, as compared to the
linear scale of the inset). 

\begin{figure}
	\centering
	\includegraphics[width=\columnwidth]{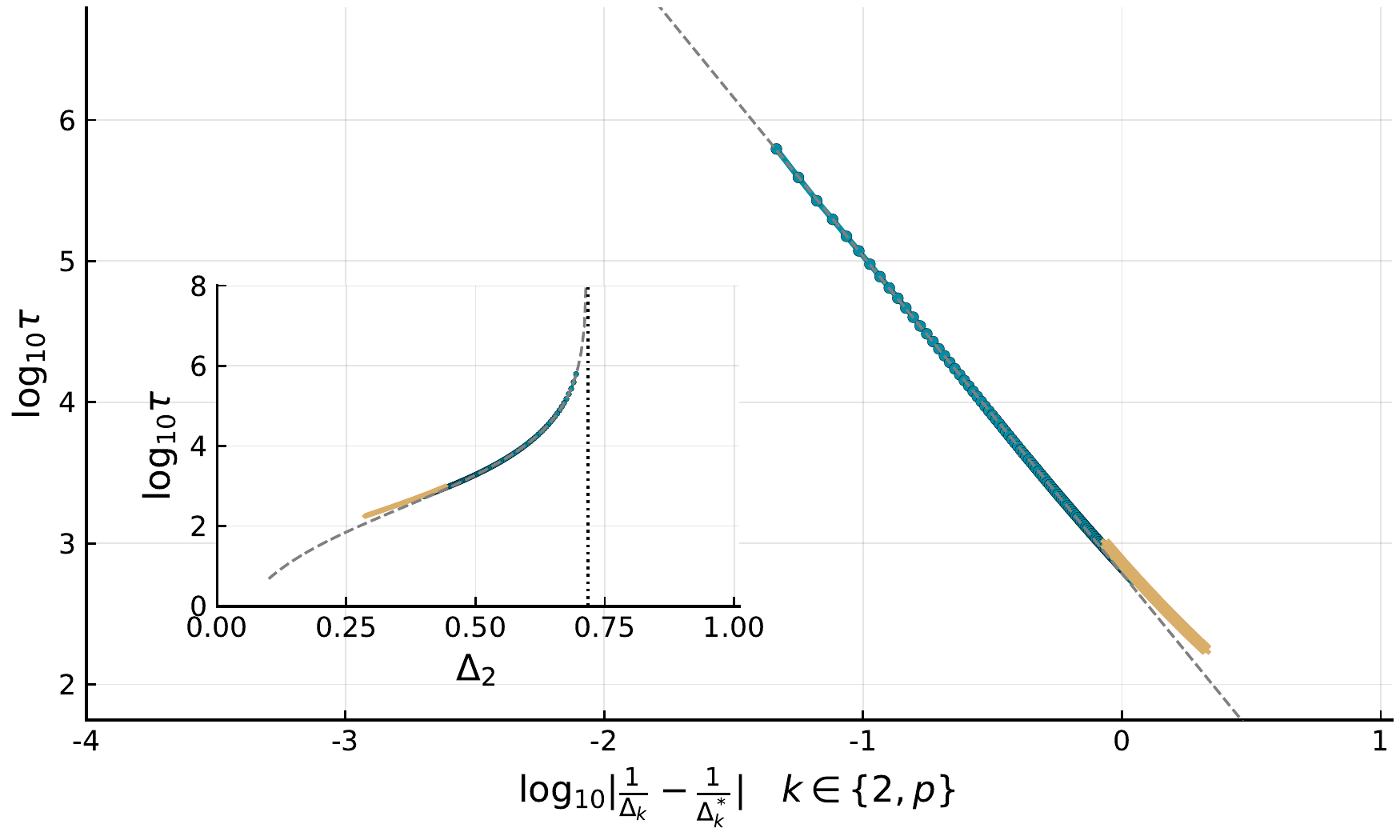}
	\caption{Extrapolation of the Langevin relaxation time. The
          inset presents the relaxation time for fixed
          $\Delta_p=1$. The main pannel then presents a fit using a
          power law consistent with a divergence at $\Delta_2^*\approx
          0.72$
                The circles are obtained with
		numerical solution of LSE that uses the dynamical grid while
		crosses are obtained using a fixed-grid, initial
                condition was $\overline{C}(t=0)= 10^{-40}$ (details in the Appendix).
		\label{convergence_time}}
\end{figure}

We define $\tau$ as the time it takes for the
correlation to reach a value $\overline C_{\rm plateau}/2$. We then
plot the value of this equilibration time in the insets of
Fig.~\ref{convergence_time} as a function of the noise 
$\Delta_2$ having fixed $\Delta_p$. The data are consistent with a divergence of $\tau$ at a
certain finite value of $\Delta^*_2$. We found that the divergence
points are affected by the initial condition of the dynamics
$\overline{C}(t=0)$, this aspect is discussed in the Appendix
Sec.~\ref{SI:numerical_initial_condition}. In the analysis of the
phase diagram we initialize the dynamics to $\overline{C}(t=0) =
10^{-40}$ (smaller values have not led to noticeable changes in $\Delta^*_2$). We calculate the divergence time and we fit the data with a power law
$\tau(\Delta) =
	\left|\frac1{\Delta}-\frac1{\Delta^*}\right|^{-\gamma}$ and we obtain in
the particular case of fixed $\Delta_p=1.0$ that
$\gamma=2.24$ and $\Delta_2^*=0.72$.  We are not
able to strictly prove that the divergence of the relaxation time
truly occurs, but at least our results imply that for
$\Delta_2>\Delta_2^*$ the Langevin
algorithm (\ref{Langevin}) is not a practical solver for the spiked
mixed matrix-tensor problem. We will call the region
$\Delta_2^*< \Delta_2<1$ where the AMP algorithm works
optimally without problems yet Langevin algorithm does not, the {\it
		Langevin-hard region}. $\Delta_2^*$ is then
plotted in Fig.~\ref{fig:phase_diagram} with green points and delimits the Langevin-hard region that extends
considerably into the region where the AMP algorithm works optimally
in a small number or iterations. Our main conclusion is thus that the
Langevin algorithm designed to sample the posterior measure works
efficiently in a considerably smaller region of parameters than the
AMP as quantified in
Fig.~\ref{fig:phase_diagram}.

\begin{figure}
	\centering
	\includegraphics[scale=.48]{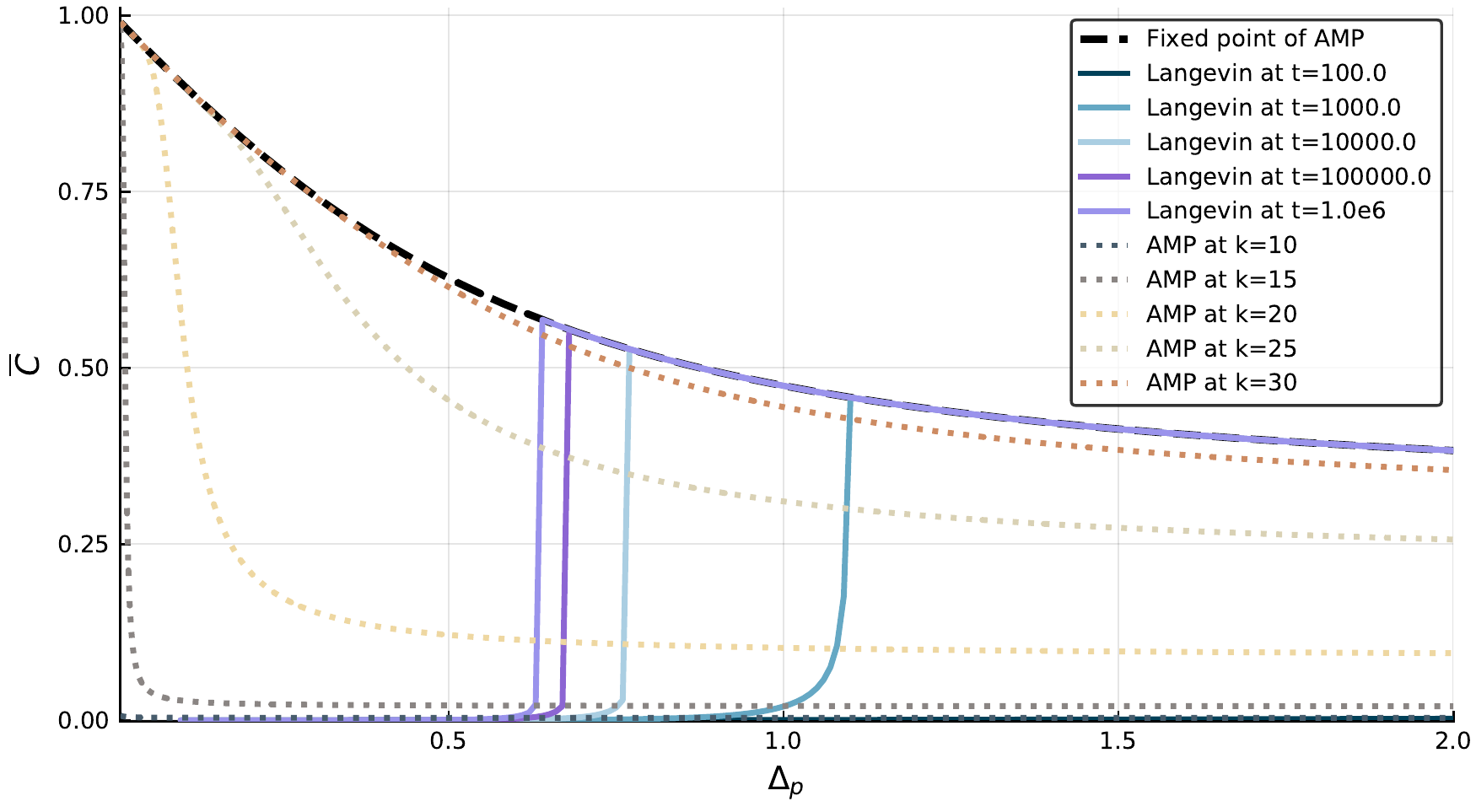}
	\caption{Correlation with the signal of AMP and Langevin
		at the $k$th iteration (at time $t$) for fixed $\Delta_2=0.7$ where both the evolutions start with initial overlap $10^{-4}$. \label{Fig:nonmonotone}}
\end{figure}

Fig.~\ref{Fig:nonmonotone} presents another way to depict the observed
data, the correlation $\overline C(t)$ reached after time $t$ is
plotted as a function of the tensor noise variance $\Delta_p$. The
results of AMP are depicted with dotted lines and, as one would expect,
decrease monotonically as the noise $\Delta_p$ increases. The equilibrium value
(black dashed) is reached within few dozens of
iterations. On the contrary, the correlation reached by the Langevin
algorithm after time~$t$ is non-monotonic and close to zero for small
values of noise $\Delta_p$ signaling again a rapidly growing relaxation time when $\Delta_p$ is decreased.

\section{Glassy nature of the Langevin-hard phase}
%

The behaviour of the Langevin dynamics as presented in the last
section might seem counter-intuitive at first sight, because one would
expect any problem to get simpler when noise $\Delta_p$ is
decreased. In the present model instead, as $\Delta_p$ is decreased,
the tensor part of the cost function (\ref{Hamiltonian}) becomes more 
important. This brings as a consequence that the landscape becomes 
rougher and causes the failure of the Langevin algorithm. 

In the presence of the hard (for AMP) phase, it was recently argued in
\cite{antenucci2018glassy} that sampling-based algorithms are indeed
to be expected to be worse than the
approximate message passing ones. This is due to residual glassiness that
extends beyond the hard phase. We repeated the analysis of \cite{antenucci2018glassy} in the present
model (details in the Appendix Sec.~\ref{sec:replica analysis}) and
conclude that while this explanation provides the correct physical picture, 
the transition line obtained in this way does not agree quantitatively
with the numerical extrapolation of the relaxation times we have
obtained numerically in the previous section, at least on the timescales on which 
we were able to solve the LSE equations. The reasons behind this
remain open.

\begin{figure}
	\centering
	\includegraphics[scale=.48]{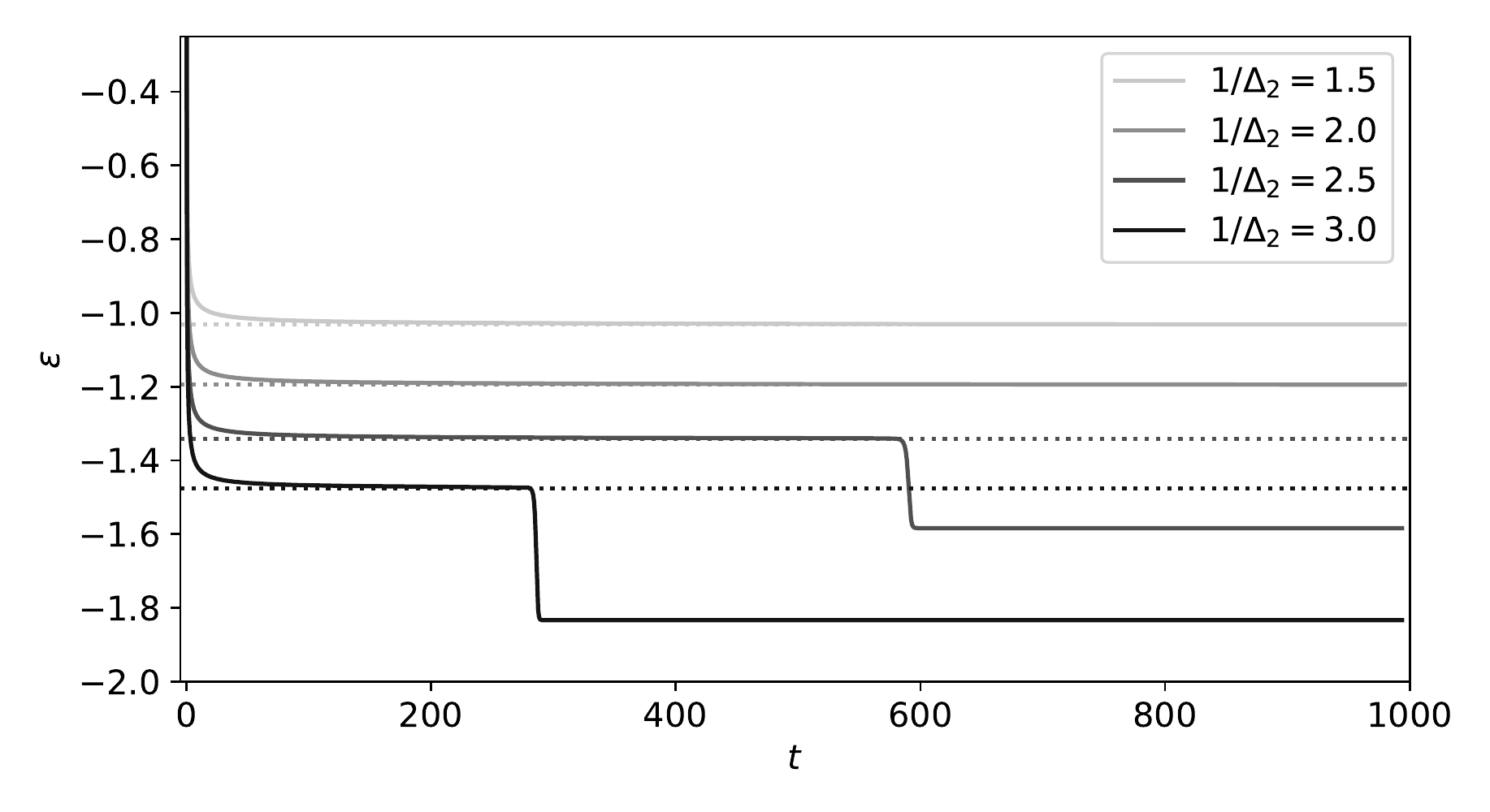}
	\caption{Dynamical evolution of the energy starting from a
          configuration with overlap $10^{-40}$ at $\Delta_p=1.0$ and
          for various $\Delta_2$. The system first tends to the
          threshold energy, Eq.~\eqref{eq:energy_th} horizontal lines,
          and then for sufficiently small $\Delta_2$ finds the direction toward the signal. \label{fig:energy_dynamics}}
\end{figure}

In order to obtain a theoretical estimate that quantitatively agrees
with the observed behaviour of the LSE, we found the following
argument. We first notice that the Langevin dynamics initialized at
very small overlap $\overline C(t=0)$ remains for a long time at
small values of the correlation with the signal. We assume that during
this time the dynamics behaves as it would in the mixed $2+p$-spin
model without the spike. The model without the spike has been studied
extensively in physics literature, precisely with the aim to
understand the dynamical properties of glasses \cite{cugliandolo1993analytical,CK95,bouchaud1998out}. 
One of the important results of those studies is
that the randomly initialized dynamics converges asymptotically to the so-called {\it  threshold states}.
Indeed in Fig.~\ref{fig:energy_dynamics} this aspect can be observed in the evolution of the energy. 
It soon approaches a value that can be evaluated \cite{cugliandolo1993analytical,CK95,bouchaud1998out}
and corresponds to threshold state energy $E^\text{\rm th}$ (horizontal lines)
\begin{equation}\label{eq:energy_th}
	E^\text{\rm th} = - Q(1) - \left[ \frac1{(1-q^\text{\rm th})Q'(q^\text{\rm th})} - \frac1{q^\text{\rm th}} \right] Q(q^\text{\rm th})\;.
\end{equation}
In the above equation $q^\text{\rm th}$ represents the correlation, a.k.a. overlap, of two configurations randomly picked from the same threshold state, which can be also evaluated as the solution of
\begin{equation}
               \frac1{1-q^\text{\rm th}} =
               \sqrt{(p-1)\frac{(q^\text{\rm th})^{p-2}}{\Delta_p}+\frac1{\Delta_2}}\,
               . \label{eq:q_th}
\end{equation}
The derivation of these expressions can be found in
Appendix~\ref{sec:replicas_1RSB} and in
Appendix~\ref{App:TAP}. Supported by the numerical results of
Fig.~\ref{fig:energy_dynamics}, we make an approximation that already on the
observed time-scales the algorithm converges to the threshold
states\footnote{This is just an approximation because the relaxation
  to the threshold state is power law and only asymptotic. Therefore our assumption is expected to provide a coarse
  grained description of the short time dynamics. Whether it provides
  an exact description of what happens on long timescales remains an open problem.}. The presence of the signal decides whether the algorithm develops a correlation with the signal. 
%
%
To understand how it occurs one has to study the statistical properties of the Thouless-Anderson-Palmer (TAP) free-energy landscape, which is the finite temperature counterpart of the energy landscape, as it has been shown in early days results of spin-glass theory \cite{TAP,MPV87} and in the recent ones of the mathematical community \cite{chen2018tap}. 
The generic picture, that comes out from several years of studies on spin-glass models, is that 
threshold states corresponds to {\it marginal} local minima of the TAP free-energy.
Critical points of the TAP free-energy functional that are below the threshold states are typically local minima, while
those above are saddles with extensively many negative directions. The
threshold states lie in between and have just a few very flat directions. 
In order to obtain an analytical prediction for the Langevin dynamics
threshold, one has to find out how the presence of the spike destabilises the threshold states.

This can be achieved by studying the free-energy Hessian at a threshold state. As shown in App. F, such Hessian reads:
\begin{equation}
\frac{\partial^2 F}{\partial m_i \partial m_j}=  G_{ij}+\delta_{ij}2\sigma_F(q_{th})-\frac{1}{\Delta_2}\frac{x_i^*x_j^*}{N}
+f''(q_{th})\frac{m_im_j}{N} \, , \label{hessian}
\end{equation} 
where $f''(q_{th})$ is positive and its expression can be found in App. F, $m_i$ is the average magnetization of site $i$ in the given threshold state, and $G_{ij}$ can be shown to be statistically equivalent to a random matrix having elements which are i.i.d. Gaussian random variables with mean zero and variance 
$$\frac{\sigma^2_F(q_{th})}{N}=\frac{(p-1)q_{th}^{p-2}}{\Delta_p}+\frac{1}{\Delta_2}\,.$$  
The free-energy Hessian evaluated at a typical threshold state is therefore a random matrix belonging to the Gaussian Orthogonal Ensemble plus two rank-one perturbations; one is negative and in the direction of the signal, whereas the other is positive and in the direction of the threshold state. \\
Results from random matrix theory allow us to completely characterise the spectral properties of the Hessian. 
Its bulk density of eigenvalues is shifted semi-circle whose left edge touches zero, hence leading to the marginality of the threshold states. For small signal to noise ratio 
the minimal eigenvalue is zero, whereas when the signal to noise ratio exceeds a certain critical value, the rank-one perturbation in the direction of the signal induces a BBP (Baik, Ben Arous, Pech\'e) transition \cite{edwards1976eigenvalue,baik2005phase}, where 
a negative eigenvalue pops out from the Wigner semi-circle, and correspondingly a downward descent direction 
toward the spike emerges and makes the threshold states unstable. Note
that the last term of the Hessian has no effect on the development of an unstable direction as it is positive and uncorrelated with the signal.
By adapting the known formulas for the BBP transition to our case, see App. F, we find a landscape-based conjecture for the algorithmic threshold, that is the larger value of
$\Delta_2^*$ between $\Delta_2^*=1$ and the roots of
\begin{equation}
     \Delta_p = (p-1) (\Delta^*_2)^2 (1-\Delta^*_2)^{p-3} \, . \label{eq:threshold}
\end{equation}
This is the threshold depicted in Fig.~\ref{fig:phase_diagram}
for $p=3$ in green dotted line. We note a very good agreement with the
data points obtained with  extrapolation of the relaxation time from numerical solution of the LSE equations. 

In the following, we present a complementary argument that interestingly also makes a direct link with AMP state evolution. 
We again assume that Langevin dynamics approaches the threshold states. 
This time we use AMP to determine whether it will remain there or not.
If the initial
correlation is $0< m^{t=0} \ll 1$ its evolution follows
\begin{equation}
 m^{t+1} = (1-q^\text{\rm th})\left[\frac{(m^t)^{p-1}}{\Delta_p}+\frac{m^t}{\Delta_2}\right]\,
 .   \label{eq:m_evol}
\end{equation}
This equation is obtained from the state evolution of the AMP
algorithm where the overlap is fixed to $q^\text{\rm th}$, as
detailed in the Appendix~\ref{subsec:SE}. 

The stability condition that decides whether an infinitesimal correlation
will grow or decrease under (\ref{eq:m_evol}) reads $q^\text{\rm th} = 1 -
\Delta_2$. Using (\ref{eq:q_th}), this then leads to eq. (\ref{eq:threshold}).

\section{Discussion and Perspectives}

Motivated by the general aim to shed light on behaviour and performance of
noisy-gradient descent algorithms that are widely used in
machine learning, we investigate analytically the performance of the
Langevin algorithm in the noisy high-dimensional limit of a spiked
matrix-tensor model.  We compare it to the performance of the approximate
message passing algorithm. While both these algorithms are designed
with the aim to sample the posterior measure of the model, we show that the Langevin algorithm
fails to find correlation with the signal in a considerable part of
the AMP-easy region. Neither of the two algorithm enters the so-called hard
phase. Our analysis is based on the Langevin State Evolution equations, generalization of
the dynamical theory for mean field spin glasses, that describe the
evolution of the algorithm in the large size limit. 

The Langevin algorithm performs worse than the AMP due to the underlying glass
transition in the corresponding region of parameters. Relying on
result from spin glass theory, we present a simple heuristic
expression of the Langevin-threshold (\ref{eq:threshold})  line which
appears to be in agreement with the value obtained from numerical
solution of the LSE equations. 

We note that so far, in our study of the spiked matrix-tensor model with
Langevin dynamics, we only accessed the cost function (\ref{Hamiltonian})
and its derivatives. We did not allow ourselves to split the cost
function in the tensor-related ${\cal H}_p$ and the matrix-related
${\cal H}_2$ parts. If we did then there is a simple way to overcome
the Langevin-hard regime by first considering only the matrix
measurements and then slowly turning on the tensor, in a similar way
as temperature is tuned in simulated annealing. We study this
procedure in the Appendix~\ref{SI:numerical_annealing}.  
It is interesting to underline that from the point of view of
Bayesian inference this finding remains somewhat paradoxical. In the setting of this paper we know perfectly the
model that generated the data and all its parameters, yet we see that
for the Langevin algorithm it is computationally advantageous to
mismatch the parameters and perform the annealing in the tensor part in
order to reach faster convergence to equilibrium. This is particularly
striking given that for AMP it has been proven in
\cite{deshpande2015finding} that mismatching the parameters can never
improve the performance.  In fact, from a physics point
  the principle thanks to which AMP does hot share the hurdles of the
  Langeving algorithm remains an interesting open question.

We stress that the above annealing procedure is a particularity of the
present model and will not generalize to a broad range of inference
problems, because it is not clear in general how to split the cost
function into simple to optimize yet informative part and the rest. 
A formidably interesting direction for future work consists instead 
in investigating whether the performance of the Langevin
algorithm can be improved in a manner that only accesses the cost
function or its derivatives.



While we studied here the spiked matrix-tensor model, we
expect that our findings, based on the existence of an underlying glass transition,  will hold more universally. We expect them to apply to
other local sampling dynamics, e.g. to Monte Carlo Markov chains, and
to a broader range of models, e.g. simple models of neural networks. 
An interesting extension of this work would be to investigate algorithms
closer to stochastic gradient descent and models
closer to current neural network architectures.

\begin{acknowledgments}

	We thank G. Folena, A. Crisanti and G. Ben Arous for precious discussions.
	We thank K. Miyazaki for sharing his code for the numerical integration of CHSCK equations.
	We acknowledge funding from the ERC under the European
	Union’s Horizon 2020 Research and Innovation Programme Grant
	Agreement 714608-SMiLe; from the European Union’s
Horizon 2020 research and innovation programme under the
Marie Skłodowska-Curie grant agreement CoSP No 823748, from the French National
	Research Agency (ANR) grant PAIL; from "Investissements d’Avenir"
	LabEx PALM (ANR-10-LABX-0039-PALM) (SaMURai
	and StatPhysDisSys); and from the Simons Foundation (\#454935, Giulio Biroli).

\end{acknowledgments}

%

\begin{widetext}
	\appendix

	\section{Definition of the spiked matrix-tensor model}\label{sec:model}

	We consider a teacher-student setting in which the teacher constructs a matrix and a tensor from a randomly sampled signal and the student is asked to recover the signal from the observation of the matrix and tensor provided by the teacher \cite{REVIEWFLOANDLENKA}.

	The signal, $x^*$ is an $N$-dimensional vector whose entries are real i.i.d. random variables sampled from the normal distribution (i.e. the prior is $P_X\sim\mathcal{N}(0,1)$).
	The teacher generates from the signal a symmetric matrix and a symmetric tensor of order $p$.
	Those two objects are then transmitted through two noisy channels with variances $\Delta_2$ and $\Delta_p$,
	so that at the end one has two noisy observations given by
	\begin{align}\label{eq:channel Y}
		 & Y_{ij} = \frac{x_i^{*}x_j^*}{\sqrt{N}}+\xi_{ij},                                                   \\
		\label{eq:channel T}
		 & T_{i_1,\dots,i_p} = \frac{\sqrt{(p-1)!}}{N^{(p-1)/2}}x_{i_1}^*\dots x_{i_p}^*+\xi_{i_1,\dots,i_p},
	\end{align}
	where, for $i< j$ and $i_1< \dots < i_p$,  $\xi_{ij}$ and
	$\xi_{i_1,\dots,i_p} $ are i.i.d. random variables distributed
	according to $\xi_{ij}\sim {\cal N}(0,\Delta_2)$ and
	$\xi_{i_1,\dots,i_p}\sim {\cal N} (0,\Delta_p)$. The $\xi_{ij}$ and
	$\xi_{i_1,\dots,i_p}$ are symmetric random matrix and tensor,
	respectively.
	Given $Y_{ij}$ and $T_{i_1,\dots,i_p}$ the inference task is to reconstruct the signal $x^*$.

	In order to solve this problem we consider the Bayesian approach.
	This starts from the assumption that both the matrix and tensor have been produced from a process of the same kind of the one described by Eq.~(\ref{eq:channel Y}-\ref{eq:channel T}).
	Furthermore we assume to know the statistical properties of the
	channel, namely the two variances $\Delta_2$ and $\Delta_p$, and the
	prior on $x$.
	Given this, the posterior probability distribution over the signal is obtained through the Bayes formula
	\begin{equation}
		P(X|Y,T) = \frac{P(Y,T|X)P(X)}{P(Y,T)}\, ,
	\end{equation}
	where
	\begin{equation} \label{eq:likelihood general}
		\begin{split}
			P(Y,T|X) &= \prod_{i< j}P_{Y}\left(Y_{ij}\Bigg|\frac{x_ix_j}{\sqrt{N}}\right)\prod_{ i_1<\dots<i_p}P_{T}\left(T_{i_1\dots i_p}\Bigg|\frac{\sqrt{(p-1)!}}{N^{(p-1)/2}}x_{i_1}\dots x_{i_p}\right) =
			\\
			&\propto \prod_{i< j}e^{-\frac1{2\Delta_2}\left(Y_{ij}-\frac{x_ix_j}{\sqrt{N}}\right)^2}\prod_{ i_1<\dots<i_p}e^{-\frac1{2\Delta_p}\left(T_{i_1\dots i_p}-\frac{\sqrt{(p-1)!}}{N^{(p-1)/2}}x_{i_1}\dots x_{i_p}\right)^2}.
		\end{split}
	\end{equation}
	Therefore we have
	\begin{equation}\label{eq:posterior general}
		\begin{split}
			P(X|Y,T) &= \frac{1}{Z(Y,T)}\prod_{i}e^{-\frac12 x_i^2}\prod_{i< j}e^{-\frac1{2\Delta_2}\left(Y_{ij}-\frac{x_ix_j}{\sqrt{N}}\right)^2}\prod_{ i_1<\dots<i_p}e^{-\frac1{2\Delta_p}\left(T_{i_1\dots i_p}-\frac{\sqrt{(p-1)!}}{N^{(p-1)/2}}x_{i_1}\dots x_{i_p}\right)^2},
		\end{split}
	\end{equation}
	where $Z(Y, T)$ is a normalization constant.

	Plugging Eqs.~(\ref{eq:channel Y}-\ref{eq:channel T}) into
	Eq.~(\ref{eq:posterior general}) allows to rewrite the posterior
	measure in the form of a \emph{Boltzmann distribution} of the mixed $2+p$-spin Hamiltonian \cite{CL04, CL06, CL13}
	\begin{equation}\label{eq:Hamiltonian}
		\begin{split}
			\mathcal{H} &= -\frac1{\Delta_2\sqrt{N}}\sum_{i<j}\xi_{ij}x_ix_j -\frac{\sqrt{(p-1)!}}{\Delta_pN^{\frac{p-1}2}}\sum_{i_1<\dots<i_p}\xi_{i_1\dots i_p}x_{i_1}\dots x_{i_p} - \frac{N}{2\Delta_2}\left(\frac1N\sum_ix_ix_i^*\right)^2 +\\
			& - \frac{N}{p\Delta_p}\left(\frac1N\sum_ix_ix_i^*\right)^p -\frac 12\sum_{i=1}^N x_i^2+ \text{const.}
		\end{split}
	\end{equation}
	so that
	\begin{equation}
		P(X|Y,T) = \frac{1}{\tilde Z(Y,T)} e^{- \mathcal{H}} \, .
		\label{Boltzmann}
	\end{equation}
	In the following we will refer to $\tilde Z(Y,T)$ as the \emph{partition function}.
	We note here that in the large $N$ limit, using a Gaussian prior on the variables $x_i$ is
	equivalent to consider a flat measure over the $N$-dimensional hypersphere $\sum_{i=1}^N x_i^2=N$.
	This choice will be used when we will describe the Langevin algorithm and in this case the last term in the Hamiltonian will become an irrelevant constant.

	\section{Approximate Message Passing, state evolution and phase diagrams}
	Approximate Message Passing (AMP) is a powerful iterative algorithms to compute the local \emph{magnetizations} $\langle x_i\rangle$ given the observed matrix and tensor.
	It is rooted in the cavity method of statistical physics of disordered
	systems \cite{TAP77, MPV87} and it has been recently developed in the
	context of statistical inference \cite{DMM09}, where in the Bayes optimal case it has been conjectured to be optimal among all local iterative algorithms.
	Among the properties that make AMP extremely useful is the fact that
	its performances can be analyzed in the thermodynamic limit. Indeed in
	such limit, its dynamical evolution is described by the so called
	State Evolution (SE) equations \cite{DMM09}.
	In this section we derive the AMP equations and their SE description for the spiked matrix-tensor model and solve them to obtain the phase diagram of the model as a function of the variances $\Delta_2$ and $\Delta_p$ of the two noisy channels.

	\subsection{Approximate Message Passing and Bethe free entropy}\label{SI:AMP}

	AMP can be obtained as a relaxed Gaussian closure of the Belief Propagation (BP) algorithm. The derivation that we present follows the same lines of \cite{MiolaneTensor,LKZ17}.
	The posterior probability can be represented as a factor graph where all the variables are represented by circles and are linked to squares representing the interactions \cite{MM09}.
	\begin{figure}[H]
		\centering
		\includegraphics[scale=0.7]{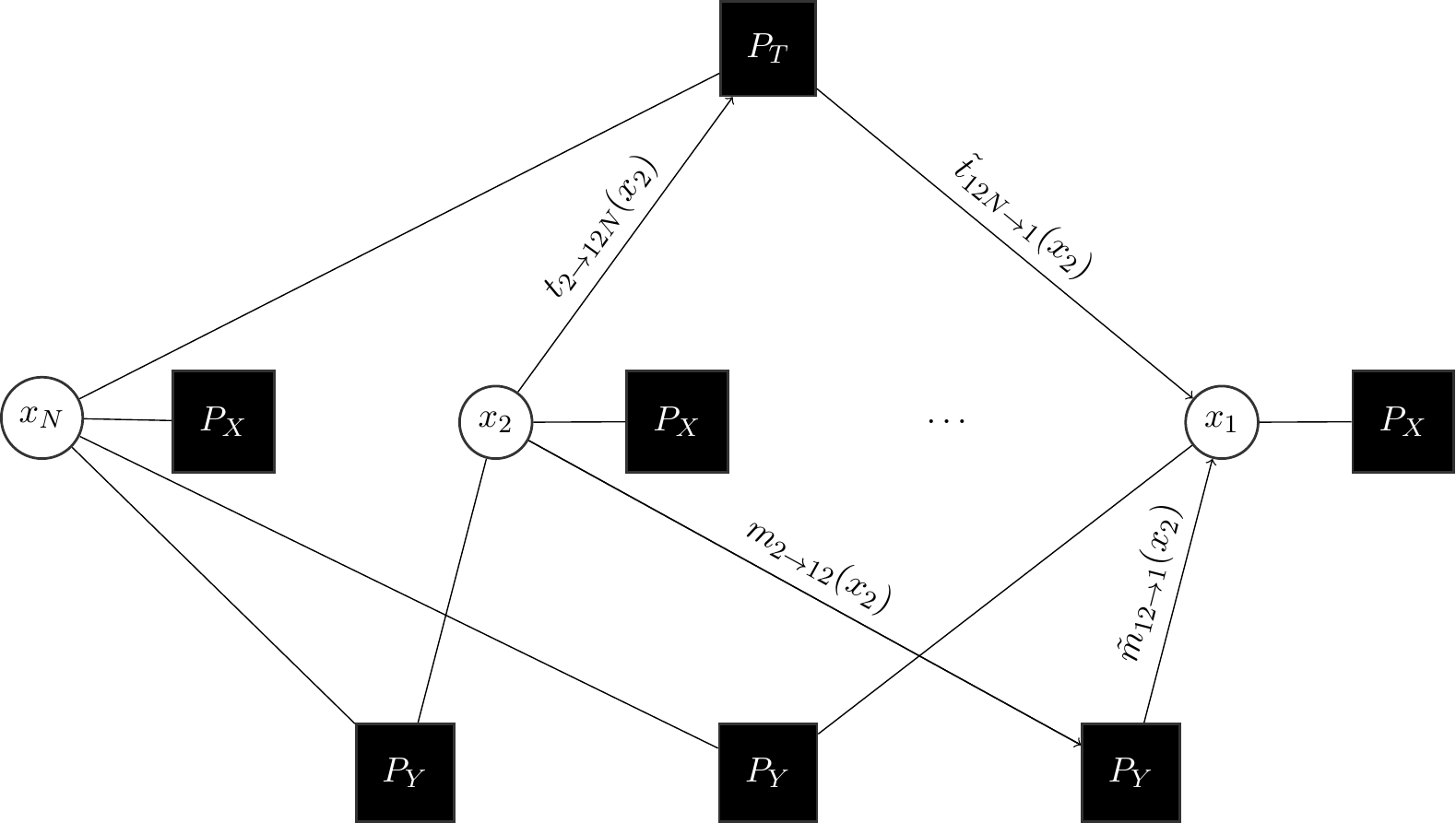}
		\caption{The factor graph representation of the posterior
			measure of the matrix-tensor factorization model. The variable nodes represented with white
			circles are the components of the signal while black
			squares are factor nodes that denote interactions between
			the variable nodes that appear in the interaction
			terms of the Boltzmann distribution in Eqs.~(\ref{eq:Hamiltonian}-\ref{Boltzmann}).
			There are three types of factor nodes: $P_X$ is the
			prior that depends on a single variable, $P_Y$ that is the
			probability of observing a matrix element $Y_{ij}$ given the
			values of the variables $x_i$ and $x_j$, and finally $P_T$
			that is the probability of observing a tensor element $T_{i_1,\dots,i_p}$.
			The posterior, apart from the normalization factor, is simply given by the product of all the factor nodes.}
	\end{figure}
	This representation is very convenient to write down the BP equations. In the BP algorithm we iteratively update until convergence a set of variables, which are beliefs of the (cavity) magnetization of the nodes. The intuitive underlying reasoning behind how BP works is the following. Given the current state of the variable nodes, take a factor node and exclude one node among its neighbors. The remaining neighbors through the factor node express a belief on the state of the excluded node. This belief is mathematically described by a probability distribution called \textit{message}, $\tilde{m}^t_{ij\rightarrow i}(x_i)$ and $\tilde{t}^t_{ii_2\dots i_p\rightarrow i}(x_i)$ depending on which factor node is selected. At the same time, another belief on the state of the excluded node is given by the rest of the network but the factor node previously taken into account, $m_{i\rightarrow ij}(x_i)$ and $t_{i\rightarrow ii_2\dots i_p}(x_i)$ respectively. All these \textit{messages} travel in the factor graph carrying partial information on the real magnetization of the single nodes, and they are iterated until convergence. The iterative scheme is described by the following equations
	\begin{align}
		\tilde{m}^{t}_{ij\rightarrow i}(x_i)            & \propto\int dx_j m^t_{j\rightarrow ij}(x_j)P_Y\left(Y_{ij}\Bigg|\frac{x_ix_j}{\sqrt{N}}\right),\label{BP1} \\
		m^{t+1}_{i\rightarrow ij}(x_i)                  & \propto P_X(x_i)\prod_{l\neq j}
		\tilde{m}^t_{il\rightarrow
		i}(x_i)\prod_{i_2 < \dots
			<  i_p} \tilde{t}^t_{ii_2\dots i_p\rightarrow i}(x_i),                                                                                                       \\
		\tilde{t}^{t}_{ii_2\dots i_p\rightarrow i}(x_i) & \propto\int
		\prod_{l=2\dots
			p}\left(dx_l
		t^t_{i_l\rightarrow
		ii_2\dots
		i_p}(x_l)\right)
		P_T\left(T_{ii_2\dots
				i_p}\Bigg|\frac{\sqrt{(p-1)!}}{N^{(p-1)/2}}x_{i}
		x_{i_2}\dots
		x_{i_p} \right),                                                                                                                                             \\
		t^{t+1}_{i\rightarrow ii_2\dots i_p}(x_i)       & \propto
		P_X(x_i)\prod_{l}
		\tilde{m}^t_{il\rightarrow
		i}(x_i)\prod_{k_2
			< \dots <
			k_p\neq i_2\dots i_p} \tilde{t}^t_{ik_2\dots k_p\rightarrow i}(x_i)
		\label{BP4}
	\end{align}
	and we have omitted the normalization constants that guarantee that the messages are probability distributions.
	When the messages have converged to a fixed point, the estimation of the local magnetizations can be obtained through the computation of the real marginal probability distribution of the variables given by
	\begin{equation}
		\mu_i(x_i)=\int\left[\prod_{j(\neq i)}dx_j\right]
		P(X|Y,T)=P_X(x_i)\prod_{l} \tilde{m}^t_{il\rightarrow
		i}(x_i)\prod_{i_2< \dots< i_p} \tilde{t}^t_{ii_2\dots
		i_p\rightarrow i}(x_i)\,.
	\end{equation}
	We note that the computational cost to produce an iteration of BP  scales as $O(N^p)$. Furthermore Eqs.~(\ref{BP1} -\ref{BP4}) are iterative equations for continuous functions and therefore
	are extremely hard to solve when dealing with continuous variables.
	The advantage of AMP is to reduce drastically the computational
	complexity of the algorithm by closing the equations on a Gaussian ansatz for the messages.
	This is justified in the present context since the factor graph is
	fully connected and therefore each iteration step of the algorithm
	involves sums of a large number of independent random variables that give rise to Gaussian distributions.
	Gaussian random variables are characterized by their mean and
	covariance that are readily obtained for $N\gg 1$ expanding the factor nodes for small $\omega_{ij} =
		x_i x_j   /  \sqrt{N}$ and $\omega_{i_1\dots i_p} =
		\sqrt{(p-1)!} x_1 \dots x_p   / N^{\frac{p-1}2}$.

	Once the BP equations are relaxed on Gaussian messages, the final
	step to obtain the AMP algorithm is the so-called TAPyfication procedure \cite{LKZ17, TAP77},
	which
	exploits the fact that the procedure of removing one node or one factor produces only a weak perturbation
	to the real marginals and therefore can be described in terms of the real marginals of the variable nodes themselves.
	By applying this scheme we obtain
	the AMP equations, which are described by a set of auxiliary variables
	$A^{(k)}$ and $B^{(k)}_i$ and by the mean $\langle x_i\rangle$ and variance $\sigma_i=\langle x_i^2\rangle$ of the marginals of variable nodes.
	The AMP iterative equations are
	\begin{align}\label{eq:AMP B2}
		B^{(2),t}_{i}   & =\frac{1}{\Delta_2\sqrt{N}}\sum_{k}Y_{ki}\hat{x}_k^t-\frac1{\Delta_2}\left(\frac1{N}\sum_k\sigma_k^t\right)\hat{x}_i^{t-1}\,;
		\\\label{eq:AMP A2}
		A^{(2),t}       & =\frac{1}{\Delta_2 N}\sum_{k}\left(\hat{x}_k^t\right)^2\,;
		\\\label{eq:AMP Bp}
		B^{(p),t}_{i}   & =\frac{\sqrt{(p-1)!}}{\Delta_p N^{(p-1)/2}}\sum_{k_2\dots k_p}T_{ik_2\dots k_p}\left(\hat{x}_{k_2}^t\dots \hat{x}_{k_p}^t\right)
		-\frac{p-1}{\Delta_p}\left[\left(\frac1{N}\sum_k\sigma_k^t\right)\;\left[\frac1{N}\sum_k
		\hat{x}_k^t\hat{x}_k^{t-1}\right]^{p-2}\right]\hat{x}_i^{t-1}\,;
		\\\label{eq:AMP Ap}
		A^{(p),t}       & =\frac1{\Delta_p}\left[\frac1{N}\sum_k\left(\hat{x}_k^t\right)^2\right]^{p-1}\,;
		\\\label{eq:AMP x}
		\hat{x}_i^{t+1} & =f(A^{(2)}+A^{(p)},B^{(2)}_i+B^{(p)}_i)\,;
		\\\label{eq:AMP sigma}
		\sigma_i^{t+1}  & =\left.\frac{\partial}{\partial B}f(A,B)\right|_{A=A^{(2)}+A^{(p)},B=B^{(2)}_i+B^{(p)}_i}\,,                                     \\
		f(A,B)          & \equiv \int dx \frac1{{\cal Z}(A,B)}xP_X(x)e^{Bx-\frac12Ax^2} =\frac{B}{1+A}\:.
	\end{align}
	It can be shown that these equations can be obtained as saddle point equations from the so called Bethe free entropy
	defined as
	$\Phi_\text{Bethe} = \log Z^{\rm Bethe}(Y,T)/N$ where $Z^{\rm Bethe}$ is the Bethe approximation to the partition function which is defined as the normalization of the posterior measure.
	The expression of the Bethe free
	entropy per variable can be computed in a standard way (see \cite{MM09}) and it is given by
	\begin{equation}
		\Phi_\text{Bethe} = \frac{1}{N} \left( \sum_i\log Z_i + \sum_{i\le j}\log Z_{ij} +
		\sum_{i_1\le \dots \le i_p}\log Z_{i_1\dots i_p} -
		\sum_{i(ij)}\log Z_{i,ij} - \sum_{i(ii_2\dots i_p)}\log
		Z_{i(ii_2\dots i_p)} \right)\;,
	\end{equation}
	where
	\begin{align*}
		 & Z_i = \int dx_i P_X(x_i) \prod_j \tilde{m}_{ij\rightarrow i}(x_i) \prod_{(i_2\dots i_p)} \tilde{t}_{ii_2\dots i_p\rightarrow i}(x_i),
		\\
		 & Z_{ij} = \int \prod_{j(\neq i)}\left[dx_{j}m_{j\rightarrow
		ij}(x_j)\right]   \prod_{i<
			j}e^{-\frac1{2\Delta_2}\left(Y_{ij}-\frac{x_ix_j}{\sqrt{N}}\right)^2}\, ,
		\\
		 & Z_{i_1\dots i_p} = \int
		\prod_{l=1}^p\left[dx_{i_l}t_{i_l\rightarrow i_1\dots
		i_p}(x_{i_l})\right] \prod_{
			i_1<\dots<i_p}e^{-\frac1{2\Delta_p}\left(T_{i_1\dots
				i_p}-\frac{\sqrt{(p-1)!}}{N^{(p-1)/2}}x_{i_1}\dots
		x_{i_p}\right)^2}\, ,
		\\
		 & Z_{i(ij)} = \int dx_im_{i\rightarrow ij}(x)\tilde{m}_{ij\rightarrow i}(x_i),
		\\
		 & Z_{i(ii_2\dots i_p)} = \int dx_it_{i\rightarrow ii_2\dots i_p}(x)\tilde{t}_{ii_2\dots i_p\rightarrow i}(x_i)
	\end{align*}
	are a set of normalization factors. Using the Gaussian approximation for the messages and employing the same TAPyification procedure used to get the AMP equations
	we obtain the Bethe free entropy density as
	\begin{equation}\label{eq:free entropy Bethe}
		\begin{split}
			\Phi_\text{Bethe} &=\frac{1}{N}\sum_i \log
			{\cal Z}(A^{(p)}+A^{(2)},B^{(p)}_i+B^{(2)}_i)+\frac{p-1}p \frac{1}{N} \sum_i\Bigg[-B_i^{(p)}\hat{x}_i + A_i^{(p)}\frac{\hat{x}_i^2+\sigma_i}2\Bigg]+\\
			&+\frac{p-1}{2p\Delta_p}\left(\frac{\sum_i\hat{x}_i^2}N\right)^{p-1}\left(\frac{\sum_i\sigma_i}N\right)+\frac{1}{2N}\sum_i\Bigg[-B_i^{(2)}\hat{x}_i
			+
			A_i^{(2)}\frac{\hat{x}_i^2+\sigma_i}2\Bigg]+\frac{1}{4\Delta_2}\left(\frac{\sum_i\hat{x}_i^2}N\right)\left(\frac{\sum_i\sigma_i}N\right)\; ,
		\end{split}
	\end{equation}
	where we used the variables defined in eqs. (\ref{eq:AMP B2}-\ref{eq:AMP Ap}) for sake of compactness
	and ${\cal Z}(A,B)$ is defined as
	\begin{equation}
		{\cal Z}(A,B)=\int dx P_X(x)e^{Bx-\frac{Ax^2}2} =
		\frac1{\sqrt{A+1}}e^{\frac{B^2}{2(A+1)}}\;.  \label{eq:ZAB}
	\end{equation}

	\subsection{Averaged free entropy and its proof}\label{SI:AMP_proof_RS}

	Eq.~(\ref{eq:free entropy Bethe}) represents the Bethe free entropy for a single realization of the factor nodes in the large size limit. Here we wish to discuss the actual, exact, value of this free entropy, that is:
	\[f_N(Y,T)=\frac {\log Z(Y,T)}N , \]\,
where the partition function $Z(Y,T)$ is defined as the normalization
of the posterior probability distribution, eq.~(\ref{posterior}).
	The free entropy is a random variable, since it depends a priori on the planted signal and the noise in the tensor and matrices. However one expects that, since free entropy is an intensive quantity, we expect from the statistical physics intuition that  it should be self averaging and concentrate around its mean value in the large $N$ limit \cite{MPV87}. In fact, this is easily proven. First, since the spherical model has a rotational symmetry, one may assume the planted assignment could be any vector on the hyper-sphere, and we might as well suppose it is the uniform one $x_i^*=1 \forall ~i$: the true source of fluctuation comes from the noise $Y$ and $T$. These can be controlled by noticing that the free entropy is a Lipshitz function of the Gaussian random variable $Y$ and $T$. Indeed:
	\[ \partial_{Y_{ij}} f_N(Y,T)=\frac 1{\Delta_2 N
            \sqrt{N}}\langle  x_i x_j \rangle \]\, ,
	so that the free energy $f_N$ is Lipschitz with respect to $Y$ with constant
	\[L=\frac 1{\Delta_2 N \sqrt{N}} \sqrt{\sum_{i<j} \langle  x_i x_j \rangle^2 }\le  \frac 1{\Delta_2 N \sqrt{N}} \sqrt{\frac12 \sum_{i,j} \langle  x_i x_j \rangle^2 }
		= \frac 1{\Delta_2 N \sqrt{N}} \sqrt{\frac 12
                  \sum_{i,j} \langle  x_i \tilde x_i x_j \tilde
                  x_j\rangle } \]\, ,
	where  $\tilde x$ represent a copy (or replica) of the system. In this case
	\[L \le \frac 1{\Delta_2 N \sqrt{N}} \sqrt{N^2 \langle \left( \frac{\sum_i x_i \tilde x_i}{N}\right)^2\rangle }=
		\frac {\sqrt{\langle q^2\rangle}}{\Delta_2
                  \sqrt{N}} \]\, ,
	where $q$ is the overlap between the two replica $x$ and $\tilde x$, that is bounded by one on the sphere, so $L \le \frac 1{\Delta_2 \sqrt{N}}$. Therefore, by Gaussian concentration of Lipschitz functions (the Tsirelson-Ibragimov-Sudakov inequality \cite{boucheron2004concentration}), we have for some constant $K$:
	\begin{equation}
		\textrm{Pr}\left[ | f_n-\mathbb{E}_Y f_n | \ge t
                \right]\le 2 e^{-N t^2/K}\, ,
	\end{equation}
	and it particular any fluctuation larger than $O(1/\sqrt{N})$ is (exponentially) rare. A similar computaton shows that $f_N$ also concentrates with respect to the tensor $T$. This shows that in the large size limit, we can consider the averaged free entropy:
	\[ {\cal F}_N \equiv \frac{1}{N} \mathbb{E} \left[ \log Z_N
			\right] \]\, .

	With our (non-rigorous) statistical physics tools, this can be obtained by averaging Eq.~(\ref{eq:free entropy Bethe}) over the disorder, see for instance \cite{LKZ17}, and this yields an expression for the free energy called the replica symmetric (RS) formula:
	\begin{equation}
		\Phi_\text{RS}=\lim_{N \to \infty} \mathbb{E}_{Y,T} \frac {\log Z(Y,T)}N\,.
		\label{RS_result}
	\end{equation}
	We now state precisely the form of $\Phi_{\rm RS}$ and prove the validity of Eq.~(\ref{RS_result}).
	The RS free entropy for any prior distribution $P_X$ reads as
	\begin{equation}\label{eq:free entropy generic}
		\begin{split}
			\Phi_\text{RS}& \equiv {\rm max}_m \tilde\Phi_\text{RS}(m)  \quad
			{\rm where} \\
			\tilde\Phi_\text{RS}(m) &=\mathbb{E}_{W,x^*}\left[\log\left[{\cal Z}\left(\frac{m}{\Delta_2}+\frac{m^{p-1}}{\Delta_p},\left(\frac{m}{\Delta_2}+\frac{m^{p-1}}{\Delta_p}\right)x^*+\sqrt{\frac{m}{\Delta_2}+\frac{m^{p-1}}{\Delta_p}}W\right)\right]\right]
			-\frac1{4\Delta_2}m^2-\frac{p-1}{2p\Delta_p}m^p\;,
		\end{split}
	\end{equation}
	where $W$ is a Gaussian random variable of zero mean and unit variance
	and $x^*$ is a random variables taken from the prior $P_X$.
	We remind that the function ${\cal Z}(A,B)$ is defined via Eq.~(\ref{eq:ZAB}).

	For Gaussian prior $P_X$, which is the one of interest here, we obtain
	\begin{equation}
		\tilde\Phi_\text{RS}(m)=-\frac12\log\left(\frac{m}{\Delta_2}+\frac{m^{p-1}}{\Delta_p}+1\right)+\frac12\left(\frac{m}{\Delta_2}+\frac{m^{p-1}}{\Delta_p}\right)-\frac1{4\Delta_2}m^2-\frac{p-1}{2p\Delta_p}
		m^p\;.
		\label{eq:Phi_sph}
	\end{equation}
	The expression given in the main text is slightly different but can be
	obtained as follow. First notice that the extremization condition for
	$\tilde\Phi_\text{RS}(m)$ reads
	\begin{equation}
		m=
                1-\frac1{1+\frac{m}{\Delta_2}+\frac{m^{p-1}}{\Delta_p}}
		\label{eq:extrem}
	\end{equation}
	and by plugging this expression in Eq.~(\ref{eq:Phi_sph}) we recover the more compact expression
	$\Phi_\text{RS}(m)$ showed in the main text:
	\begin{equation}
		\Phi_\text{RS}(m)=\frac12\log\left(1-m\right)+\frac{m}2+\frac{m^2}{4\Delta_2}+\frac{m^p}{2p\Delta_p}\;.
	\end{equation}
	The two expressions $\Phi_\text{RS}(m)$ and $\tilde \Phi_\text{RS}(m)$
	are thus equal for each value of $m$ that satisfy Eq.~(\ref{eq:extrem}).
	The parameter $m$ can be interpreted as the average correlation between the true and the estimated signal
	\begin{equation}
		m=\frac{1}{N} \sum_{i=1}^N x_i^*\hat x_i\:.
	\end{equation}
	The average minimal mean squared error (MMSE) can be obtained from the maximizer $m$ of the average Bethe free entropy as
	\begin{equation}
		{\rm MMSE}\equiv \frac{1}{N}\sum_{i=1}^N \overline{(x_i^*-\hat x_i)^2} = 1 - m^*\, , \quad {\rm where} \quad    m^* = {\rm
		argmax}\; \tilde\Phi_\text{RS}(m) \, .
		\label{V-MMSE}
	\end{equation}
	where the overbar stands for the average over the signal $x^*$ and the noise of the two Gaussian channels.

	The validity of Eq.~(\ref{eq:free entropy generic}) can be
	proven rigorously for every prior having a bounded second moment. The
	proof we shall present is a straightforward generalization of the one
	presented in \cite{MiolaneTensor} for the pure tensor case, and in
	\cite{MiolaneXX} for the matrix case, and it is based on two main
	ingredients. The first one is the Guerra interpolation method applied on the Nishimori line
	\cite{korada2009exact,krzakala2016mutual,MiolaneXX}, in which we
	construct an interpolating Hamiltonian that depends on a parameter $t\in[0;1]$ that is
	used to move from the original Hamiltonian of Eq.~(\ref{eq:Hamiltonian}), to the one
	corresponding to a scalar denoising problem whose free entropy is
	given by the first term in Eq.~(\ref{eq:free entropy generic}).  The second ingredient is
	the Aizenman-Sims-Starr method \cite{aizenman2003extended} which is
	the mathematical version of the cavity method (note that other
	techniques could also be employed to obtain the same results, see
	\cite{XXT,barbier_stoInt,AlaouiKrzakala,mourrat2018hamilton}). The theorem we want to prove is:

	\begin{theorem}[Replica-Symmetric formula for the free energy]\label{thm1}
		Let $P_X$ be a probability distribution over \,$\R$, with
		finite second moment $\Sigma_X$. Then, for all $\Delta_2>0$ and $\Delta_p>0$
		\begin{equation} \label{eq:lim_F}
			{\cal F}_N \equiv \frac{1}{N} \mathbb{E} \left[ \log Z_N
				\right] \xrightarrow[N \to \infty]{} \sup_{m \geq 0}
			\tilde  \Phi_{\rm RS}(m) \equiv \Phi_\text{\rm RS}(\Delta_2,\Delta_p)\,.
		\end{equation}
		For almost every $\Delta_2>0$ and $\Delta_p>0$, $\tilde \Phi_{\rm RS}$
		admits a unique maximizer $m$ over $\R_+ \times \R_+$ and
		$$
			{\rm T-MMSE}_N \xrightarrow[N \to \infty]{}
			\Sigma_X^p - (m^*)^p \, ,
		$$
		$$
			{\rm M-MMSE}_N \xrightarrow[N \to \infty]{}
			\Sigma_X^2 - (m^*)^2 \,.
		$$
	\end{theorem}
	Here, we have defined the tensor-MMSE  $\text{T-MMSE}_N$ by the error in
	reconstructing the tensor:
	\begin{align*}
		\text{T-MMSE}_N(\Delta_2,\Delta_p)=
		\inf_{\hat{\theta}} \left\{
		\frac{p!}{N^p} \sum_{i_1 < \dots < i_p} \left(
		x^0_{i_1} \dots x^0_{i_p}
		- \hat{\theta}(Y)_{i_1 \dots i_p}
		\right)^2
		\right\} \,,
	\end{align*}
	and the  matrix-MMSE  $\text{M-MMSE}_N$ by the error in
	reconstructing the matrix:
	\begin{align*}
		\text{M-MMSE}_N(\Delta_2,\Delta_p)=
		\inf_{\hat{\theta}} \left\{
		\frac{2}{N^2} \sum_{i < j} \left(
		x^0_{i} x^0_{j}
		- \hat{\theta}(Y)_{i,j}
		\right)^2
		\right\} \,,
	\end{align*}
	where in both cases the infimum is taken over all measurable functions
	$\hat{\theta}$ of the observations $Y$.

	The result concerning the MMSE is a simple application of the I-MMSE
	theorem \cite{GuoShamaiVerdu_IMMSE}, that relates the derivative of the
	free energy with respect to the noise variances and the MMSE.
	The details of the arguments are the same than in
	the matrix ($p=2$) case (\cite{MiolaneXX}, corollary 17) and the tensor
	one (\cite{MiolaneTensor}, theorem 2). Indeed, as discussed  in \cite{MiolaneXX, MiolaneTensor},
	these M-MMSE and T-MMSE results implies the vector MMSE
	result of Eq.~(\ref{V-MMSE}) when $p$ is odd, and thus in particular for the
	$p=3$ case discussed in the main text.

	\paragraph{Sketch of proof}
	In this section we give a detailed sketch of the proof theorem
	$1$. Following the techniques used in many recent works
	\cite{korada2009exact,krzakala2016mutual,
		XXT,MiolaneXX,MiolaneTensor,barbier_stoInt,AlaouiKrzakala,barbier2017phase}, we shall
	make few technical remarks:
	\begin{itemize}
		\item We will consider only priors with bounded
		      support, $\text{supp}(P_X) = S \subset [-K;K]$. This allows to switch
		      integrals and derivatives without worries. This condition can then
		      be relaxed to unbounded distributions with bounded second moment
		      using the same techniques as the ones that we are going to present, and
		      the proof is therefore valid in this case. This is detailed for
		      instance in \cite{MiolaneXX} sec. 6.2.2.
		\item Another key ingredient is the introduction of a
		      small perturbation in the model that takes the form of a small
		      amount of side information.  This kind of techniques are frequently
		      used in statistical physics, where a small ``magnetic field''
		      forces the Gibbs measure to be in a single pure state
		      \cite{georgii2011gibbs}. It has also been used in the context
		      of coding theory \cite{Macris2007} for the same reason.
		      In the context of Bayesian inference, we follow the generic scheme
		      proposed by Montanari in \cite{andrea2008estimating} (see also
		      \cite{coja2017information}) and add a small additional source of
		      information that allows the system to be in a single pure state so that
		      the overlap concentrates on a single value.  This source depends on
		      Bernoulli random variables
		      $L_i\stackrel{\text{i.i.d.}}{\sim}\text{Bern}(\epsilon)$, $i\in[N]$;
		      if $L_i=1$, the channel, call it $A$, transmits the correct
		      information. We can then consider the posterior of this new problem,
		      $P(X|A,Y,T)$, and focus on the associated free energy density
		      $F_{N,\epsilon}$ defined as the expected value of the average of the
		      logarithm of normalization constant divided by the number of
		      spins. Then we can immediately prove that for all $N\ge1$ and
		      $\epsilon,\epsilon'\in[0;1]$ it follows:
		      $\left|F_{N,\epsilon}-F_{N,\epsilon'}\right|\leq
			      \left(\frac{K^{2p}}{\Delta_p}+\frac{K^4}{\Delta_2}\right)|\epsilon-\epsilon'|$. This
		      allows (see for instance \cite{MiolaneTensor}) to obtain the
		      concentration of the posterior distribution around the replica
		      parameter ($q=\frac1N\langle x^{(1)}\cdot x^{(2)}\rangle$)
		      \begin{align}
			       & \mathbb{E} \left\langle\left(\frac{x^{(1)}\cdot x^{(2)}}N - q\right)^2\right\rangle \stackrel{N\rightarrow\infty}{\longrightarrow0} \;; \\
			       & \mathbb{E} \left\langle\left(\frac{x^*\cdot x}N -
			      q\right)^2\right\rangle
			      \stackrel{N\rightarrow\infty}{\longrightarrow0} \;,
		      \end{align}
		      where $x,x^{(1)},x^{(2)}$ are sampled from the posterior distribution
		      and the averages $\langle\cdot\rangle$ and $\mathbb{E}[\cdot]$ are
		      respectively the average over the posterior measure and the
		      remaining random variables.
		\item Finally, a fundamental property of inference problems which is a direct
		      consequence of the Bayes theorem, and of the fact that we are in the Bayes optimal setting where we know the statistical properties of the signal, namely the prior, and the statistical properties of the channels, namely $\Delta_2$ and $\Delta_p$, is
		      the so-called Nishimori symmetry \cite{nishimori2001statistical,REVIEWFLOANDLENKA}: Let $(X,Y)$ be a couple of random variables
		      on a polish space. Let $k \geq 1$ and let $X^{(1)}, \dots, X^{(k)}$
		      be $k$ i.i.d.\ samples (given $Y$) from the distribution
		      $P(X=\cdot \, | \, Y)$, independently of every other random
		      variables. Let us denote $\langle \cdot \rangle$ the expectation
		      with respect to $P(X=\cdot \, | \, Y)$ and $\mathbb{E}$ the
		      expectation with respect to $(X,Y)$. Then, for all continuous
		      bounded function $f$
		      $$
			      \mathbb{E} \langle f(Y,X^{(1)}, \dots, X^{(k)}) \rangle
			      =
			      \mathbb{E} \langle f(Y,X^{(1)}, \dots, X^{(k-1)}, X) \rangle \,.
		      $$
		      While the consequences of this identity are important, the proof is rather simple: It
		      is equivalent to sample the couple $(X,Y)$ according to its
		      joint distribution or to sample first $Y$ according to its
		      marginal distribution and then to sample $X$ conditionally to
		      $Y$ from its conditional distribution $P(X=\cdot \, | \,
			      Y)$. Thus the $(k+1)$-tuple $(Y,X^{(1)}, \dots,X^{(k)})$ is
		      equal in law to $(Y,X^{(1)},\dots,X^{(k-1)},X)$.
	\end{itemize}
	The proof of Theorem \ref{thm1} is obtained by using the Guerra interpolation technique to prove a lower bound for the free entropy and
	then by applying the Aizenman-Sims-Star scheme to get a matching upper bound.

	\paragraph{Lower bound: Guerra interpolation} We now move to the core
	of the proof. The first part combines the Guerra interpolation method
	\cite{guerra2002thermodynamic} developed for matrices in
	\cite{krzakala2016mutual} and tensors in \cite{MiolaneTensor}.

	Consider the interpolating Hamiltonian depending of $t\in[0,1]$
	\begin{equation}\label{eq:interpolating Hamiltonian}
		\begin{split}
			\mathcal{H}_{N,t} &= - \sum_{i<j}\left[\frac{\sqrt{t}}{\Delta_2\sqrt{N}}Y_{ij}x_ix_j +\frac{t}{2\Delta_2N}(x_ix_j)^2\right] +\\
			& - \sum_{i_1<\dots<i_p}\left[\frac{\sqrt{t(p-1)!}}{\Delta_pN^{\frac{p-1}2}}T_{i_1\dots i_p}x_{i_1}\dots x_{i_p} + \frac{t(p-1)!}{2\Delta_pN^{p-1}}(x_{i_1}\dots x_{i_p})^2\right] + \\
			& - \sum_j\left[\sqrt{1-t}\sqrt{\frac{m^{p-1}}{\Delta_p}+\frac{m}{\Delta_2}}W_jx_j+(1-t)\left(\frac{m^{p-1}}{\Delta_p}+\frac{m}{\Delta_2}\right)x_j^*x_j+\frac{1-t}2\left(\frac{m^{p-1}}{\Delta_p}+\frac{m}{\Delta_2}\right)x_j^2\right]\,,
		\end{split}
	\end{equation}
	where we have for $t=1$ the regular Hamiltonian and for $t=0$
	the first term of Eq.~(\ref{eq:free entropy generic}) where $W_j$ are
	i.i.d. canonical Gaussian variables.  More importantly, for all
	$t\in[0,1]$ we can show that the Hamiltonian above can be seen as the one emerging for an appropriate inference problem, so that
	the Nishimori property is kept valid  for generic $t\in [0,1]$ \cite{krzakala2016mutual}.

	Given the interpolating Hamiltonian we can write the corresponding
	Gibbs measure,
	\begin{equation}
		P(x|W,Y,T) = \frac1{\mathcal{Z}_{N,t}}P_X(x)e^{H_{N,t}(x)}\, ,
	\end{equation}
	and the interpolating free entropy
	\begin{equation}
		\psi_N(t) \doteq
		\frac1N\mathbb{E}\left[\log\mathcal{Z}_{N,t}\right]\, ,
	\end{equation}
	whose boundaries are $\psi_N(1) = \frac1N\mathcal{F}_N$ (our target) and
	$\psi_N(0) = \frac1N\tilde\Phi_\text{RS} + \frac1{4\Delta_2}m^2 +
		\frac{p-1}{2p\Delta_p}m^p$.
	We then use the fundamental theorem of calculus to write
	\begin{equation}
		\mathcal{F}_N = \psi_N(1) = \psi_N(0) +\frac1N\underbrace{\mathbb{E}\int_0^1\left(-\frac{\partial\log\mathcal{Z}_{N,t}}{\partial t}\right)dt}_{\doteq \mathcal R}\,.
	\end{equation}
	We work with the second term and use Stein's lemma which, given a
	well behaving function $g$, provides the useful relation for a
	canonical Gaussian variable $Z$: $\mathbb{E}_Z[Zg(Z)] =
		\mathbb{E}_Z[g'(Z)]$. This yields
	\begin{equation*}
		\begin{split}
			\mathcal{R} & = -\mathbb{E}\int_0^1\left[\frac1{\mathcal{Z}_{N,t}}\int dx^N \frac{\partial \mathcal{H}_{N,t}(x)}{\partial t}P_X(x)e^{\mathcal{H}_{N,t}(x)} \right]dt = -\mathbb{E}\int_0^1\left\langle \frac{\partial \mathcal{H}_{N,t}(x)}{\partial t} \right\rangle dt \\
			& = -\mathbb{E}\int_0^1\left\langle \sum_{i<j}\frac1{\Delta_2N}(x_i^*x_ix_j^*x_j) + \sum_{i_1<\dots<i_p}\frac{(p-1)!}{\Delta_2N^{p-1}}(x_{i_1}^*x_{i_1}\dots x_{i_p}^*x_{i_p}) - \sum_i\left(\frac{m}{2\Delta_2}+\frac{m^{p-1}}{2\Delta_p}\right)x_i^*x_i \right\rangle dt \\
			& =
			\mathbb{E}\int_0^1\left[\frac1{4\Delta_2}\left\langle
				\left(\frac{x\cdot x^*}N\right)^2 - 2
				m\left(\frac{x\cdot x^*}N\right)\right\rangle +
				\frac1{2p\Delta_p}\left\langle \left(\frac{x\cdot
					x^*}N\right)^p - p m^{p-1}\left(\frac{x\cdot
					x^*}N\right)\right\rangle\right] dt\,.
		\end{split}
	\end{equation*}
	where we have used the Nishimori property to replace terms such as
	$\langle x\rangle^2$ by $\langle x x^* \rangle$. At this point, we can
	write
	\begin{eqnarray}
		\mathcal{R}&=&
		\mathbb{E}\int_0^1\left[\frac1{4\Delta_2}\left\langle
			\left(\frac{x\cdot x^*}N\right)^2 - 2
			m\left(\frac{x\cdot x^*}N\right)\right\rangle \right]dt
		+\mathbb{E}\int_0^1\left[
			\frac1{2p\Delta_p}\left\langle \left(\frac{x\cdot
				x^*}N\right)^p - p m^{p-1}\left(\frac{x\cdot
				x^*}N\right)\right\rangle\right]
		dt\, \nonumber  \\
		&=&-\frac {m^2}{4\Delta_2} + \frac {1}{4\Delta_2}
		\mathbb{E}\int_0^1\frac1{4\Delta_2}\left\langle
		\left(\frac{x\cdot x^*}N-m\right)^2 \right\rangle dt
		+ \frac1{2p\Delta_p}\mathbb{E}\int_0^1\
		\left\langle \left(\frac{x\cdot
			x^*}N\right)^p - p m^{p-1}\left(\frac{x\cdot
			x^*}N\right)\right\rangle
		dt\,. \nonumber \\
	\end{eqnarray}
	The first integral is clearly positive. The second one,
	however, seems harder to estimate. We may, however, use a simple
	convexity argument on the function $f(x) = x^k$. Indeed observe that
	$\forall\;a,b\ge0$ and $p\ge1$: $a^p-p b^{p-1}a \ge (1-p)b^p$. We would like
	to use this property but there is the subtlety that we need $x\cdot x^*$ to be
	non-negative. To bypass this problem we can add again a small
	perturbation that forces $x \cdot x^*$ to concentrate around a
	non-negative value, without affecting the ``interpolating free
	entropy'' $\psi_N(t)$ in the $N \to \infty$ limit. This is, again, the
	argument used in \cite{MiolaneTensor} and originally in
	\cite{korada2009exact}. In this way we can write
	\begin{eqnarray}
		\mathcal{R} &\ge& -\frac {m^2}{4\Delta_2}+ \mathbb{E}\int_0^1\left[\frac1{4\Delta_2}\left\langle
			\left(\frac{x\cdot x^*}N\right)^2 - 2
			m\left(\frac{x\cdot x^*}N\right)\right\rangle \right]dt +
		\frac{(1-p)m^p}{4\Delta_2}         \, \nonumber
		\\
		&\ge& -\frac {m^2}{4\Delta_2} - \frac{(p-1)m^p}{4\Delta_2}       \:.
	\end{eqnarray}
	This concludes the proof and yields the lower bound:
	\begin{equation}
		\mathcal{F}_N \ge \psi_N(0) - \frac1{4\Delta_2}m^2 -
		\frac{p-1}{2p\Delta_p}m^p   =
		\frac1N\tilde\Phi_\text{RS}(m)   \, ,
	\end{equation}
	so that for all $m\ge 0$
	\begin{align*}
		\liminf_{N \to \infty} {\cal F}_N = \liminf_{N \to \infty}
		\psi_N(1) & = \liminf_{N \to \infty} \left[\psi_N(0) +
			\int_0^1 \psi'_N(t) dt \right]  \geq \tilde \Phi_{\rm RS}(m) \,.
	\end{align*}

	\paragraph{Upper bound: Aizenman-Sims-Starr scheme.} The matching
	upper bound is obtained using the Aizenman-Sims-Starr
	scheme \cite{aizenman2003extended} . This is a particularly effective
	tool that has been already used for these problems, see for example
	\cite{MiolaneXX,coja2017information,MiolaneTensor}. The method goes as follows. Consider the
	original system with $N$ variables, $\mathcal{H}_N$ and add an new variable
	$x_0$ so that we get an Hamiltonian $\mathcal{H}_{N+1}$.
	Define the Gibbs measures of the two systems, the first with $N$ variables and the second with $N+1$ variables, and consider the two relative free entropies.
	Call
	$A_N = \mathbb{E}\left[\log\mathcal{Z}_{N+1}\right] -
		\mathbb{E}\left[\log\mathcal{Z}_{N}\right]$ their difference.  First, we notice that we
	have $\lim\sup_N\mathcal{F}_N \le \lim\sup_NA_N$ because
	$$
		{\cal  F}_N=\mathbb{E} \frac 1N \log {Z_N} = \frac 1N \mathbb{E} \log  \left(\frac{Z_N}{Z_{N-1}} \frac{Z_{N-1}}{Z_{N-2}} \ldots\frac{Z_1}{Z_0} \right)= \frac 1N \sum_i A_i \le \sup_i A_i\, .
	$$
	Moreover, we can separate the contribution of the additional variable in the Hamiltonian ${\cal H}_{N+1}$ so that
	$\mathcal{H}_{N+1} = \tilde{\mathcal{H}}_N + x_0 z(x) + x_0^2 s(x)$,
	with $x = (x_1,\dots,x_N)$, and
	\begin{align*}
		 & z(x) = \frac1{\sqrt{\Delta_2(N+1)}}\sum_{i=1}^N Z_{0i}x_i + \frac{\sqrt{(p-1)!}}{\sqrt{\Delta_p}(N+1)^{(p-1)/2}}\sum_{1\le i_1<\dots<i_{p-1}\le N}Z_{0i_1\dots i_{p-1}}x_{i_1}\dots x_{i_{p-1}} + \\
		 & \quad + \frac1{\Delta_2(N+1)}\sum_{i=1}^N x_0^*x_i^*x_i +\frac{(p-1)!}{\Delta_p(N+1)^{p-1}}\sum_{1\le i_1<\dots<i_{p-1}\le N} x_0^*x_{i_1}^*x_{i_1}\dots x_{i_{p-1}}^*x_{i_{p-1}}
		\\
		 & s(x) = - \frac1{2\Delta_2(N+1)}\sum_{i=1}^N x_i^2 - \frac{(p-1)!}{2\Delta_p(N+1)^{p-1}}\sum_{1\le i_1<\dots<i_{p-1}\le N}(x_{i_1}\dots x_{i_{p-1}})^2
	\end{align*}
	and $\mathcal{H}_{N+1}$ is the same expression as Eq.~(\ref{eq:Hamiltonian}) where the $N$ in the denominators are replaced by $N+1$. We rewrite also $\mathcal{H}_N(x)$ as a perturbation of $\tilde{\mathcal{H}}_N$: $\mathcal{H}_N(x) = \tilde{\mathcal{H}}_N(x) + y(x) + O(1)$ with

	\begin{align*}
		 & y(x) = \frac1{\sqrt{\Delta_2N}}\sum_{i<j}V_{ij}x_ix_j + \sqrt{p-1}\frac{\sqrt{(p-1)!}}{\sqrt{\Delta_p}N^{p/2}}\sum_{i_1<\dots<i_p}V_{i_1\dots i_p}x_{i_1}\dots x_{i_p} +                                    \\
		 & \quad + \frac1{N^2}\sum_{i<j}\left[x_i^*x_ix_j^*x_j-\frac12(x_ix_j)^2\right] +(p-1)!\frac{p-1}{N^p}\sum_{i_1<\dots<i_p}\left[x_{i_1}^*x_{i_1}\dots x_{i_p}^*x_{i_p}-\frac12(x_{i_1}\dots x_{i_p})^2\right],
	\end{align*}
	where the $Z$s and the $V$s are standard Gaussian random variables.

	Finally we can observe the partition functions $Z_N$ can be interpreted
	as ensemble averages with respect to $\tilde{\mathcal{H}}_N$. Thus
	$A_N = \mathbb{E}\log\left\langle\int P_X(x_0)e^{x_0 z(x)+x_0^2
				s(x)}dx_0 \right\rangle_{\tilde{\mathcal{H}}_N} -
		\mathbb{E}\log\left\langle
		e^{y(x)}\right\rangle_{\tilde{\mathcal{H}}_N}$. Now, using the
	Nishimori property and the concentration of the overlap around a
	non-negative value ---that we denote $m(Y,T)$ since it depends explicitly on the disorder--- it yields (see \cite{MiolaneXX}, see section 4.3 for
	details) \eqref{eq:free entropy generic} in the thermodynamic limit, with $m(Y,T)$ instead of $m$. From this, we can now obtain the upper bound that concludes the proof:
	\begin{equation}
		\limsup_N \mathcal{F}_N \le \limsup_N A_N \le \limsup_N \mathbb{E}_{Y,T}\tilde\Phi_\text{RS}[m(Y,T)] \le  \limsup_N \sup_{{m}}\Phi_\text{RS}({m}) \le \tilde\Phi_\text{RS}\,.
	\end{equation}

	\subsection{State evolution of AMP and its analysis}\label{subsec:SE}

	\begin{figure}[h]
		\centering
		\includegraphics[scale=.5]{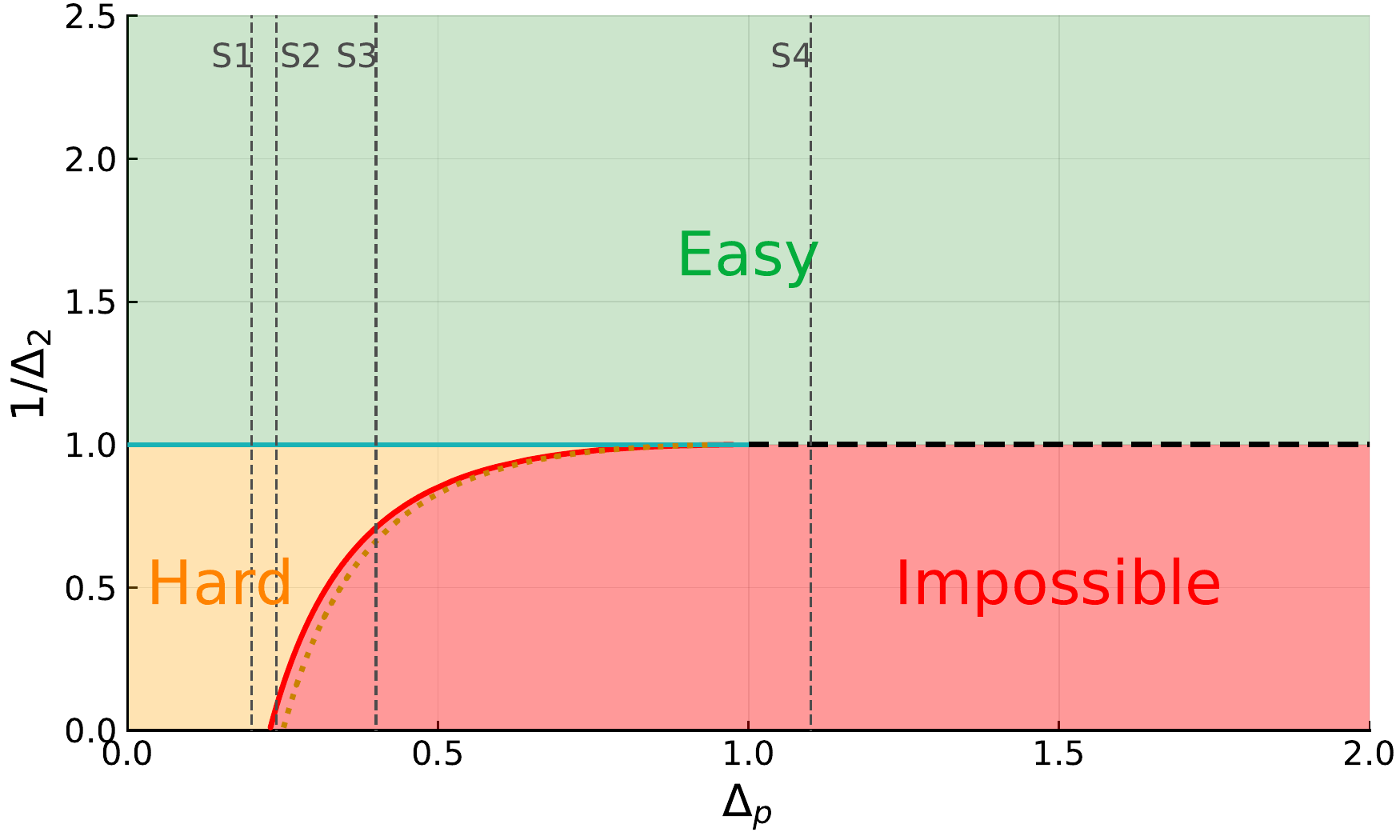}
		\includegraphics[scale=.5]{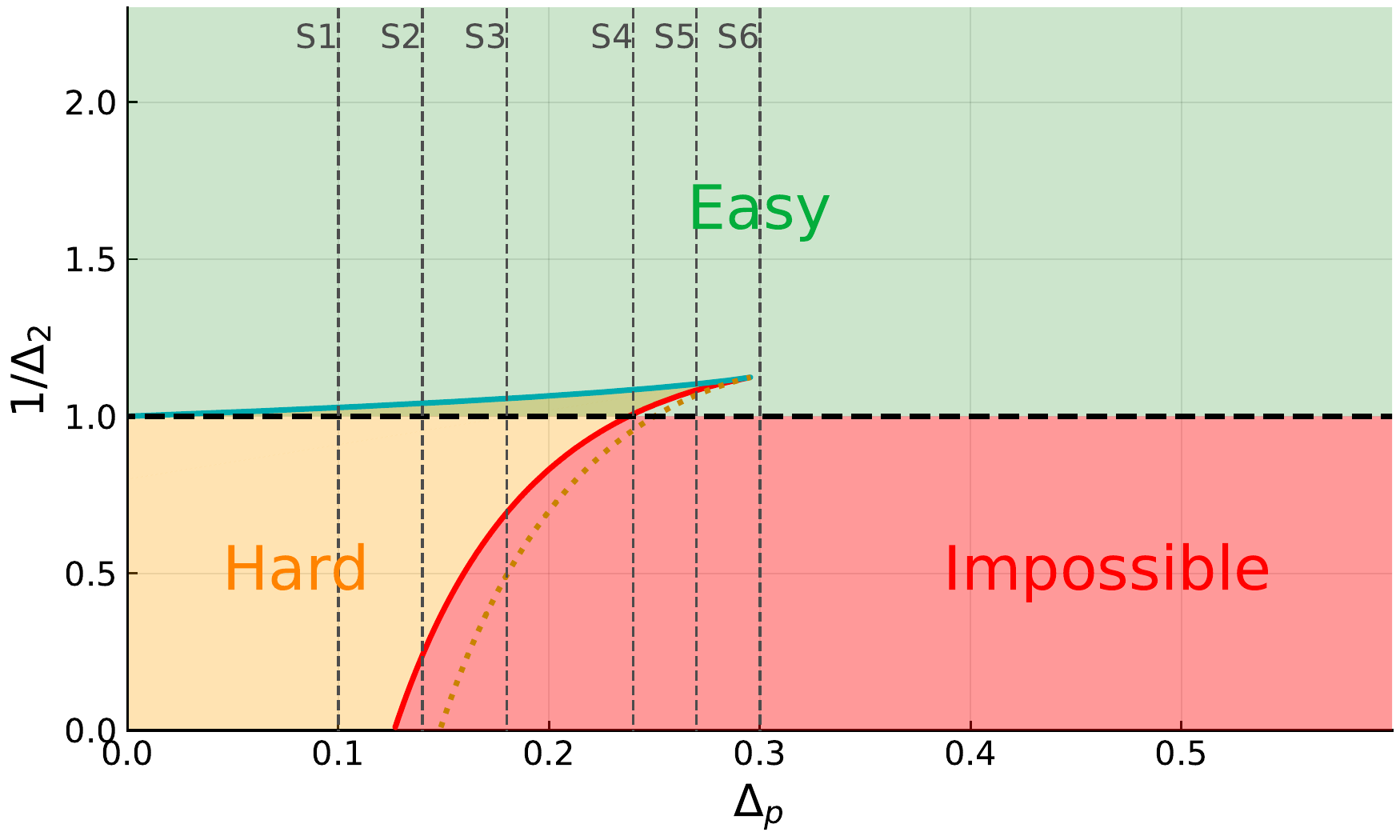}
		\caption{\emph{On the left}: Phase diagram of the spiked matrix-tensor model for
			$p=3$. The phase diagram identifies four regions: easy (green),
			impossible (red), and hard (orange). The lines correspond to
			different phase transitions namely the stability threshold (dashed black),
			the information theoretic threshold (solid red), the algorithmic
			threshold (solid cyan), and the dynamical threshold (dotted
			orange). The vertical cuts represent the section along which the magnetization is plotted in Fig.~\ref{fig:phase diagram p=3 sections}.
			\emph{On the right}: Phase diagram of the spiked matrix-tensor model for
			$p=4$. The main difference with respect to case $p=3$,
			is that the algorithmic
			spinodal (solid cyan) is strictly above the stability threshold
			(dashed black). The hybrid-hard phase appears between these two
			lines (combined green and orange color). The vertical cuts represent the section along which the magnetization is plotted in Fig.~\ref{fig:phase diagram p=4 sections}.
		}
		\label{fig:phase diagram p=3}
	\end{figure}

	The dynamical evolution of the AMP algorithm in the large $N$ limit is described by the so-called State Evolution (SE) equations.
	The derivation of these equations can be straightforwardly
        done using the same techniques as developed in \cite{LKZ17}.
	They can be written in terms of two dynamical order parameters namely $m^t = \sum_i
		\hat{x}_i^tx_i^*/N$, which encodes for the alignment of the current estimation $\hat x_i^t$ of the components of the signal with the signal itself at
                time $t$ and $q^t = \sum_i
                \hat{x}_i^t\hat{x}_i^t/N$. Keeping the spherical
                constraint in mind we obtain the following SE
                equations 
                \begin{eqnarray}
		\frac{m^{t+1}}{1-q^{t+1}}&=& 
		\frac{m^t}{\Delta_2}+\frac{(m^t)^{p-1}}{\Delta_p} \label{eq:AMP_SE_m}
                \;\;, \\		
\frac{q^{t+1}}{(1-q^{t+1})^2}&=& 
		\left[  \frac{m^t}{\Delta_2}+\frac{(m^t)^{p-1}}{\Delta_p}
                \right]^2 + 	\left[  \frac{q^t}{\Delta_2}+\frac{(q^t)^{p-1}}{\Delta_p}
                \right] \;\; .    \label{eq:AMP_SE_q}
              \end{eqnarray}
        Note that eq. (\ref{eq:AMP_SE_m}) at fixed values of $q$
        describes the evolution of the parameter $m$, this is why we use in
        the main text to derive the Langevin threshold eq.~(\ref{eq:threshold}).               
	Finally, using the Nishimori symmetry it can be shown that $m^t = q^t$ at all times, see e.g. \cite{REVIEWFLOANDLENKA}, and therefore the evolution
	of the algorithm is characterized by a single order parameter
	$m^t$ whose dynamical evolution is given by
	\begin{equation}
		m^{t+1}=
		1-\frac1{1+\frac{m^t}{\Delta_2}+\frac{(m^t)^{p-1}}{\Delta_p}}\;\;.
		\label{eq:AMP SE}
              \end{equation}
              Note that AMP satisfies the Nishimori property at all times while this condition is violated on the run by the Langevin dynamics. In that case the Nishimori symmetry is recovered only when equilibrium is reached and therefore it is violated when the Langevin algorithm gets trapped in the glass phase, see below.
	If we initialize the configuration of the estimator $\hat x$ at random, the initial value of $m$ will be equal to zero on average.
	However, finite size fluctuations produce by chance a small bias towards the signal and therefore we consider the initialization to be $m^{t=0}=\epsilon$
	being $\epsilon$ an arbitrarily small positive number.
	We will call $m_{\rm AMP}$ the fixed point of Eq.~(\ref{eq:AMP SE}) reached
	from this infinitesimal initialization.
	The
	mean-square-error (MSE) reached by AMP after convergence is then given by ${\rm MSE}_{\rm AMP} = 1-m_{\rm
		AMP}$.

	We underline that Eq.~(\ref{eq:AMP SE})  can be proven
	rigorously following \cite{javanmard2013state,richard2014statistical}.
	Finally we note that the fixed point of the SE satisfies the very same Eq.~(\ref{eq:extrem}) that
	gives the replica
	free entropy.
	In the rest of this section we will study the fixed points of Eq.~(\ref{eq:AMP SE}). This will allow too determine the phase diagram of the spiked
	matrix-tensor model.

	We start by observing that $m=0$ is a fixed point of Eq.~(\ref{eq:AMP SE}).
	However, in order to understand whether it is a possible attractor of the AMP dynamics
	we need to understand its local stability. This can be obtained perturbatively by
	expanding Eq.~(\ref{eq:AMP SE}) around $m=0$
	\begin{equation}
		m^{t+1}
		=\frac{m^t}{\Delta_2}+\left(\frac{m^t}{\Delta_2}\right)^2-\frac{(m^t)^{p-1}}{\Delta_p}+O\left((m^t)^3\right)\;.
	\end{equation}
	It is clear that the non-informative fixed point $m=0$ is stable as long as
	$\Delta_2>1$. We will call $\Delta_2=1$ the \emph{stability threshold}.

	\begin{figure}
		\centering
		\includegraphics[scale=.5]{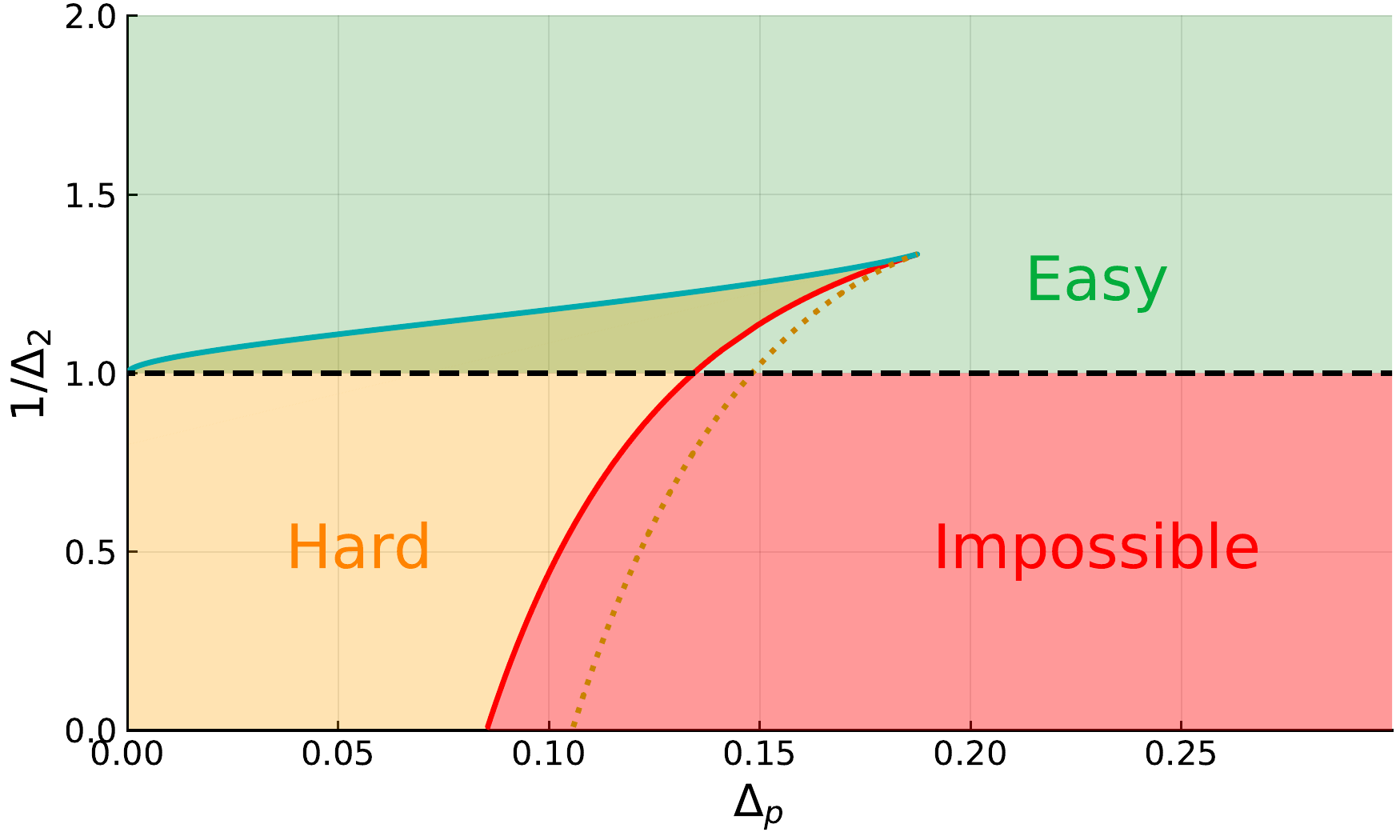}
		\includegraphics[scale=.5]{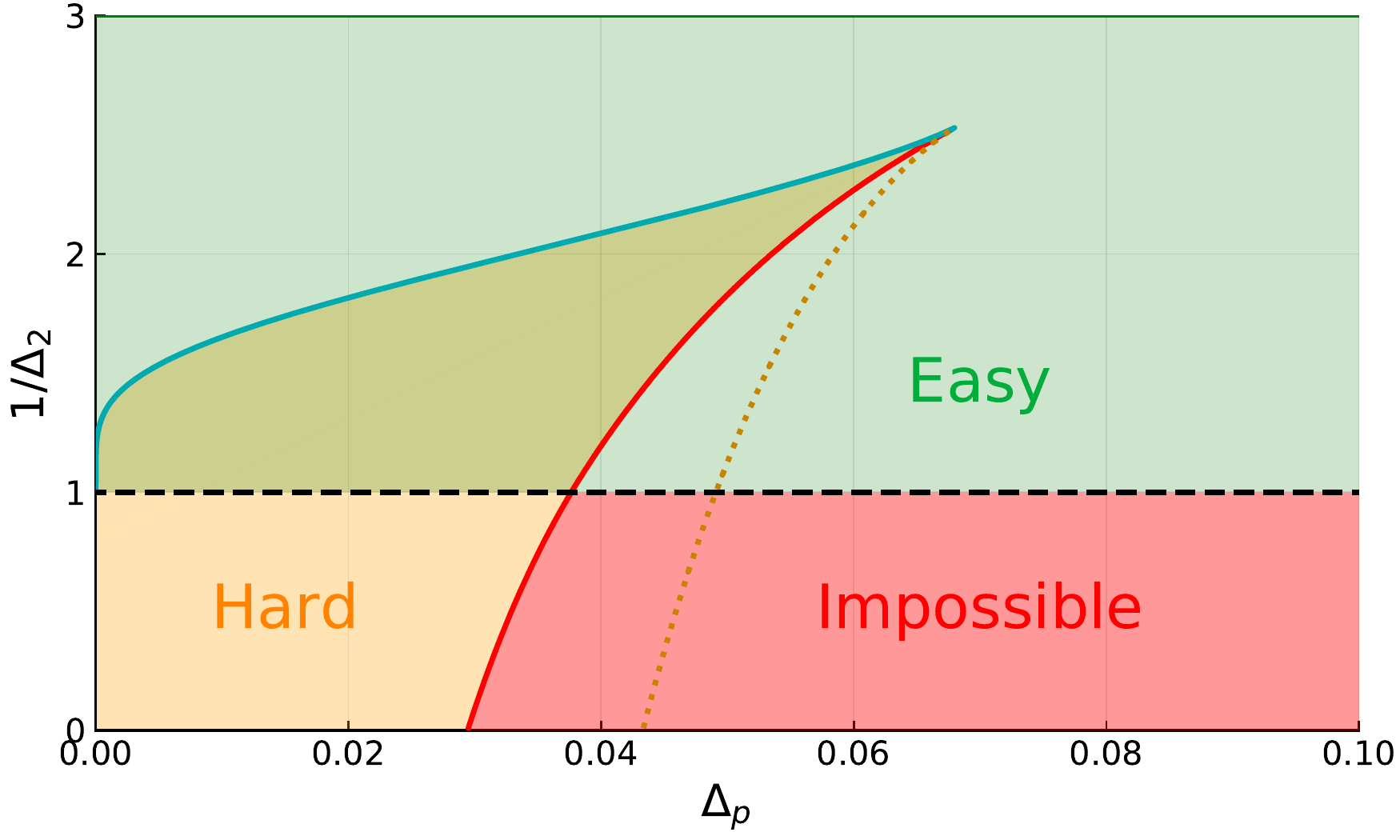}

		\caption{\emph{On the left}: Phase diagram of the spiked matrix-tensor model for
			$p=5$.
			\emph{On the right}: Phase diagram of the spiked matrix-tensor model for
			$p=10$.
			In both cases we observe qualitatively the same scenario found in the right panel of Fig.~\ref{fig:phase diagram p=3}.
		}
		\label{fig:phase diagram p=5}
	\end{figure}

	When $p=3$ the SE equations are particularly simple
	and the fixed points are written explicitly as
	\begin{equation}
		m_0=0\,;\quad \quad
		m_\pm=\frac12\left[1-\frac{\Delta_3}{\Delta_2} \pm
			\sqrt{\left(1+\frac{\Delta_3}{\Delta_2}\right)^2-4\Delta_3}\right]\;.
	\end{equation}
	In the regime where $\Delta_2>1$, $m_0$ and $m_+$ are stable while
	$m_-$ in unstable. When $\Delta_2$ becomes smaller than one, $m_+$
	becomes the only non-negative stable solution and therefore $\Delta_2=1$ is also known as the \emph{algorithmic spinodal} since it corresponds to the point
	where the AMP algorithm converges to the informative fixed point.
	The informative solution $m_+$ exists as long as $\Delta_2  \le   \Delta^{\rm dyn}_2 $, where we have defined the
	\emph{dynamical spinodal} by
	\begin{equation}
		\Delta^{\rm dyn}_2  = \frac{\Delta_3}{2 \sqrt{\Delta_3} -1} \, .
	\end{equation}

	For a generic $p$ we cannot determine the values of the informative
	fixed points explicitly but we can easily study Eq.~(\ref{eq:AMP SE}) numerically to get the full phase diagram.

	Furthermore we can obtain the spinodal transition lines as follows.
	The key
	observation is that the two spinodals are critical points of the
	equation $\Delta_p(m;\Delta_2)$ where $\Delta_2$ is fixed, or
	analogously $\Delta_2(m;\Delta_p)$ where $\Delta_p$ is fixed (to have
	a pictorial representation of the idea you can see Fig.~\ref{fig:phase
		diagram p=4 sections}). We call $x = m/\Delta_2 + m^{p-1}/\Delta_p$,
	and $f_\text{SE}(x)\equiv 1 - \frac1{1+x}$, then
	\begin{equation}
		\Delta_p\equiv\Delta_p(x;\Delta_2)=\frac{\left(f_\text{SE}(x)\right)^{p-1}}{x-\frac{f_\text{SE}(x)}{\Delta_2}}\;.
	\end{equation}
	Then the stationary points are implicitly defined by
	\begin{equation*}
		0=\frac{d\log\Delta_p}{d m}=\frac{\partial\log\Delta_p}{\partial x}(1+x)^2\propto (p-1)\frac{f_\text{SE}'(x)}{f_\text{SE}(x)}-\frac{1-\frac{f_\text{SE}'(x)}{\Delta_2}}{x-\frac{f_\text{SE}'(x)}{\Delta_2}}=\frac{\frac{2-p}{\Delta_2}+(1+x)(p-x-2)}{x(1+x)\left[x+1-\frac1{\Delta_p}\right]}\;,
	\end{equation*}
	giving
	\begin{equation}\label{eq:spinodals x}
		x_\pm(\Delta_2)=\frac12\left[p-3\pm\sqrt{(p-1)^2-\frac4{\Delta_2}(p-2)}\right]\;.
	\end{equation}
	Finally $\Delta_p\left(x_\pm(\Delta_2);\Delta_2\right)$ describes the
	two spinodals.
	We can also derive the tri-critical point, when the two spinodals meet, which is given by the zero discriminant condition on \eqref{eq:spinodals x}
	\begin{equation}
		\left(\Delta_p^{\rm tri};1/\Delta_2^{\rm tri}\right)=\left(\frac{4(p-2)\left(\frac{p-3}{p-1}\right)^{p-1}}{(p-3)^2};\frac{(p-1)^2}{4(p-2)}\right)\;.
	\end{equation}

	\subsection{Phase diagrams of spiked matrix-tensor model}

	In this section we present the phase diagrams for the spiked
	matrix-tensor model as a function of the two noise levels $\Delta_2$ and $\Delta_p$ and for several
	values of $p$.
	These phase diagrams are plotted in Figs.~\ref{fig:phase diagram p=3} and \ref{fig:phase diagram p=5}.

	\noindent Generically we can have four regions:
	\begin{itemize}
		\item {\bf Easy phase} (green), where the MSE obtained through AMP coincides with the MMSE which is better than random sampling of the prior.
		\item {\bf Impossible phase} (red), where the MMSE and MSE of AMP coincide and are equal to 1 (meaning that $m^*=m_{AMP}$=0).
		\item {\bf Hard phase} (orange),  where the MMSE is smaller than the MSE obtained from AMP and $m^*>m_{\rm AMP} \ge 0$.
		\item {\bf Hybrid-hard phase} \cite{Typology18} (mix of green and orange),
		      is a part of the hard phase where the AMP
		      performance is strictly better than random sampling from the prior,
		      but still the MSE obtained this way does not match the MMSE, i.e. $m^*>m_{\rm AMP} > 0$.  The hybrid-hard phase can be found
		      for $p \ge 4$.
	\end{itemize}

	\noindent All these phases are separated by the following transition lines:
	\begin{itemize}
		\item The {\bf stability threshold} (dashed black line) at $\Delta_2=1$ for all $p$. This corresponds to the point where the uninformative fixed
		      point $m=0$ looses its local stability.
		\item The {\bf information theoretic
		      threshold} (solid red line).  Here $m^*>0$ and the MMSE jumps to a value strictly smaller than one.
		\item The {\bf algorithmic threshold} (solid cyan line). This is
		      where the fixed point of AMP jumps to the MMSE$<1$.
		      For $p=3$ this line coincides with a segment
		      of the stability threshold while for $p\ge 4$ it is
		      strictly above.
		\item The {\bf dynamic
		      threshold} (dotted orange line). Here the most informative fixed point (the one with largest $m_{\rm AMP}$) disappears.
	\end{itemize}

	In Figs.~\ref{fig:phase diagram p=3 sections}, and \ref{fig:phase diagram p=4 sections} we plot the
	evolution of the magnetization $m$, as found through the fixed points of the SE equation, for several fixed values of $\Delta_p$ and $p=3$ and $p=4$,
	respectively. The values of $\Delta_p$ are identified by the vertical cuts in the phase diagrams of Fig.~\ref{fig:phase diagram p=3}.

	\begin{figure}[H]
		\centering
		\subfigure[Section S1: $p=3$, $\Delta_p=0.2$]{
		\includegraphics[scale=0.4]{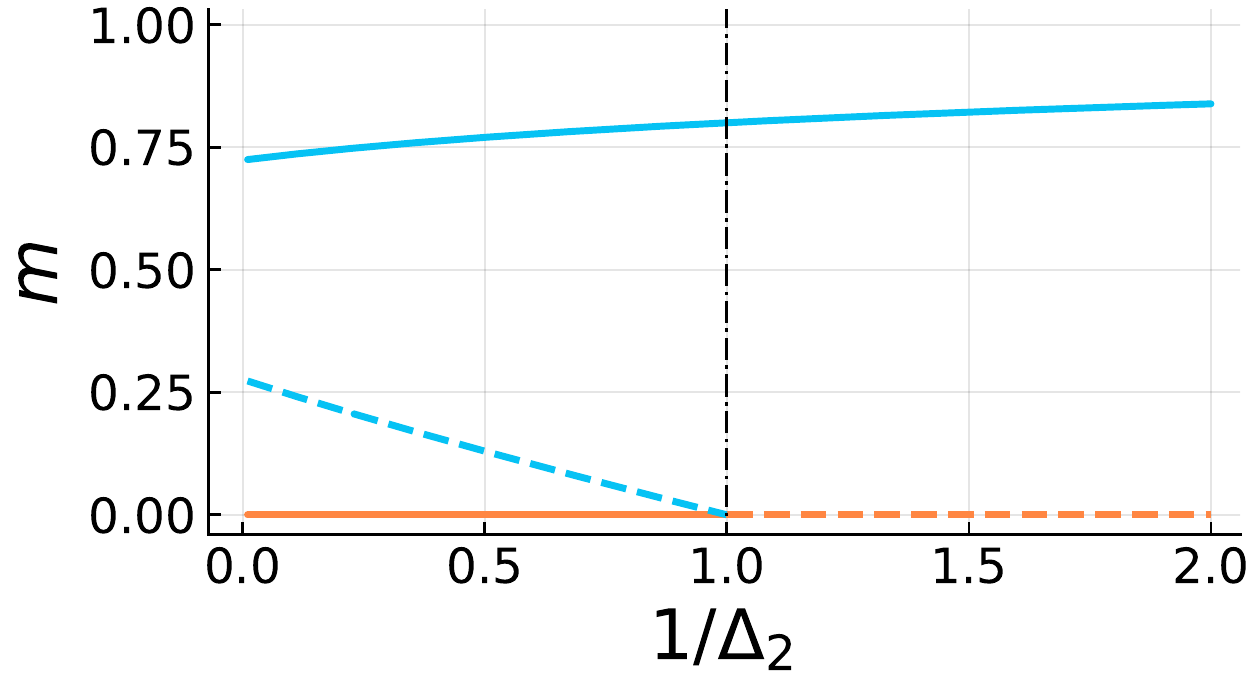}
		}
		\subfigure[Section S2:$p=3$, $\Delta_p=0.24$]{
		\includegraphics[scale=0.4]{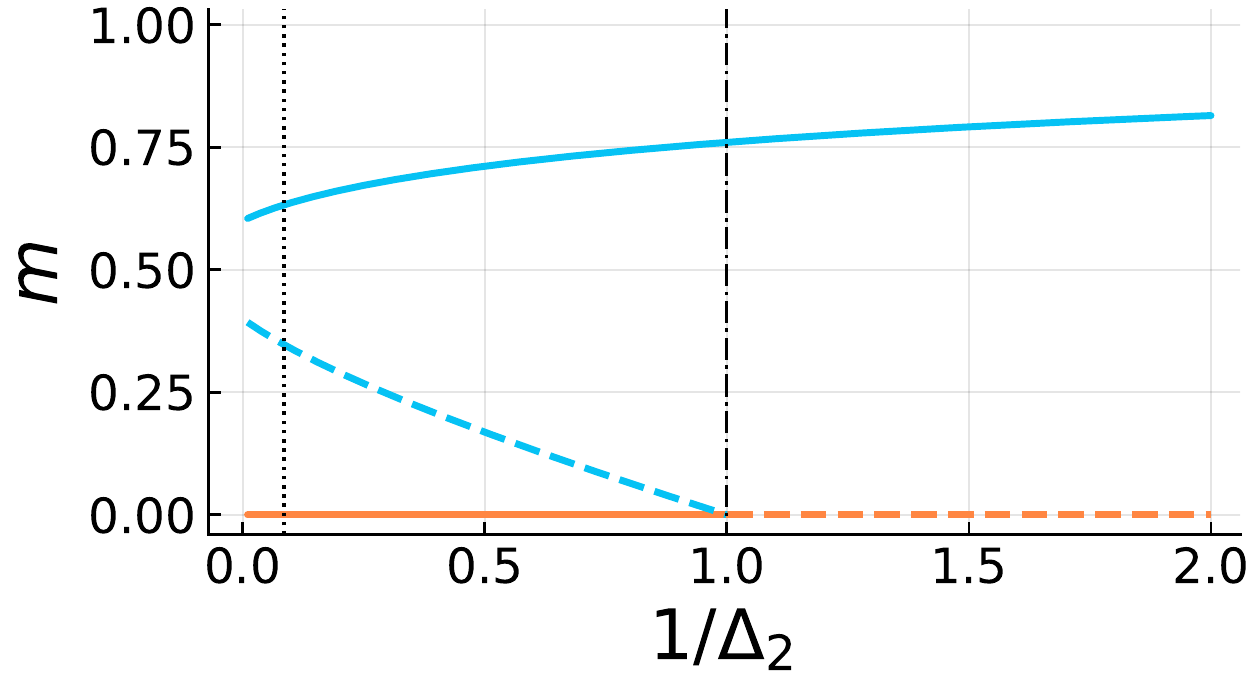}
		}
		\\
		\subfigure[Section S3: $p=3$, $\Delta_p=0.4$]{
		\includegraphics[scale=0.4]{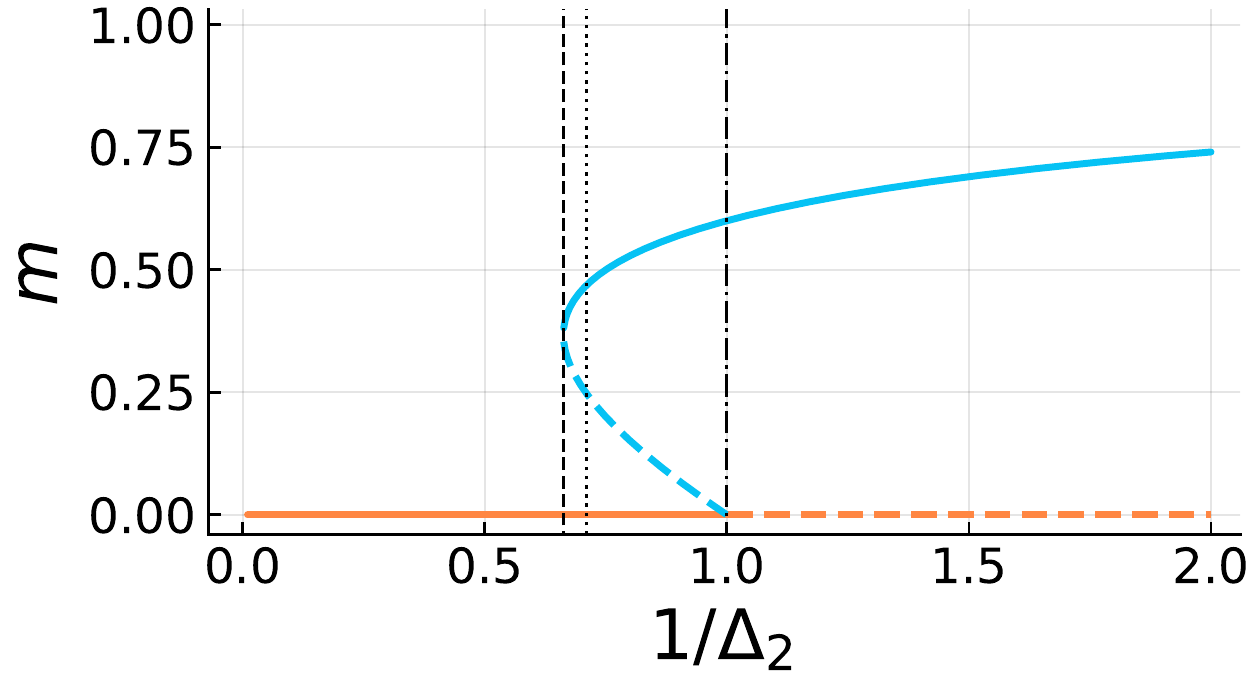}
		}
		\subfigure[Section S4: $p=3$, $\Delta_p=1.1$]{
		\includegraphics[scale=0.4]{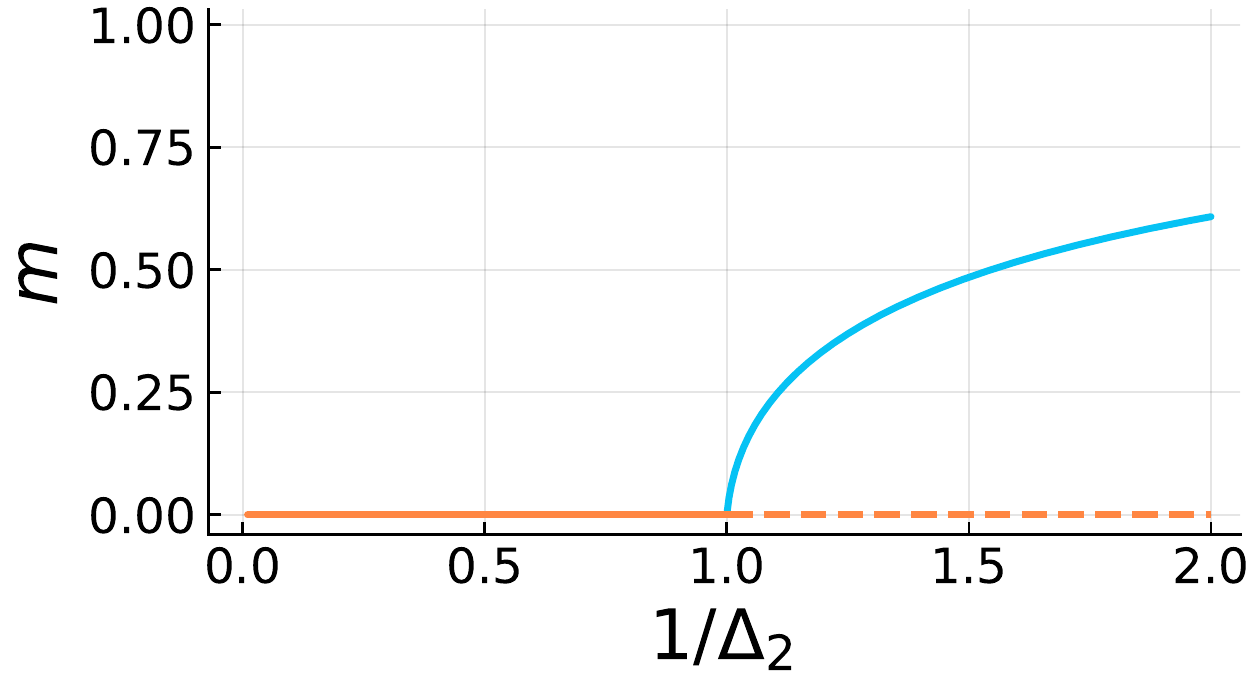}
		}
		\caption{Fixed points of Eq.~(\ref{eq:AMP SE}) as
			a function of $\Delta_2$
			for $p=3$ and several fixed values of $\Delta_p$.
			The values of $\Delta_p$ correspond to the vertical cuts in the left panel of Fig.~\ref{fig:phase diagram p=3}. Solid
			lines are stable fixed point, dashed lines are unstable fixed
			points.  The blue line represent informative fixed points with positive
			overlap with the signal while the orange line represent a uninformative fixed points
			with no overlap with the signal.
			Starting from high $\Delta_2$ an informative fixed point appears at
			the dynamical threshold (vertical dashed line) but is energetically
			disfavored until the information theoretic threshold (vertical dotted
			line) and finally it becomes the only stable solution crossing the
			algorithmic threshold (vertical dotted-dashed line). When the
			transition is continuous the three vertical threshold lines merge and
			we have a single second order phase transition, which here occurs at
			$\Delta_p \ge 1$.}
		\label{fig:phase diagram p=3 sections}
	\end{figure}

	\begin{figure}[H]
		\centering
		\subfigure[Section S1: $p=4$, $\Delta_p=0.1$]{
		\includegraphics[scale=0.4]{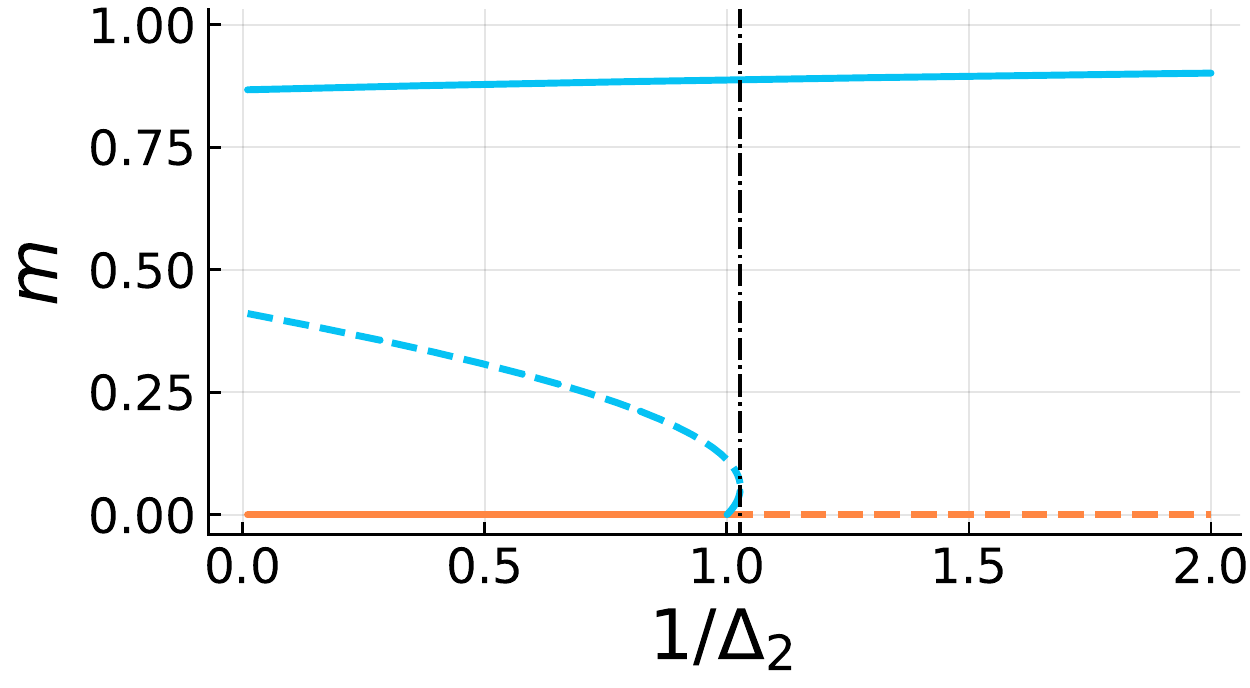}
		}
		\subfigure[Section S2: $p=4$, $\Delta_p=0.14$]{
		\includegraphics[scale=0.4]{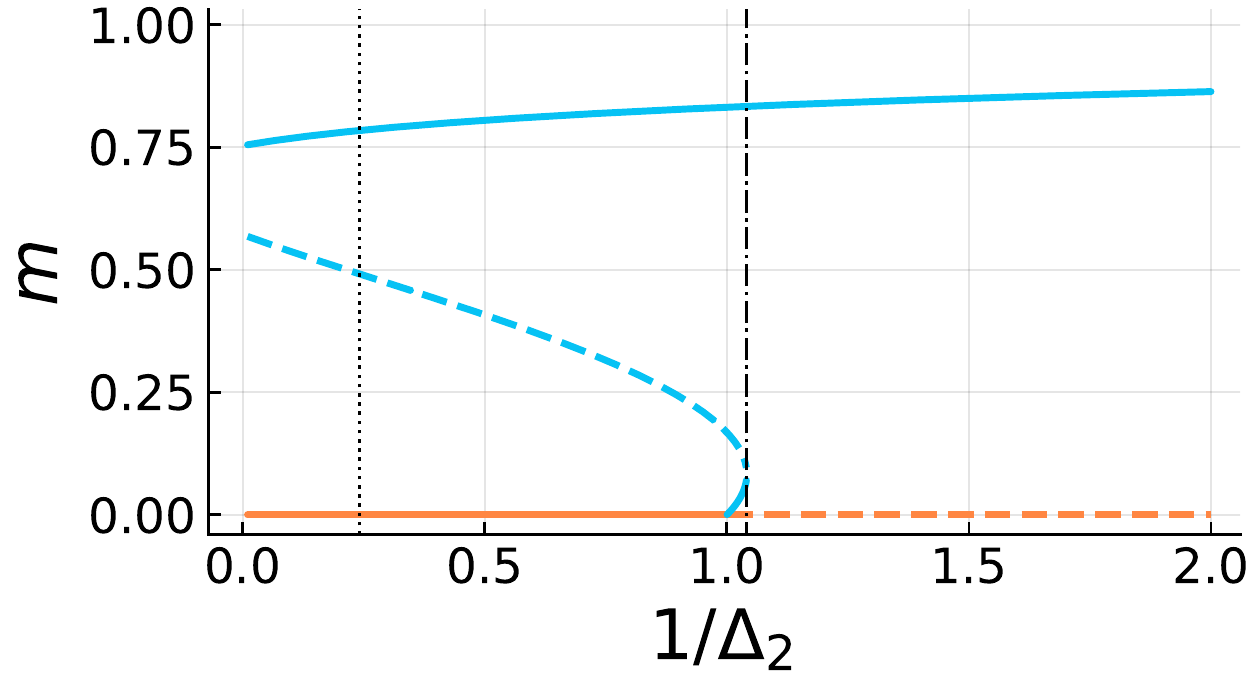}
		}
		\centering
		\subfigure[Section S3: $p=4$, $\Delta_p=0.18$]{
		\includegraphics[scale=0.4]{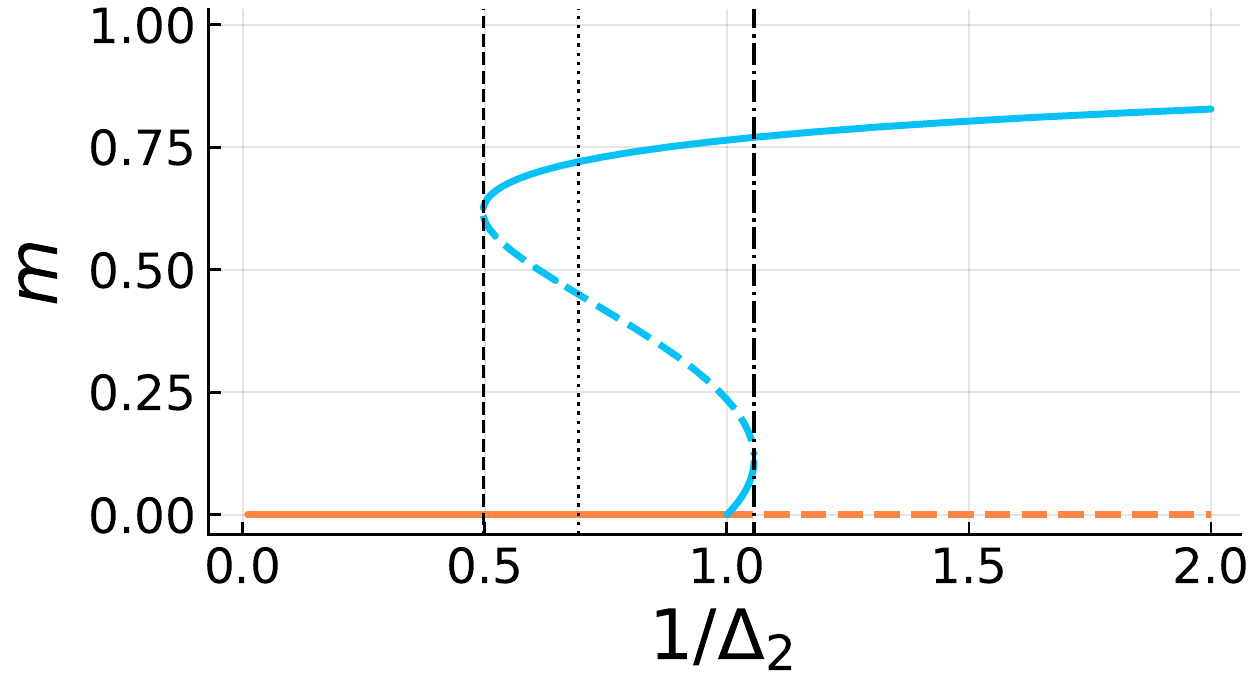}
		}
		\subfigure[Section S4: $p=4$, $\Delta_p=0.24$]{
		\includegraphics[scale=0.4]{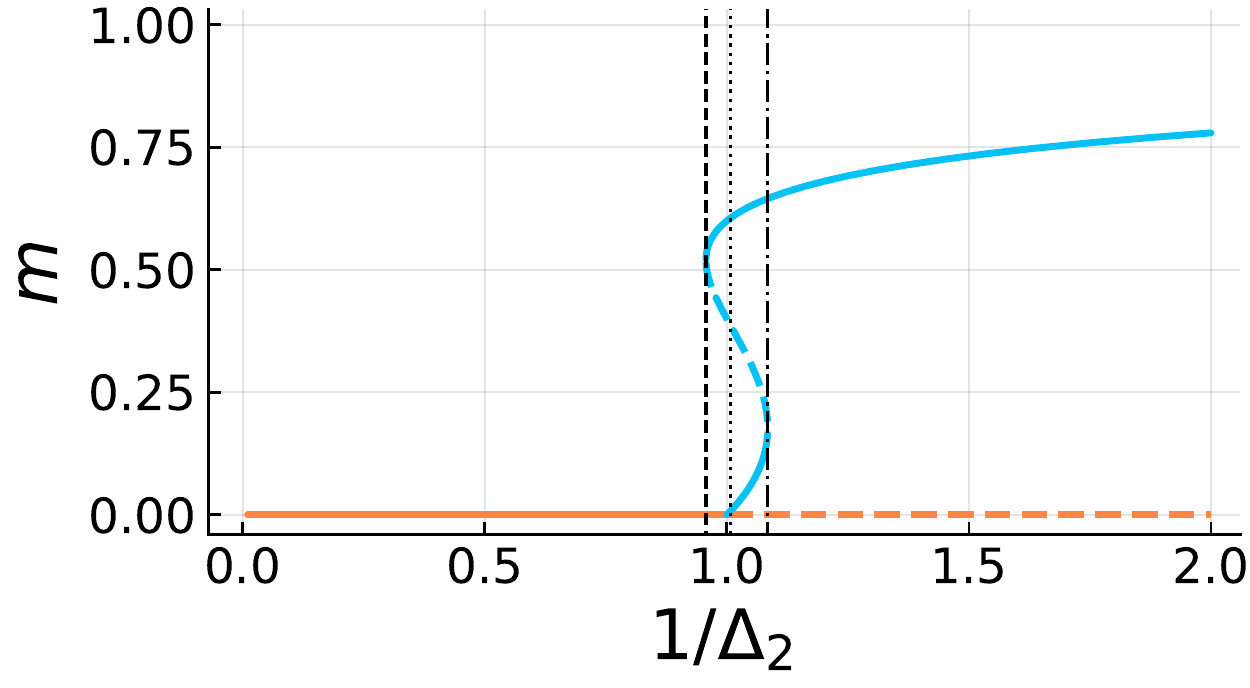}
		}
		\centering
		\subfigure[Section S5: $p=4$, $\Delta_p=0.27$]{
		\includegraphics[scale=0.4]{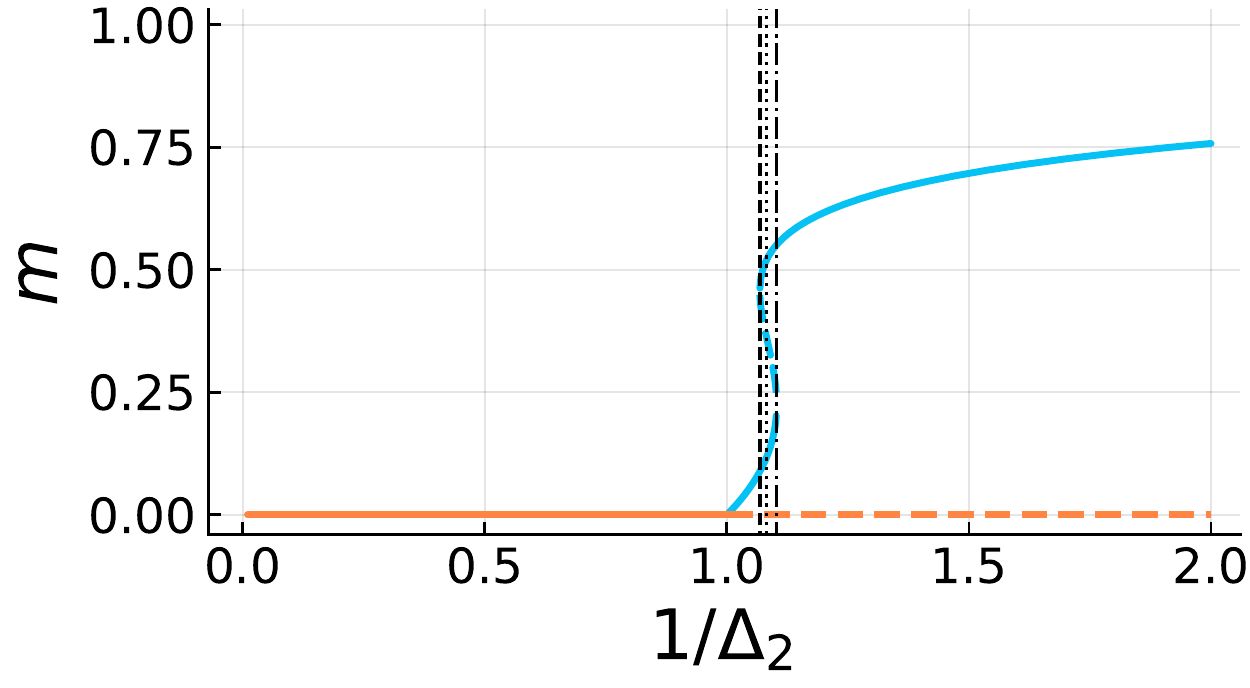}
		}
		\subfigure[Section S6: $p=4$, $\Delta_p=0.3$]{
		\includegraphics[scale=0.4]{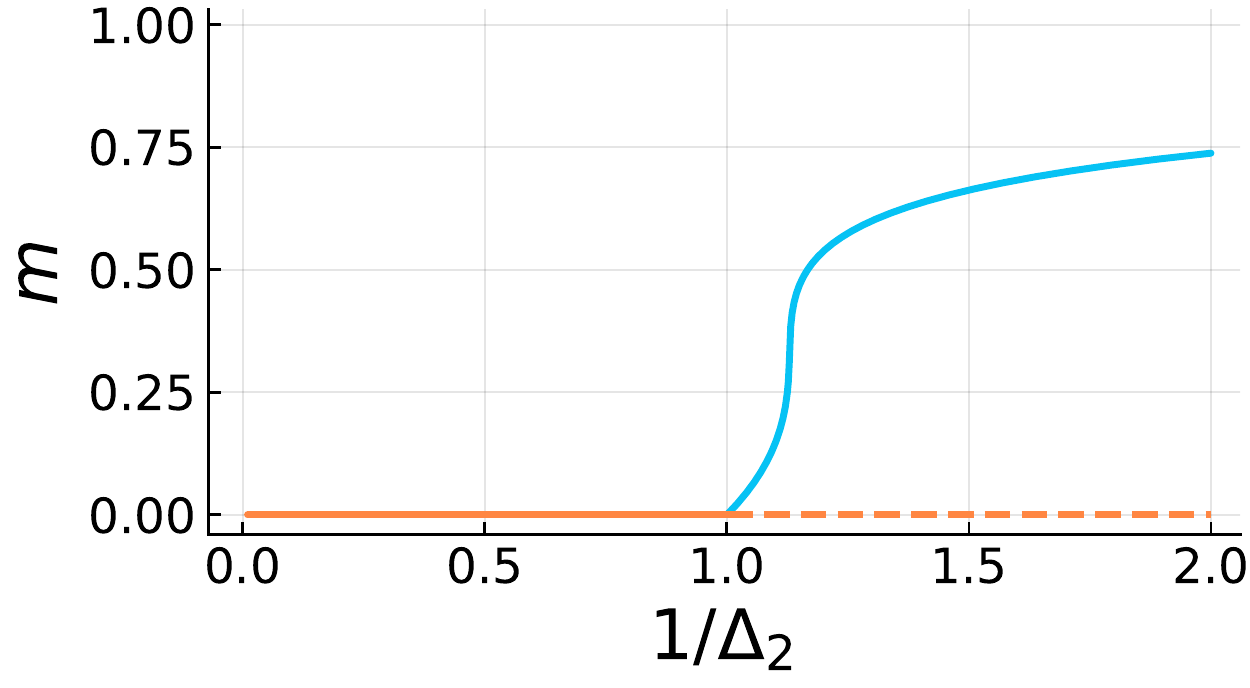}
		}
		\caption{Fixed points Eq.~(\ref{eq:AMP SE}) as
			a function of $\Delta_2$
			for $p=4$ and several fixed values of $\Delta_p$. The values of $\Delta_p$ correspond to the vertical cuts of the right panel of Fig.~\ref{fig:phase diagram p=3}.
			The situation is qualitatively similar to Fig.~\ref{fig:phase diagram p=3 sections}, the difference being only the presence of the hybrid-hard phase.
			We can observe that when the transition is discontinuous, figure from (a) to (e), for $1/\Delta_2>1.0$ the uninformative solution becomes unstable and continuously goes to a stable-informative solution which is not the optimal one.}
		\label{fig:phase diagram p=4 sections}
	\end{figure}

	\section{Langevin Algorithm and its state evolution}\label{sec:dynamics}

	The main goal of our analysis is to compare AMP with the performance of the Langevin dynamics. The advantage of the spiked matrix-tensor model is that in this case the Langevin dynamics can be studied in the large $N$ limit through integro-differential equations for the correlation function, $C(t,t')=\lim_{N\to \infty} \sum_i\langle x_i(t)x_i(t')\rangle/N$, the response function $R(t,t')=\lim_{N\to \infty} \frac1N\sum_i\frac{d\langle x_i(t)\rangle}{d\eta_i(t')}$ and the magnetization  $\Cmag(t)=\lim_{N\to \infty} \sum_i\langle x_i(t)x_i^*\rangle /N$.

	To obtain these equations we use the techniques developed in the context of mean-field spin glass systems \cite{MPV87, Cu03}.
	We call $\eta_i(t)$ a time dependent noise and we indicate with $\langle \cdot \rangle$ the average with respect to it. The noise is Gaussian and characterized by $\langle \eta_i(t)\rangle=0$ for all $t$ and $i=1,\ldots N$ and $\langle \eta_i(t) \eta_j(t')\rangle = 2\delta_{ij} \delta(t-t')$. As before we will denote by $\mathbb{E}[\dots]$ the average with respect to the realization of disorder that in this case goes back to the specific realization of the signal.

	Before proceeding, it is useful to introduce a set of auxiliary variables that will help in the following.
	For  $k\in\{2,p\}$ we define $r_k \equiv r_k(t) = 2/\left(kT_k(t)\Delta_k\right)$, $f_k(x)=x^k/2$ and $m(t)\doteq\frac1N\sum_ix_i(t)x_i^*$, and the random variable $\tilde{\xi}_{i_1\dots i_k}\equiv \frac1{\Delta_k}\xi_{i_1\dots i_k} \sim \mathcal{N}(0,1/\Delta_k)$. The time dependence in $T_k$, will be used in the tensor-annealing protocol that will be used to avoid part of the Langevin hard phase.
	We introduce a time dependent Hamiltonian
	\begin{equation*}
		\begin{split}
			\mathcal{H}(t) &= -\frac1{T_2(t)\sqrt{N}}\sum_{i<j}\tilde{\xi}_{ij}x_i(t)x_j(t) -\frac{\sqrt{(p-1)}!}{T_p(t)N^{\frac{p-1}2}}\sum_{i_1<\dots<i_p}\tilde{\xi}_{i_1\dots i_p}x_{i_1}(t)\dots x_{i_p}(t) \\
			&- Nr_2(t)f_2(m(t)) - Nr_p(t)f_p(m(t))\,,
		\end{split} \label{time_hamiltonian}
	\end{equation*}
	and the associated Langevin dynamics
	\begin{equation}\label{lng}
		\begin{split}
			\dot{x}_i(t) &=-\mu(t) x_i(t) -\frac{\partial \mathcal{H}}{\partial x_i}(t) - \eta_i(t) = -\mu(t) x_i(t) -\frac1{T_2(t)\sqrt{N}}\sum_{j(\neq i)}\tilde{\xi}_{ij}x_j(t)+
			\\
			&+ r_2(t)f'_2(m(t))-\frac{\sqrt{(p-1)!}}{T_p(t)N^{\frac{p-1}2}}\sum_{(i,i_1,\dots ,i_{p-1})\setminus i}\tilde{\xi}_{ii_1\dots i_{p-1}}x_{i_1}(t)\cdots x_{i_{p-1}}(t)+r_p(t)f'_p(m(t)) - \eta_i(t)\,,
		\end{split}
	\end{equation}
	with $\mu$ a Langrange multiplier that enforces the spherical constraint $\sum_{i=1}^N x_i^2(t)=N$.
	If $T_k(t)=1$ for all $k=2,\ p$, the stationary equilibrium distribution for the Langevin dynamics is given by the posterior measure.
	Using Ito's lemma one finds
	\[
		\frac 1 N\frac{d}{dt}\sum_i x_i^2(t)=\frac 2 N\sum_i x_i(t)
		\dot{x}_i(t)+2 \, .
	\]
	Since the spherical constraint imposes the left-hand-side to be zero, one obtains a condition on the right-hand-side. By plugging the expression (\ref{lng}) in it, one gets that in the large $N$ limit
	\begin{equation}
		\mu(t) = 1 - 2 \mathcal{H}_2(t) - p \mathcal{H}_p(t)
	\end{equation}
	where
	\begin{equation}
		{\cal H}_k = -\frac{\sqrt{(k-1)}!}{T_k(t)N^{\frac{k-1}2}}\sum_{i_1<\dots<i_k}\tilde{\xi}_{i_1\dots i_k}x_{i_1}(t)\dots x_{i_k}(t)- Nr_k(t)f_k(m(t))\ \ \ \ \ \ k=2,p
	\end{equation}
	are the parts of the Hamiltonian defined in Eq.~(\ref{time_hamiltonian}) relative to the matrix ($k=2$) and to the tensor ($k=p$).

	Note that we have not specified any initial condition for the variables $x_i(t=0)$. Therefore, since we always employ the spherical constraint, the initial condition for the dynamics is a point on the $N$ dimensional hypersphere $|x|^2=N$ extracted with the flat measure.

	In order to analyze the Langevin dynamics in the large $N$ limit, we will use the dynamical cavity method \cite{MPV87, CC05, agoritsas2018out}. We will consider a system of $N$ variables, with $N\gg1$, and add a new one. This new variable will be considered as a small perturbation to the original system but at the same time will be treated self consistently.

	\subsection{Dynamical Mean-Field Equations}\label{SI:DMFT_derivation}
	In the following we will drop the time dependence for simplicity restoring it only when it is needed.
	Given the system with $N$ variables $i=1\dots N$, we add a new one, say $i=0$, and define $\tilde{m}=\frac1{N+1}\sum_{i=0}^Nx_ix_i^*\simeq\frac1N\sum_{i=0}^Nx_ix_i^*$ (henceforth we use the symbol $\simeq$ to denote two quantities that are
	equal up to terms that vanish in the large-$N$ limit).
	The Langevin equation associated to the new variable is
	\begin{equation}
		\begin{split}
			\dot{x}_0 &=-\mu x_0 -\frac1{T_2(t)\sqrt{N}}\sum_{j(\neq 0)}\tilde{\xi}_{0j}x_j+r_2f'_2(\tilde{m})-\frac{\sqrt{(p-1)!}}{T_p(t)N^{\frac{p-1}2}}\sum_{(0,i_1,\dots ,i_{p-1})\setminus 0}\tilde{\xi}_{0i_1\dots i_{p-1}}x_{i_1}\cdots x_{i_{p-1}}+r_pf'_p(\tilde{m}) - \eta_0\,,
		\end{split}
	\end{equation}
	where we used that $N\simeq N+1$ for $N\gg 1$.
	We will consider the contribution of the new variable on the others in perturbation theory.
	In the dynamical equations for the variables $i=1,\ldots, N$ we can isolate the variable $i=0$ and write
	\begin{equation}
		\begin{split}
			\dot{x}_i &= -\mu x_i -\frac1{T_2(t)\sqrt{N}}\sum_{j(\neq i,0)}\tilde{\xi}_{ij}x_j+r_2f'_2(m)-\frac{\sqrt{(p-1)!}}{T_p(t)N^{\frac{p-1}2}}\sum_{(i,i_1,\dots ,i_{p-1})\setminus i,0}\tilde{\xi}_{ii_1\dots i_{p-1}}x_{i_1}\cdots x_{i_{p-1}}\\
			&+r_pf'_p(m) - \eta_i +H_i\,,
		\end{split}
	\end{equation}
	with
	\begin{equation}
		H_i(t)=\left(r_2f_2''(m)+r_pf_p''(m)\right)\frac1Nx_0-\frac1{T_2(t)\sqrt{N}}\tilde{\xi}_{0i}x_0-\frac{\sqrt{(p-1)!}}{T_p(t)N^{\frac{p-1}2}}\sum_{(i,0,i_1,\dots ,i_{p-2})\setminus i,0}\tilde{\xi}_{i0i_1\dots i_{p-2}}x_0x_{i_1}\cdots x_{i_{p-2}}\,.
	\end{equation}
	Consider the unperturbed variables $x_i^0=x_i\big|_{H_i=0}$. At leading order in $N$ we can write
	\begin{equation}
		x_i\simeq x_i^0+\int_{t_o}^tdt'\frac{\delta x_i(t)}{\delta H_i(t')}\bigg|_{H_i=0}H_i(t')\:.
	\end{equation}
	In the dynamical equation for the variable 0 we can identify a piece associated to the unperturbed variables $x_i^0$. This term can be thought of collectively  as a stochastic term $\Xi(t)$
	\begin{equation}\label{eq:i0 spin perturbed}
		\begin{split}
			\dot{x}_0 &=-\mu x_0 \overbrace{-\frac1{T_2(t)\sqrt{N}}\sum_{j(\neq 0)}\tilde{\xi}_{0j}x^0_j-\frac{\sqrt{(p-1)!}}{T_p(t)N^{\frac{p-1}2}}\sum_{(0,i_1,\dots ,i_{p-1})\setminus 0}\tilde{\xi}_{0i_1\dots i_{p-1}}x^0_{i_1}\cdots x^0_{i_{p-1}} - \eta_0}^{\doteq\;\Xi(t)}+\\
			&+r_2f'_2(m)+r_pf'_p(m)+\left(r_2f_2''(m)+r_pf_p''(m)\right)\frac1Nx_0-\frac1{T_2(t)\sqrt{N}}\sum_{j(\neq0)}\tilde{\xi}_{0j}\int_{t_o}^tdt'\frac{\delta x_j(t)}{\delta H_j(t')}\bigg|_{H_j=0}H_j(t')+\\
			&-\Bigg[\frac{\sqrt{(p-1)!}}{T_p(t)N^{\frac{p-1}2}}\sum_{(0,i_1,\dots ,i_{p-1})\setminus 0}\tilde{\xi}_{0i_1\dots i_{p-1}}\int_{t_o}^tdt'\frac{\delta x_{i_1}(t)}{\delta H_{i_1}(t')}\bigg|_{H_{i_1}=0}H_{i_1}(t')x^0_{i_2}\cdots x^0_{i_{p-1}}+\text{permutations}\Bigg]\,.
		\end{split}
	\end{equation}
	Indeed $\Xi(t)$ encodes the effect of a kind of bath made by of the unperturbed variables $i=1,\ldots, N$ to the new one.
	We can show that at leading order in $N$, $\Xi(t)$ is a Gaussian noise with zero mean and variance given by
	\begin{equation*}
		\begin{split}
			\mathbb{E}\langle \Xi(t)\Xi(t')\rangle  &= 2\delta(t-t')-\mathbb{E}\left[ \frac1{T_2(t)T_2(t')N}\sum_{j(\neq0)}\sum_{l(\neq0)}\tilde{\xi}_{0j}\tilde{\xi}_{0l}x_j^0(t)x_l^0(t')\right] +\\
			&-\mathbb{E}\left[ \frac{(p-1)!}{T_p(t)T_p(t')N^{p-1}}\sum_{(0,i_1,\dots ,i_{p-1})\setminus 0}\sum_{(0,j_1,\dots ,j_{p-1})\setminus 0}\tilde{\xi}_{0i_1\dots i_{p-1}}\tilde{\xi}_{0j_1\dots j_{p-1}}x^0_{i_1}\cdots x^0_{i_{p-1}}x^0_{j_1}\cdots x^0_{j_{p-1}}\right]
		\end{split}
	\end{equation*}
	and the second term can be simplified as
	\begin{equation*}
		\begin{split}
			\mathbb{E}\Bigg[ & \frac{(p-1)!}{T_p(t)T_p(t')N^{p-1}} \sum_{(0,i_1,\dots ,i_{p-1})\setminus 0}\sum_{(0,j_1,\dots ,j_{p-1})\setminus 0}\tilde{\xi}_{0i_1\dots i_{p-1}}\tilde{\xi}_{0j_1\dots j_{p-1}}x^0_{i_1}\cdots x^0_{i_{p-1}}x^0_{j_1}\cdots x^0_{j_{p-1}}\Bigg]  =\\
			&\simeq \frac{(p-1)!}{N^{p-1}}\frac1{T_p(t)T_p(t')\Delta_p} \sum_{(0,i_1,\dots ,i_{p-1})\setminus 0}\langle x^0_{i_1}(t)x^0_{i_1}(t')\cdots x^0_{i_{p-1}}(t)x^0_{i_{p-1}}(t')\rangle =\frac1{T_p(t)T_p(t')\Delta_p}C^{p-1}(t,t')\,,
		\end{split}
	\end{equation*}
	where we used $\sum_{(i_1,\dots,i_k)}=\frac1{k!}\sum_{1\le i_1,\dots,i_k\le N}$, we neglected terms sub-leading in $N$, and we used the definition of the dynamical correlation function
	\begin{equation}
		\nonumber
		C(t,t') = \frac 1N \sum_{i=1}^N \langle x_i(t) x_i(t')\rangle \:.
	\end{equation}
	Therefore we have
	\begin{align}
		 & \mathbb{E}\langle \Xi(t)\rangle =0\,;                                                                                            \\
		 & \mathbb{E}\langle \Xi(t)\Xi(t')\rangle =2\delta(t-t')+\frac1{T_2(t)T_2(t')}C(t,t')+\frac1{T_p(t)T_p(t')\Delta_p}C^{p-1}(t,t')\,.
	\end{align}
	Now we can focus of the deterministic term coming from the first order
	perturbation in \eqref{eq:i0 spin perturbed}. Consider just the
	integral for the $p$-body term, the other will be given by setting
	$p=2$
	\begin{equation}\label{eq:i0 deterministic perturbation}
		\begin{split}
			\frac{\sqrt{(p-1)!}}{T_p(t)N^{\frac{p-1}2}}&\sum_{(0,i_1,\dots ,i_{p-1})\setminus 0}\tilde{\xi}_{0i_1\dots i_{p-1}}\int_{t_o}^tdt'\frac{\delta x_{i_1}(t)}{\delta H_{i_1}(t')}\bigg|_{H_{i_1}=0}H_{i_1}(t')x^0_{i_2}\cdots x^0_{i_{p-1}}+\text{permutations}=\\
			&\simeq \frac{(p-1)!}{T_p(t)N^{p-1}}\sum_{(0,i_1,\dots ,i_{p-1})\setminus 0}\tilde{\xi}^2_{0i_1\dots i_{p-1}}\int_{t_o}^tdt'\frac1{T_p(t')}\frac{\delta x_{i_1}(t)}{\delta H_{i_1}(t')}\bigg|_{H_{i_1}=0} x^0_{i_1}(t)x^0_{i_1}(t')\cdots x^0_{i_{p-2}}(t)x^0_{i_{p-2}}(t')x_0(t')+\\
			&+\text{permutations} \simeq -\frac{p-1}{T_p(t)\Delta_p}\int_{t_o}^tdt'\frac1{T_p(t')}R(t,t')C^{p-2}(t,t')x_0(t')
		\end{split}
	\end{equation}
	where we have used the definition of the response function
	\begin{equation}\nonumber
		R(t,t') = \frac{1}{N} \sum_{i=1}^N \left\langle \frac{\delta x_i(t)}{\delta H_i(t')}\right \rangle \:.
	\end{equation}
	Plugging \eqref{eq:i0 deterministic perturbation} into \eqref{eq:i0 spin perturbed} we obtain an effective dynamical equation for the new variable in terms of the correlation and response function
	of the system with $N$ variables
	\begin{equation}\label{eq:i0 spin}
		\begin{split}
			\dot{x}_0(t)&=-\mu(t) x_0(t)+\Xi(t)+r_pf_p'(\Cmag(t))+r_2f_2'(\Cmag(t))+
			\\
			&+\frac{p-1}{T_p(t)\Delta_p}\int_{t_o}^tdt''\frac1{T_p(t'')}R(t,t'')C^{p-2}(t,t'')x_0(t'')+\frac1{T_2(t)\Delta_2}\int_{t_o}^tdt''\frac1{T_p(t'')}R(t,t'')x_0(t'')\,.
		\end{split}
	\end{equation}
	In order to close Eq.~(\ref{eq:i0 spin}) we need to give the recipe to compute the correlation and response function.

	\subsection{Integro-differential equations}
	In order to obtain the final equations for dynamical order parameters we will assume that the new variable $x_0$ is a typical one, namely it has the same statistical nature of all the others.
	Therefore we can assume that
	\begin{equation}
		\begin{split}
			C(t,t')&\doteq \mathbb{E}\langle x_0(t)x_0(t')\rangle\\
			R(t,t')&\doteq \mathbb{E}\left\langle \frac{\delta x_0(t)}{\delta \Xi(t')}\right\rangle\\
			\Cmag(t)&\doteq \mathbb{E}\langle x_0(t)x_0^*\rangle\:.
		\end{split}
		\label{closure}
	\end{equation}

	Eqs.~(\ref{closure}) give a way to obtain the equation for all the correlation functions.
	Indeed we can consider Eq.~(\ref{eq:i0 spin}), multiply it by $x_0(t')$, or differentiate it with respect to an external field $h_0(t')$,  or multiply it it by $x_0^*$ and we can average the results over the disorder and thermal noise.
	Using the following identity
	\begin{equation}\label{eq:Xi x0 covariance}
		\begin{split}
			\mathbb{E}\langle \Xi(t) & x_0(t')\rangle =\int\mathcal{D}\Xi(t)\;\Xi(t)x_0(t')e^{-\int d\bar{t}d\tilde{t}\Xi(\bar{t})\mathbb{K}^{-1}(\bar{t},\tilde{t})\Xi(\tilde{t})}=\\
			&=-\int dt'' \int\mathcal{D}\Xi(t)\;x_0(t')\frac{\delta}{\delta\Xi(t'')}e^{-\int d\bar{t}d\tilde{t}\Xi(\bar{t})\mathbb{K}^{-1}(\bar{t},\tilde{t})\Xi(\tilde{t})} \mathbb{K}(t,t'')=\\
			&=\int dt''\mathbb{E}\left\langle \frac{\delta x_0(t')}{\delta \Xi(t'')}\mathbb{K}(t,t'')\right\rangle =\int dt''R(t',t'')\mathbb{K}(t,t'')=\\
			&=2R(t',t)+\frac1{T_p(t)\Delta_p}\int_{t_o}^{t'}dt''\frac1{T_p(t'')}R(t',t'')C^{p-1}(t,t'')+\frac1{T_2(t)\Delta_2}\int_{t_o}^{t'}dt''\frac1{T_2(t'')}R(t',t'')C(t,t'')
		\end{split}
	\end{equation}
	we get the following Langevin State Evolution (LSE) equations
	\begin{align}
		\begin{split}
			\label{eq:dynamic correlation}
			\frac{\partial}{\partial t}C(t,t')&= \mathbb{E}\langle \dot{x}_0(t)x_0(t')\rangle = 2R(t',t)-\mu(t)C(t,t')+r_p(t)f_p'(\Cmag(t))\Cmag(t')+r_2(t)f_2'(\Cmag(t))\Cmag(t')+\\
			&+(p-1)\frac1{T_p(t)\Delta_p}\int_{t_o}^tdt''\frac1{T_p(t'')}R(t,t'')C^{p-2}(t,t'')C(t',t'')+\\
			&+\frac1{T_p(t)\Delta_p}\int_{t_o}^{t'}dt''\frac1{T_p(t'')}R(t',t'')C^{p-1}(t,t'')+\\
			&+\frac1{T_2(t)\Delta_2}\int_{t_o}^tdt''\frac1{T_2(t'')}R(t,t'')C(t',t'')+\frac1{T_2(t)\Delta_2}\int_{t_o}^{t'}dt''\frac1{T_2(t'')}R(t',t'')C(t,t'')\,;
		\end{split} \\
		\begin{split}
			\label{eq:dynamic response function}
			\frac{\partial}{\partial t}R(t,t')&=\mathbb{E}\left\langle \frac{\delta \dot{x}_0(t)}{\delta\Xi(t')}\right\rangle =\\
			&=\delta(t-t')-\mu(t)R(t,t')+(p-1)\frac1{T_p(t)\Delta_p}\int_{t'}^tdt''\frac1{T_p(t'')}R(t,t'')R(t'',t')C^{p-2}(t,t'')+\\
			&+\frac1{T_2(t)\Delta_2}\int_{t'}^tdt''\frac1{T_2(t'')}R(t,t'')R(t'',t')\,;
		\end{split} \\
		\begin{split}
			\label{eq:dynamic magnetization}
			\frac{\partial}{\partial t}\Cmag(t)&=\mathbb{E}\langle \dot{x}_0(t)x^*_0\rangle =
			\\
			&=-\mu(t)\Cmag(t)+r_p(t)f_p'(\Cmag(t))+r_2(t)f_2'(\Cmag(t))+\\
			&+(p-1)\frac1{T_p(t)\Delta_p}\int_{t_o}^tdt''\frac1{T_p(t'')}R(t,t'')C^{p-2}(t,t'')\Cmag(t'')+\frac1{T_2(t)\Delta_2}\int_{t_o}^tdt''\frac1{T_2(t'')}R(t,t'')\Cmag(t'')\,;
		\end{split} \\
		\label{eq:dynamic mu}
		\begin{split}
			\mu(t)&=1+r_p(t)f_p'(\Cmag(t))\Cmag(t)+r_2(t)f_2'(\Cmag(t))\Cmag(t)+\\
			&+p\frac1{T_p(t)\Delta_p}\int_{t_o}^tdt''\frac1{T_p(t'')}R(t,t'')C^{p-1}(t,t'')+2\frac1{T_2(t)\Delta_2}\int_{t_o}^tdt''\frac1{T_2(t'')}R(t,t'')C(t,t'')\,.
		\end{split}
	\end{align}
	Note that the last equation for $\mu(t)$ is obtained by imposing the spherical constraint $C(t,t)=1\;\forall t$ using the fact that $0 = \frac{dC(t,t)}{dt} = \frac{\partial C(t,t')}{\partial t}\Big|_{t'=t} + \frac{\partial C(t',t)}{\partial t}\Big|_{t'=t}$. The boundary conditions of this equations are: $C(t,t)=1$ the spherical constrain, $R(t,t)=0$ which comes from causality in the It\^{o} approach and $R(t,t'\to t^-)=1$.
	The initial condition for $\Cmag(0)=\Cmag_0$ is the overlap with the initial configuration with the true signal.
	If the initial configuration is random, $\Cmag_0=0$ but will have finite size fluctuations, as in the case of AMP.
	Therefore we can think that $\Cmag_0=\epsilon$ being $\epsilon$ an arbitrary small positive number.

	\section{Numerical solution of the LSE equations}
	\label{section:numerical implementation}

	The dynamical equations (\ref{eq:dynamic correlation}-\ref{eq:dynamic response function}-\ref{eq:dynamic magnetization}-\ref{eq:dynamic mu}) were integrated numerically using two schemes:
	\begin{itemize}
		\item fixed time-grid: the derivatives were discretized and integrated according to their causal structure. This method is suited only for short times (up to $500$ time units);
		\item dynamic time-grid: the step size is doubled after a given number of steps and the equations are solved self-consistently for every waiting-time. This is the approach proposed in \cite{kim2001dynamics} and described in Appendix C of \cite{berthier2007spontaneous}. It allows integration up to very large times (up to $10^6$ time units).
	\end{itemize}

	The results of these algorithms are concisely reported in the phase diagram shown in the main paper. In what follows we will present the algorithms and a series of investigations that we carried out to check their stability, we will explain the procedure followed to delimit the Langevin hard region, and we will discuss how we can enter into part of that region by choosing a proper annealing protocol. The codes are available online \cite{LSEcode}.

	\subsection{Fixed time-grid $(2+p)$-spin}\label{sec:numerical implementation - fixed grid}

	In this approach time-derivatives and integrals were discretized using $\frac{\partial}{\partial t}f(t,t') \simeq \frac1{\Delta t}\left[f(t+\Delta t,t')-f(t,t')\right]$, and the trapezoidal rule for integration $\int_0^t f(t)dt \simeq \frac{\Delta t}2 \sum_{l=0}^{t/\Delta t-1} \left[f(l\Delta t)+f((l+1)\Delta t)\right]$.
	For instance we defined a function for computing the update in the the
	response function, \eqref{eq:dynamic response function} as follows
	\begin{equation*}
		\begin{split}
			R(t +\Delta t &,t') = R(t,t') -\Delta t\,\mu(t)R(t,t') + \frac12\frac{\Delta t^2}{\Delta_2} \sum_{l=t'/\Delta t}^{t/\Delta t-1} \left[R(t,l\Delta t)R(l\Delta t,t')+R(t,(l+1)\Delta t)R((l+1)\Delta t,t')\right] +\\
			&+ (p-1)\frac{\Delta t^2}{\Delta_p} \sum_{l=t'/\Delta t}^{t/\Delta t-1}\left[C^{p-2}(t,l\Delta t)R(t,l\Delta t)R(l\Delta t,t')+C^{p-2}(t,(l+1)\Delta t)R(t,(l+1)\Delta t)R((l+1)\Delta t,t')\right]\;.
		\end{split}
	\end{equation*}
	Analogously we defined the other integrators. A simple causal
	integration scheme, being careful with the It\^o prescription, gives the
	pseudo-code below.
	\begin{algorithmic}
		\STATE $C(0,0) \gets1$;  $R(0,0) \gets0$;  $\Cmag(0) \gets \Cmag_0$;
		\FOR {$t \le t_{\max}$}
		\STATE $C(t+\Delta t,t+\Delta t) \gets1$;  $R(t+\Delta t,t+\Delta t) \gets0$;
		\STATE $\mu(t) \gets $ compute\_mu($C$, $R$, $\Cmag$, $t$);
		\STATE $\Cmag(t+\Delta t) \gets$ compute\_mag($\mu$, $C$, $R$, $\Cmag$, $t$);
		\FOR {$t' \le t$}
		\STATE $C(t+\Delta t,t') \gets$ compute\_C($\mu$, $C$, $R$, $\Cmag$, $t$);
		\STATE $R(t+\Delta t,t') \gets$ compute\_R($\mu$, $C$, $R$, $\Cmag$, $t$);
		\ENDFOR
		\STATE $R(t+\Delta t,t) \gets1$;
		\ENDFOR
	\end{algorithmic}

	\subsection{Dynamical time-grid $(2+p)$-spin}

	\begin{figure}[H]
		\centering
		\subfigure[Memory allocation;]{
		\includegraphics[scale=.5]{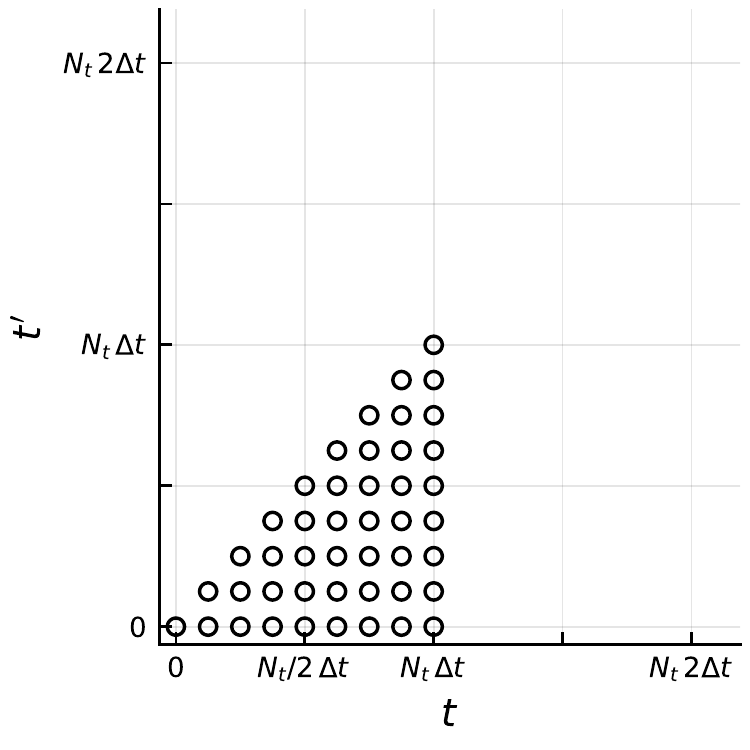}
		}
		\subfigure[step 1;]{
		\includegraphics[scale=.5]{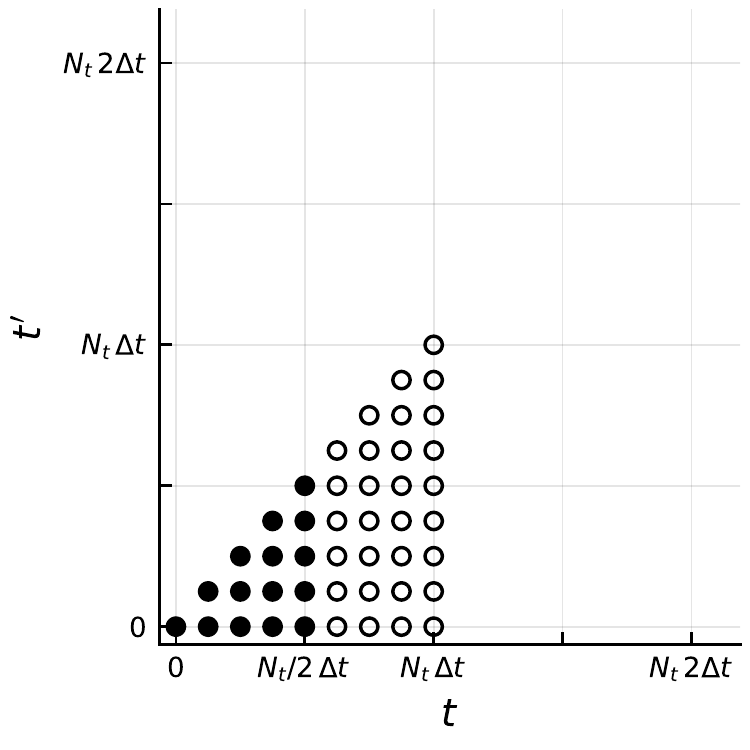}
		}
		\subfigure[step 2;]{
		\includegraphics[scale=.5]{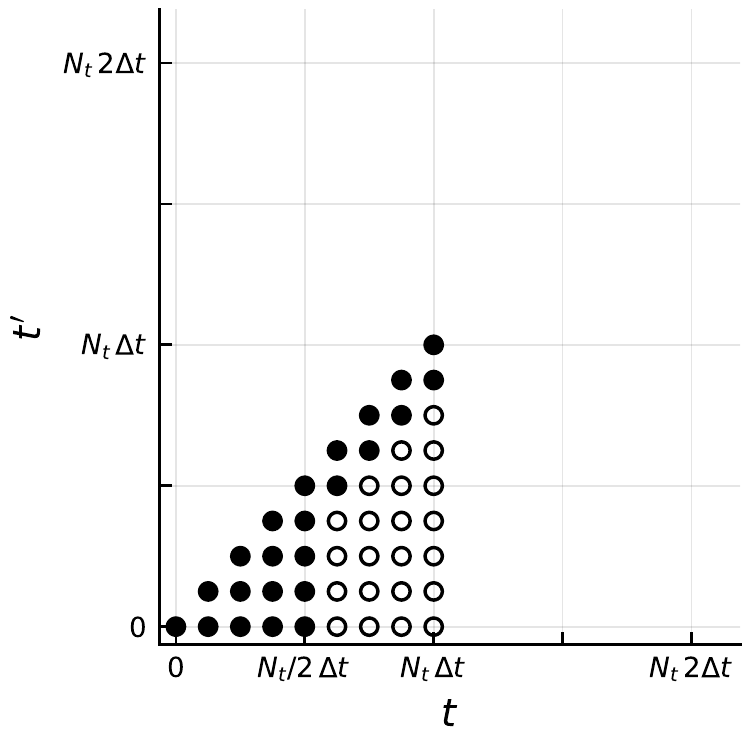}
		}
		\subfigure[step 3;]{
		\includegraphics[scale=.5]{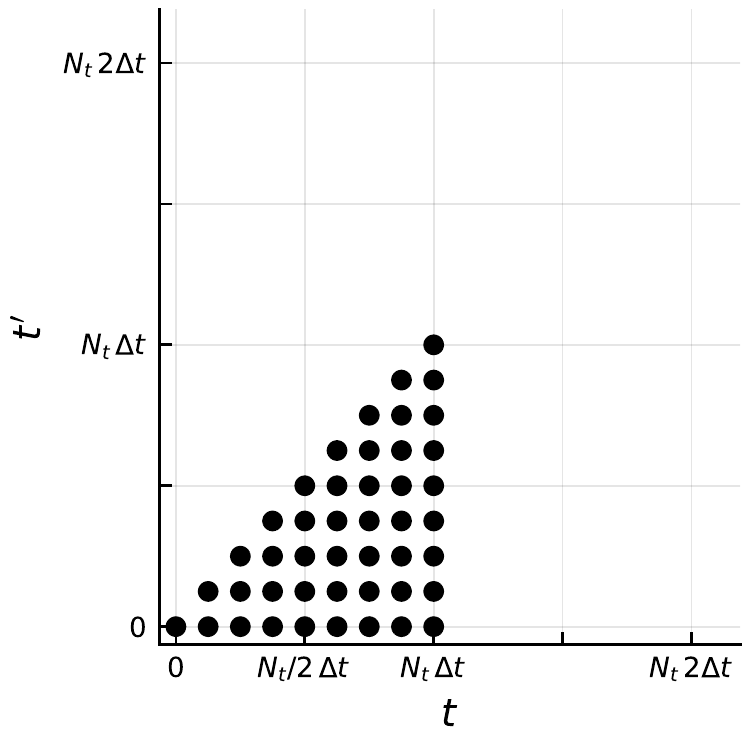}
		}\\
		\centering
		\subfigure[step 4.1;]{
		\includegraphics[scale=.5]{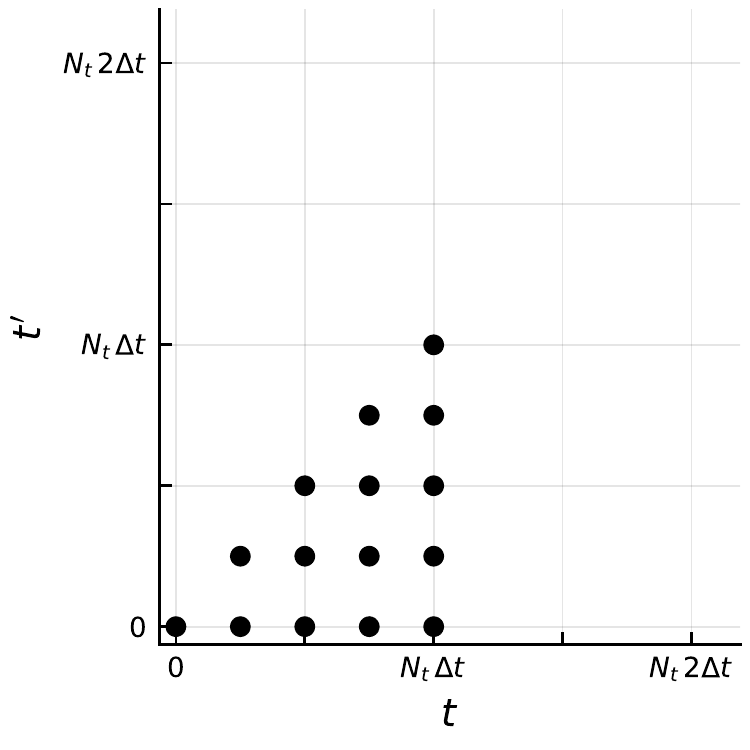}
		}
		\subfigure[step 4.2;]{
		\includegraphics[scale=.5]{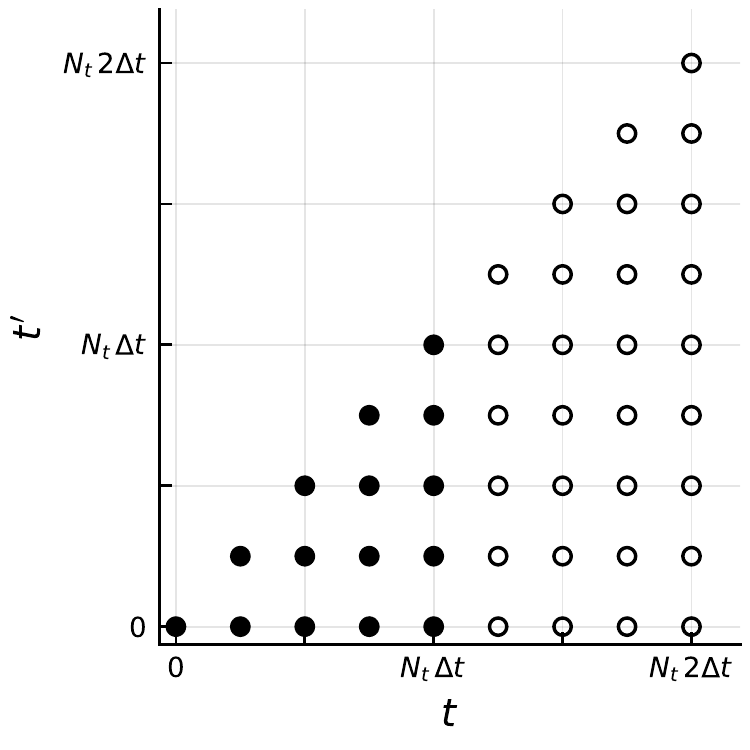}
		}
		\subfigure[step 2;]{
		\includegraphics[scale=.5]{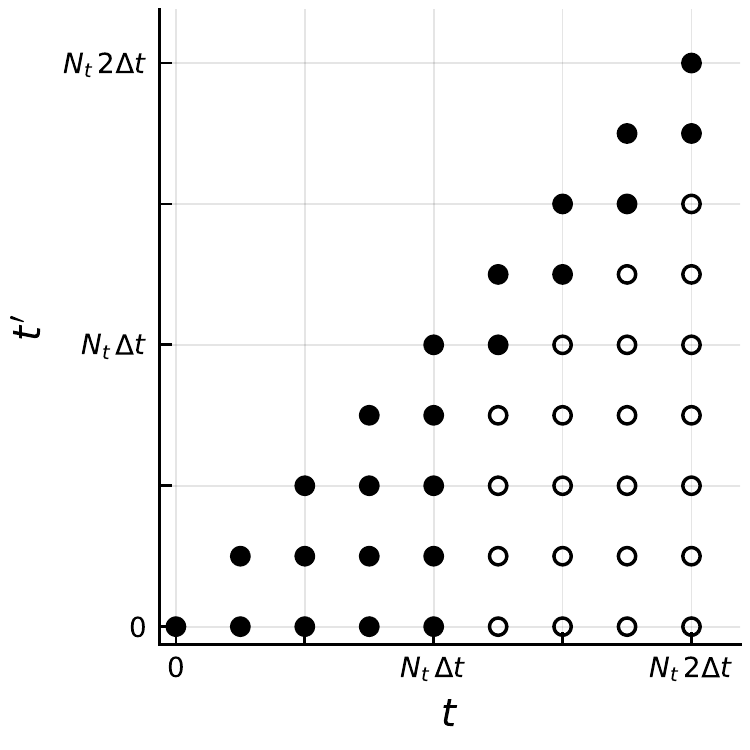}
		}
		\subfigure[step 3.]{
		\includegraphics[scale=.5]{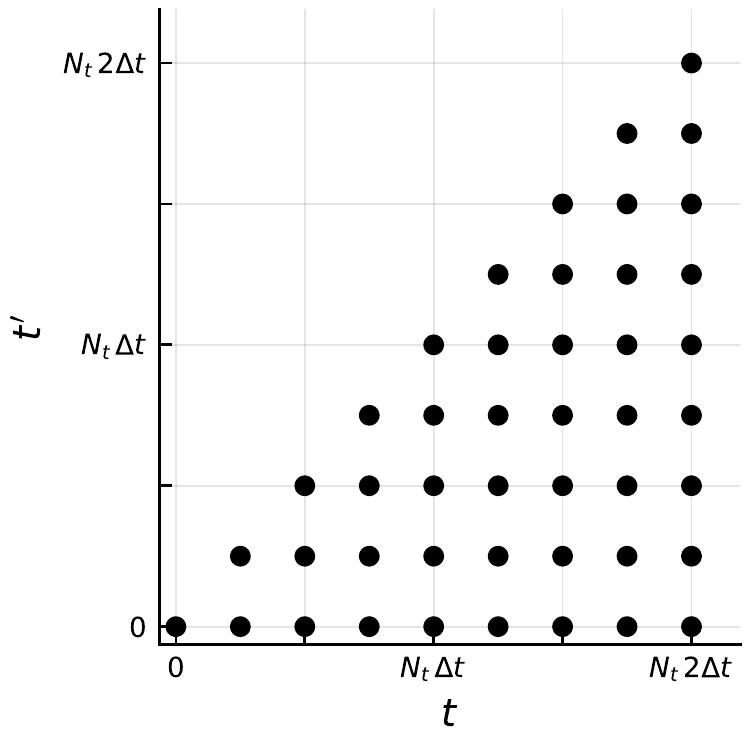}
		}
		\caption{Representation of the initialization and the first two iterations for the evaluation of a two-times observable using the dynamic-grid algorithm. The empty circles represent slots allocated in memory but not associated to any specific value, while the full circles are memory slots already associated. For any two time function, it first allocates the memory (a), than it fills half of the grid by linear propagation (b). Still using linear propagation it fills the slots with $t-t'\ll1$ (c), and it sets the other values by imposing self-consistency (d). Finally it halves the grid (e), doubles time step and it allocates the memory (f). Then the algorithm loops following the same scheme as in (b-c-d-e).}
		\label{fig:algorithm KL}
	\end{figure}
	The numerical scheme we are going to discuss is presented in
        the Bayes-optimal case where $T_2(t) \equiv T_p(t) \equiv
        1$. However the derivation that we propose can be easily
        generalized to the case where the $T$s assume different
        values, but are constants. Therefore we do not employ this algorithm to solve the LSE equations in the annealing protocol for which instead we use the fixed time-grid algorithm. It is convenient to manipulate the equations to obtain an equivalent set of equations for the functions $C(t,t')$, $Q(t,t') \doteq 1 - C(t,t') - \int_{t'}^t R(t,t'')dt''$, $\Cmag(t)$, where $Q(t,t')$ represents the deviation from Fluctuation Dissipation Theorem (FDT) at time $t$ starting from time $t'$.
	Indeed when the FDT theorem holds, it states that $R(t,t') = -\partial_t C(t,t')$.

	We briefly anticipate the strategy that the algorithm uses to solve the equations. The algorithm discretizes the times into $N_t$ intervals, first starting from the boundary conditions, $C(t,t) = 1$, $Q(t,t) = 0$ and $\Cmag = \Cmag_0\in[0;1]$, it fills the grid for small times (or small time differences $\tau=t-t'\ll1$) using linear propagation. Given a time $t$ and the initial guess for the Lagrange multiplier obtained by the linear propagator, the integrals are discretized and evaluated, then the results is used to update the value of the Lagrange multiplier. This procedure is repeated iteratively until convergence. Once that the first grid is filled, it follows a coarse-graining procedure where the sizes of the time intervals is doubled and only half of the information is retained. This procedure is repeated a fixed number of doubling of the original grid. The doubling scheme allows to explore exponentially long times at the cost of loosing part of the information, the direct consequence of this is the loss of stability for very large times (especially when the functions $C(t,t')$, $R(t,t')$, $\Cmag(t)$ undergo fast changes at large times).

	\paragraph{Dynamical equations in the algorithm.} We recall the function $f_k(x) = \frac{x^k}2$ and its derivatives, $f_k'(x) = \frac{kx^{k-1}}2$ and $f_k''(x) = \frac{k(k-1)x^{k-2}}2$. For simplicity in the notation, we introduce also $f_k(t,t') \doteq f_k\left(C(t,t')\right)$
	\begin{align*}
		(\partial_t+\mu(t))C(t,t')  & = 2R(t',t)+r_2\Cmag(t')f_2'(\Cmag(t))+r_p\Cmag(t')f_p'(\Cmag(t))+                                                          \\
		                            & +\frac1{\Delta_2}\int_0^{t'}dt''f_2'(t,t'')R(t',t'')+\frac1{\Delta_2}\int_0^{t}dt''f_2''(t,t'')R(t,t'')C(t',t'')+          \\
		                            & +\frac2{p\Delta_p}\int_0^{t'}dt''f_p'(t,t'')R(t',t'')+\frac2{p\Delta_p}\int_0^{t}dt''f_p''(t,t'')R(t,t'')C(t',t'')\,,
		\\
		(\partial_t+\mu(t))R(t,t')  & = \delta(t-t')+\frac1{\Delta_2}\int_{t'}^t dt''f_2''(t,t'')R(t,t'')R(t'',t')+                                              \\
		                            & +\frac2{p\Delta_p}\int_{t'}^t dt''f_p''(t,t'')R(t,t'')R(t'',t')\,,
		\\
		(\partial_t+\mu(t))\Cmag(t) & =r_2f_2'(\Cmag(t))+r_pf_p'(\Cmag(t))+                                                                                      \\
		                            & +\frac1{\Delta_2}\int_0^tdt''f_2''(t,t'')R(t,t'')\Cmag(t'')+\frac2{p\Delta_p}\int_0^tdt''f_p''(t,t'')R(t,t'')\Cmag(t'')\,,
		\\
		\mu(t)                      & = 1+r_2\Cmag(t)f_2'(\Cmag(t))+r_p\Cmag(t)f_p'(\Cmag(t))+                                                                   \\
		                            & +\frac2{\Delta_2}\int_0^tdt''f_2'(t,t'')R(t,t'')+\frac2{\Delta_p}\int_0^tdt''f_p'(t,t'')R(t,t'')\,.
	\end{align*}
	Following the lines of \cite{berthier2007spontaneous}, we introduce the FDT violation function, $Q(t,t')$, and after some manipulation the systems becomes
	\begin{align}\label{eq:sys for numerics}
		\begin{split}
			(\partial_t+ &\mu(t))C(t,t') = \Cmag(t')\left[r_2f_2'(\Cmag(t))+r_pf_p'(\Cmag(t))\right]+\\
			&+\frac1{\Delta_2}\Big\{\int_0^{t'}dt''\left[f_2'(t,t'')\frac{\partial Q(t',t'')}{\partial t''}+f_2''(t,t'')\frac{\partial Q(t,t'')}{\partial t''}C(t',t'')\right]+\\
			&-\int_{t'}^{t}dt''\left[f_2'(t,t'')\frac{\partial C(t'',t')}{\partial t''}-f_2''(t,t'')\frac{\partial Q(t,t'')}{\partial t''}C(t'',t')\right]+f_2'(1)C(t,t')-f_2'(t,0)C(t',0)\Big\}+\\
			&+\frac2{p\Delta_p}\Big\{\int_0^{t'}dt''\left[f_p'(t,t'')\frac{\partial Q(t',t'')}{\partial t''}+f_p''(t,t'')\frac{\partial Q(t,t'')}{\partial t''}C(t',t'')\right]+\\
			&-\int_{t'}^{t}dt''\left[f_p'(t,t'')\frac{\partial C(t'',t')}{\partial t''}-f_p''(t,t'')\frac{\partial Q(t,t'')}{\partial t''}C(t'',t')\right]+f_p'(1)C(t,t')-f_p'(t,0)C(t',0)\Big\}\,,
		\end{split}
		\\
		\begin{split}
			(\partial_t+ &\mu(t))Q(t,t') = \mu(t)-1+\frac1{\Delta_2}\Big\{-\int_{t'}^t dt''f_2'(t,t'')\frac{\partial Q(t'',t')}{\partial t''}+\int_{t'}^t dt''f_2''(t,t'')\frac{\partial Q(t,t'')}{\partial t''}[Q(t'',t')-1]+\\
			&+f_2'(1)[Q(t,t')-1]+f_2'(t,0)C(t',0)-\int_0^{t'}dt''\left[f_2'(t,t'')\frac{\partial Q(t',t'')}{\partial t''}+f_2''(t,t'')\frac{\partial Q(t,t'')}{\partial t''}C(t',t'')\right]\Big\}+\\
			&+\frac2{p\Delta_p}\Big\{-\int_{t'}^t dt''f_p'(t,t'')\frac{\partial Q(t'',t')}{\partial t''}+\int_{t'}^t dt''f_p''(t,t'')\frac{\partial Q(t,t'')}{\partial t''}[Q(t'',t')-1]+\\
			&+f_p'(1)[Q(t,t')-1]+f_p'(t,0)C(t',0)-\int_0^{t'}dt''\left[f_p'(t,t'')\frac{\partial Q(t',t'')}{\partial t''}+f_p''(t,t'')\frac{\partial Q(t,t'')}{\partial t''}C(t',t'')\right]\Big\}+\\
			&-\Cmag(t')\left[r_2f_2'(\Cmag(t))+r_pf_p'(\Cmag(t))\right]\,,
		\end{split}
		\\
		\begin{split}
			(\partial_t+ &\mu(t))\Cmag(t) = r_2f_2'(\Cmag(t))+r_pf_p'(\Cmag(t))+\frac1{\Delta_2}\Big\{f_2'(1)\Cmag(t)-f_2'(t,0)\Cmag(0)
			-\int_0^tdt''f_2'(t,t'')\frac{d}{d t''}\Cmag(t'')+\\
			&+\int_0^tdt''f_2''(t,t'')\frac{\partial Q(t,t'')}{\partial t''}\Cmag(t'')\Big\}+\frac2{p\Delta_p}\Big\{f_p'(1)\Cmag(t)-f_p'(t,0)\Cmag(0)-\int_0^tdt''f_p'(t,t'')\frac{d}{d t''}\Cmag(t'')+\\
			&+\int_0^tdt''f_p''(t,t'')\frac{\partial Q(t,t'')}{\partial t''}\Cmag(t'')\Big\}\,,
		\end{split}
		\\
		\begin{split}
			\mu(t) &= 1+r_2\Cmag(t)f_2'(\Cmag(t))+r_p\Cmag(t)f_p'(\Cmag(t))+\frac2{\Delta_2}[f_2(1)-f_2(t,0)]+\frac2{\Delta_p}[f_p(1)-f_p(t,0)]+\\
			&+\int_0^tdt''\left[\frac2{\Delta_2}f_2'(t,t'')+\frac2{\Delta_p}f_p'(t,t'')\right]\frac{\partial Q(t,t'')}{\partial t''}\,,
		\end{split}
	\end{align}
	further simplifications can be obtained introducing $\mu'(t)=\mu(t)-\frac2{\Delta_2}f_2(1)-\frac2{\Delta_p}f_p(1)$
	\begin{align}\label{eq:sys for numerics mu'}
		\begin{split}
			\mu'(t) &= 1+r_2\Cmag(t)f_2'(\Cmag(t))+r_p\Cmag(t)f_p'(\Cmag(t))-\frac2{\Delta_2}f_2(t,0)-\frac2{\Delta_p}f_p(t,0)+\\
			&+\int_0^tdt''\left[\frac2{\Delta_2}f_2'(t,t'')+\frac2{\Delta_p}f_p'(t,t'')\right]\frac{\partial Q(t,t'')}{\partial t''}\,,
		\end{split}
		\\
		\begin{split}
			(\partial_t+ &\mu'(t))C(t,t') = \Cmag(t')\left[r_2f_2'(\Cmag(t))+r_pf_p'(\Cmag(t))\right]+\\
			&+\frac1{\Delta_2}\Big\{\int_0^{t'}dt''\left[f_2'(t,t'')\frac{\partial Q(t',t'')}{\partial t''}+f_2''(t,t'')\frac{\partial Q(t,t'')}{\partial t''}C(t',t'')\right]+\\
			&-\int_{t'}^{t}dt''\left[f_2'(t,t'')\frac{\partial C(t'',t')}{\partial t''}-f_2''(t,t'')\frac{\partial Q(t,t'')}{\partial t''}C(t'',t')\right]-f_2'(t,0)C(t',0)\Big\}+\\
			&+\frac2{p\Delta_p}\Big\{\int_0^{t'}dt''\left[f_p'(t,t'')\frac{\partial Q(t',t'')}{\partial t''}+f_p''(t,t'')\frac{\partial Q(t,t'')}{\partial t''}C(t',t'')\right]+\\
			&-\int_{t'}^{t}dt''\left[f_p'(t,t'')\frac{\partial C(t'',t')}{\partial t''}-f_p''(t,t'')\frac{\partial Q(t,t'')}{\partial t''}C(t'',t')\right]-f_p'(t,0)C(t',0)\Big\}\,,
		\end{split}
		\\
		\begin{split}
			(\partial_t+ &\mu'(t))Q(t,t') = \mu'(t)-1-\Cmag(t')\left[r_2f_2'(\Cmag(t))+r_pf_p'(\Cmag(t))\right]+\\
			&+\frac1{\Delta_2}\Big\{-\int_{t'}^t dt''f_2'(t,t'')\frac{\partial Q(t'',t')}{\partial t''}+\int_{t'}^t dt''f_2''(t,t'')\frac{\partial Q(t,t'')}{\partial t''}[Q(t'',t')-1]+\\
			&+f_2'(t,0)C(t',0)-\int_0^{t'}dt''\left[f_2'(t,t'')\frac{\partial Q(t',t'')}{\partial t''}+f_2''(t,t'')\frac{\partial Q(t,t'')}{\partial t''}C(t',t'')\right]\Big\}+\\
			&+\frac2{p\Delta_p}\Big\{-\int_{t'}^t dt''f_p'(t,t'')\frac{\partial Q(t'',t')}{\partial t''}+\int_{t'}^t dt''f_p''(t,t'')\frac{\partial Q(t,t'')}{\partial t''}[Q(t'',t')-1]+\\
			&+f_p'(t,0)C(t',0)-\int_0^{t'}dt''\left[f_p'(t,t'')\frac{\partial Q(t',t'')}{\partial t''}+f_p''(t,t'')\frac{\partial Q(t,t'')}{\partial t''}C(t',t'')\right]\Big\}\,,
		\end{split}
		\\
		\begin{split}
			(\partial_t+ &\mu'(t))\Cmag(t) = r_2f_2'(\Cmag(t))+r_pf_p'(\Cmag(t))+\frac1{\Delta_2}\Big\{-f_2'(t,0)\Cmag(0)+\\
			&-\int_0^tdt''f_2'(t,t'')\frac{d}{d t''}\Cmag(t'')+\int_0^tdt''f_2''(t,t'')\frac{\partial Q(t,t'')}{\partial t''}\Cmag(t'')\Big\}+\\
			&+\frac2{p\Delta_p}\Big\{-f_p'(t,0)\Cmag(0)+\\
			&-\int_0^tdt''f_p'(t,t'')\frac{d}{d t''}\Cmag(t'')+\int_0^tdt''f_p''(t,t'')\frac{\partial Q(t,t'')}{\partial t''}\Cmag(t'')\Big\}\,.
		\end{split}
	\end{align}
	\paragraph{First order expansion coefficients.} In the numerics we will initialize the grid by a linear propagation of the initial conditions. To determine the coefficients to use we can expand the functions up the second term for small values of $\tau$ (and in the last equation of $t$)
	\begin{equation}\label{eq:numerical_linear_expantion}
		\begin{split}
			C(t'+\tau,t')&=C(t',t')+C^{(1,0)}(t',t')\tau+\frac12C^{(2,0)}(t',t')+O(\tau^3)\,
			, \\
			Q(t'+\tau,t')&=Q(t',t')+Q^{(1,0)}(t',t')\tau+\frac12Q^{(2,0)}(t',t')+O(\tau^3)\,
			, \\
			\Cmag(t)&=\Cmag(0)+\Cmag^{(1)}(0)\tau+\frac12\Cmag^{(2)}(0)+O(t^3)
			\, .
		\end{split}
	\end{equation}
	This gives the following coefficients: $C(t,t)=1$, $C^{(1,0)}(t,t)=-1$, $Q(t,t)=0$, $Q^{(1,0)}(t,t)=0$, $\Cmag(0)=\Cmag_0$ and $\Cmag^{(1)}(0)=\left[r_2f_2'(\Cmag_0)+r_pf_p'(\Cmag_0)\right](1-(\Cmag_0)^2)-\Cmag_0$, where $\Cmag_0$ is the initial value of the overlap with the signal.

	\paragraph{Numerical integration and derivation.} The set of equations derived above presents six types of integrals
	\begin{align*}
		 & I_{ij}^{(1AB)}=\int_{t_j}^{t_i}dt''A(t_i,t'')\frac{\partial B(t'',t_j)}{\partial t''}\,;            \\
		 & I_{ij}^{(2ABC)}=\int_{t_j}^{t_i}dt''A(t_i,t'')\frac{\partial B(t_i,t'')}{\partial t''}C(t'',t_j)\,; \\
		 & I_{ij}^{(3AB)}=\int_{0}^{t_j}dt''A(t_i,t'')\frac{\partial B(t_i,t'')}{\partial t''}\,;              \\
		 & I_{ij}^{(4ABC)}=\int_{0}^{t_j}dt''A(t_i,t'')\frac{\partial B(t_i,t'')}{\partial t''}C(t_j,t'')\,;   \\
		 & I_{i}^{(5AB)}=\int_{0}^{t_i}dt''A(t_i,t'')\frac{\partial B(t'')}{\partial t''}\,;                   \\
		 & I_{i}^{(6ABC)}=\int_{0}^{t_i}dt''A(t_i,t'')\frac{\partial B(t_i,t'')}{\partial t''}C(t'')\,.
	\end{align*}
	The integrals can be easily discretized
	\begin{align*}
		I_{ij}^{(2ABC)} & =\sum_{t_l=t_j+\delta t}^{t_i}\int_{t_l-\delta t}^{t_l}dt''A(t_i,t'')\frac{\partial B(t_i,t'')}{\partial t''}C(t'',t_j)\simeq                                                                 \\
		                & \simeq \sum_{t_l=t_j+\delta t}^{t_i}\int_{t_l-\delta t}^{t_l}dt_1A(t_i,t_1)\int_{t_l-\delta t}^{t_l}dt_2\frac{\partial B(t_i,t_2)}{\partial t_2}\int_{t_l-\delta t}^{t_l}dt_3C(t_3,t_j)\simeq \\
		                & \simeq \sum_{t_l=t_j+\delta t}^{t_i}\frac12[A(t_i,t_l)+A(t_i,t_l-\delta t)][B(t_i,t_l)-B(t_i,t_l-\delta t)]\frac12[C(t_l,t_j)+C(t_l-\delta t,t_j)]\;.
	\end{align*}
	In particular the 6 integrals become
	\begin{align}
		\begin{split}
			&I_{ij}^{(1AB)}=A_{im}B_{mj}-A_{ij}B_{jj}+\sum_{l=m+1}^{i}\frac12(A_{il}+A_{i(l-1)})(B_{lj}-B_{(l-1)j})+\\
			&\quad \quad \quad -\sum_{l=j+1}^{m}\frac12(B_{lj}+B_{(l-1)j})(A_{il}-A_{i(l-1)})=\\
			&\quad \quad \quad =A_{im}B_{mj}-A_{ij}B_{jj}+\sum_{l=m+1}^{i}dA_{il}^{(v)}(B_{lj}-B_{(l-1)j})-\sum_{l=j+1}^{m}(A_{il}-A_{i(l-1)})dB_{lj}^{(h)}\,;
		\end{split}
		\\
		\begin{split}
			&I_{ij}^{(2ABC)}=\sum_{l=j+1}^{i}\frac12(A_{il}+A_{i(l-1)})(B_{lj}-B_{(l-1)j})\frac12(C_{lj}+C_{(l-1)j})=\\
			&\quad \quad \quad =\sum_{l=m+1}^{i}dA_{il}^{(h)}(B_{il}-B_{i(l-1)})\frac12(C_{lj}+C_{(l-1)j})+\sum_{l=j+1}^{m}\frac12(A_{il}+A_{i(l-1)})(B_{il}-B_{i(l-1)})dC_{lj}^{(v)}\,;
		\end{split}
		\\
		 & I_{ij}^{(3AB)}=A_{ij}B_{jj}-A_{i0}B_{j0}-\sum_{l=1}^{j}(A_{il}-A_{i(l-1)})dB_{jl}^{(v)}\,;
		\\
		 & I_{ij}^{(4ABC)}=\sum_{l=1}^{j}\frac12(A_{il}+A_{i(l-1)})(B_{il}-B_{i(l-1)})dC_{jl}^{(v)}\,;
		\\
		 & I_{i}^{(5AB)}=\sum_{l=1}^{i}dA_{il}^{(v)}(B_{l}-B_{l-1})\,;
		\\
		 & I_{i}^{(6ABC)}=\sum_{l=1}^{i}\frac12(A_{il}+A_{i(l-1)})(B_{il}-B_{i(l-1)})dC_{l}\,,
	\end{align}
	where the superscript $(v)$ and $(h)$ represent the vertical ($t'$)
	and horizontal ($t$) derivatives in the discretized times, see
	Fig.~\ref{fig:algorithm KL} for an intuitive understanding.

	We also discretized the derivative using the last two time steps
	\begin{equation}
		\frac{d}{dt}g(t)=\frac3{2\delta t}g(t)-\frac2{\delta t}g(t-\delta t)+\frac1{2\delta t}g(t-2\delta t)+O(\delta t^3)\,.
	\end{equation}
	Given the time indices $i$ and $j$, we will define and evaluate the following quantities
	\begin{equation*}
		\begin{split}
			\{&C_{ij},Q_{ij},M2_{ij},N2_{ij},Mp_{ij},Np_{ij},\Cbar_i,P2_i,Pp_i,mu_i\}=\\
			&=\{C(t_i,t_j),Q(t_i,t_j),f_2'(C(t_i,t_j)),f_2''(C(t_i,t_j)),f_p'(C(t_i,t_j)),f_p''(C(t_i,t_j)),\Cmag(t_i),f_2'(\Cmag(t_i)),f_p'(\Cmag(t_i)),\mu(t_i)\}
		\end{split}
	\end{equation*}
	plus the respective vertical and horizontal derivatives.

	Calling $D_i=\frac3{2dt}+\mu'_i-\frac1{\Delta_2}M2_{ii}-\frac2{p\Delta_p}Mp_{ii}$, the original dynamical equations are integrated as follow
	\begin{align} \label{eq:self-consistent algo 1}
		\begin{split}
			C_{ij}D_i&= \frac2{dt}C_{(i-1)j}-\frac1{2dt}C_{(i-2)j}+\Cbar_j(r_2P2_i+r_pPp_i)+\\
			&+\frac1{\Delta_2}\left(-\dot{I}_{ij}^{(1f_2'C)}+I_{ij}^{(2f_2''QC)}+I_{ij}^{(3f_2'Q)}+I_{ij}^{(4f_2''QC)}-M2_{i0}C_{j0}\right)+\\
			&+\frac2{p\Delta_p} \left(-\dot{I}_{ij}^{(1f_p'C)}+I_{ij}^{(2f_p''QC)}+I_{ij}^{(3f_p'Q)}+I_{ij}^{(4f_p''QC)}-Mp_{i0}C_{j0}\right)\,,
		\end{split}
		\\
		\label{eq:self-consistent algo 2}
		\begin{split}
			Q_{ij}D_i&= \mu'_i-1+\frac2{dt}Q_{(i-1)j}-\frac1{2dt}Q_{(i-2)j}+\Cbar_j(r_2P2_i+r_pPp_i)+\\
			&+\frac1{\Delta_2}\left(-\dot{I}_{ij}^{(1f_2'Q)}+I_{ij}^{(2f_2''Q(Q-1))}-I_{ij}^{(3f_2'Q)}-I_{ij}^{(4f_2''QC)}-M2_{i0}C_{i0}\right)+\\
			&+\frac2{p\Delta_p} \left(-\dot{I}_{ij}^{(1f_p'Q)}+I_{ij}^{(2f_p''Q(Q-1))}-I_{ij}^{(3f_p'Q)}-I_{ij}^{(4f_p''QC)}-Mp_{i0}C_{i0}\right)\,,
		\end{split}
		\\
		\label{eq:self-consistent algo 3}
		\begin{split}
			\Cbar_iD_i&= \frac2{dt}\Cbar_{i-1}-\frac1{2dt}\Cbar_{i-2}+r_2P2_i+r_pPp_i+\\
			&+\frac1{\Delta_2}\left(-\dot{I}_i^{(5f_2'\Cbar)}+I_i^{(6f_2''Q\Cbar)}-M2_{i0}\Cbar_{0}\right)+\\
			&+\frac2{p\Delta_p} \left(-\dot{I}_i^{(5f_p'\Cbar)}+I_i^{(6f_p''Q\Cbar)}-Mp_{i0}\Cbar_{0}\right)\,.
		\end{split}
	\end{align}
	In the systems we used $\dot{I}$ to characterize the integrals where we remove from the sum the term present in the left-hand side (e.g. for $C_{ij}$ eq.~\ref{eq:self-consistent algo 1}).
	Using Simpson's integration formula we define the increments
	\begin{equation*}
		\begin{split}
			\Delta_{il}&=\frac1{12}(Q_{il}-Q_{i(l-1)})\{W_2^2[-(M2_{i(l+1)}+N2_{i(l+1)}C_{i(l+1)})+8(M2_{il}+N2_{il}C_{il})+5(M2_{i(l-1)}+N2_{i(l-1)}C_{i(l-1)})]+\\
			&+W_p^2[-(Mp_{i(l+1)}+Np_{i(l+1)}C_{i(l+1)})+8(Mp_{il}+Np_{il}C_{il})+5(Mp_{i(l-1)}+Np_{i(l-1)}C_{i(l-1)})]\}
		\end{split}
	\end{equation*}
	and we determine $\mu'$ as
	\begin{equation} \label{eq:self-consistent algo mu}
		\mu'=1+r_2P2_i+r_pPp_i+\delta \mu'+\sum_{l=1}^{i-N_t/4}\Delta_{il}-(W_2^2M2_{i0}+W_p^2Mp_{i0})C_{i0}\,,
	\end{equation}
	with $\delta \mu'$ initially set to 0.

	\paragraph{Algorithm:} Here we describe the main steps of the algorithm, pictorially represented Fig.~\ref{fig:algorithm KL}.

	Discretize the time $(t,t')$ in $N_t$ (even) intervals, the results shown use $N_t = 1024$.
	\begin{enumerate}
		\item \textbf{Initialization.} Fill the first $N_t/2$ times by linear propagation of the value obtained from the perturbative analysis
		      \begin{align}
			       & C_{ij}=1-(i-j)dt\,;                                                                                         \\
			       & Q_{ij}=0\,;                                                                                                 \\
			       & \Cbar_{i}=\Cmag_0+\left\{\left[r_2f_2'(\Cmag_0)+r_pf_p'(\Cmag_0)\right](1+(\Cmag_0)^2)-\Cmag_0\right\}dt\,; \\
			       & M2_{ij}=f_2'(C_{ij})\,;                                                                                     \\
			       & N2_{ij}=f_2''(C_{ij})\,;                                                                                    \\
			       & Mp_{ij}=f_p'(C_{ij})\,;                                                                                     \\
			       & Np_{ij}=f_p''(C_{ij})\,.
		      \end{align}
		\item \textbf{Fill the grid (small $\tau$).} Continue to propagate the values for small time differences $\tau=t-t'\ll1$. In terms of the algorithm it means that we have some elements of the grid, $N_c$ of them, close to the diagonal that will be updated by linear propagation because the approximation of small $\tau$ is still valid. In our simulation the first $\Delta t$ is of the order $10^{-7}$ and $N_c=2$.
		\item \textbf{Fill the grid (larger $\tau$).} The rest of the values will be copied from the previous $t$ ($A_{t+\Delta t,t'}=A_{t,t'}$). These values are the initial guess for solving the self-consistent equations (\ref{eq:self-consistent algo 1}-\ref{eq:self-consistent algo 2}-\ref{eq:self-consistent algo 3}) and (\ref{eq:self-consistent algo mu}), in this procedure the derivatives are updated using the 2nd order discretization.
		\item \textbf{Half the grid and expand.} The grid is decimated which means that each observable is contracted $A_{i,j}\leftarrow A_{2i,2j}$ and the derivate are updated as follows $dA^{(h)}_{i,j}\leftarrow \frac12(dA^{(h)}_{2i,2j}+dA^{(h)}_{2i-1,2j})$, $dA^{(v)}_{i,j}\leftarrow \frac12(dA^{(v)}_{2i,2j}+dA^{(v)}_{2i,2j-1})$. The new time step is now: $\Delta t \leftarrow 2\Delta t$.
		\item \textbf{Start over from step 2}.
	\end{enumerate}

	\subsection{Numerical checks on the dynamical algorithm}

	The dynamic-grid algorithm has been checked in a variety of ways.

	\begin{figure}[H]
		\centering
		\subfigure[$p=3 \quad \Delta_2=1.01$]{
		\includegraphics[scale=.4]{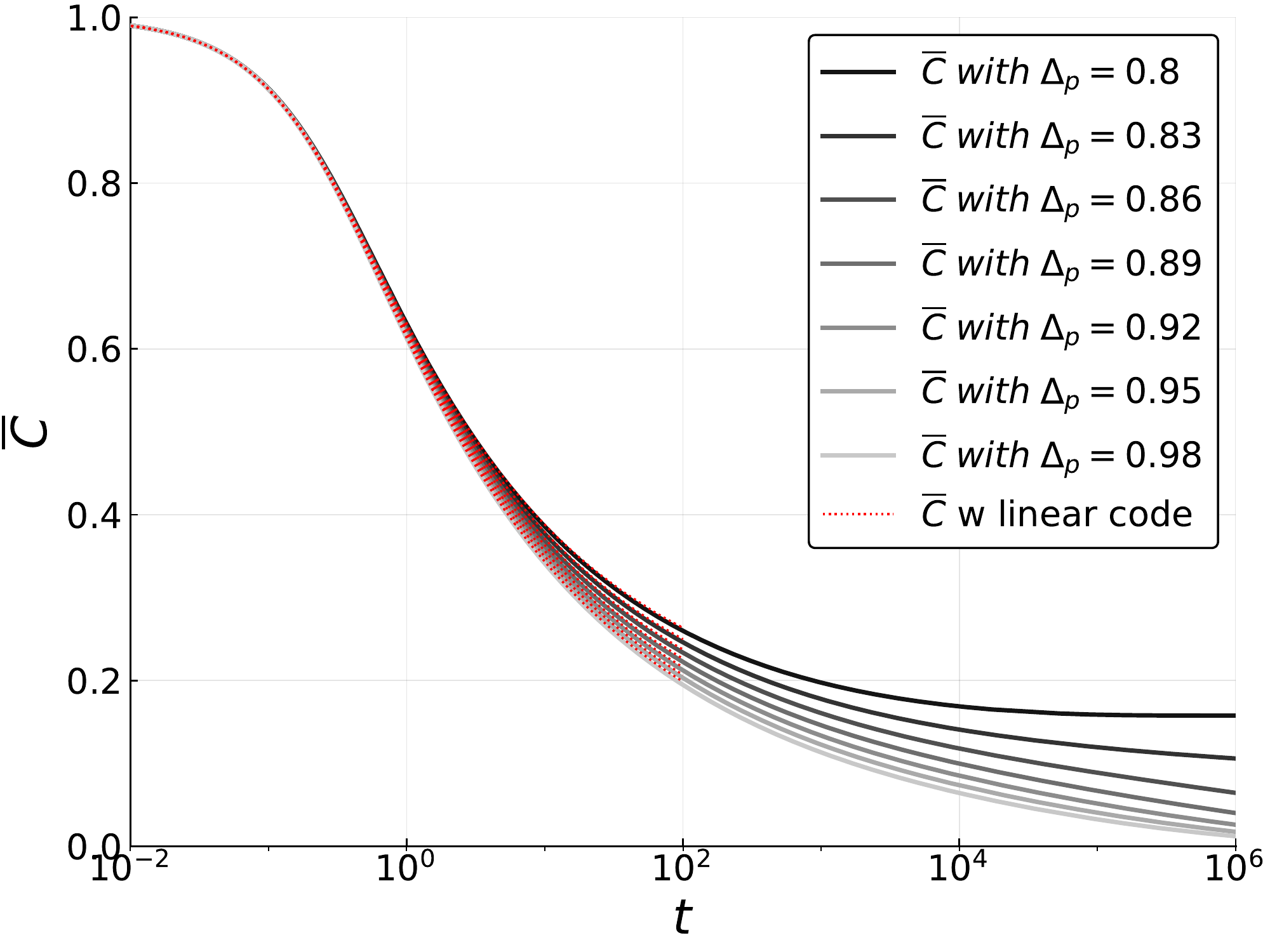}
		}
		\subfigure[$p=3 \quad \Delta_2=1.05$]{
		\includegraphics[scale=.4]{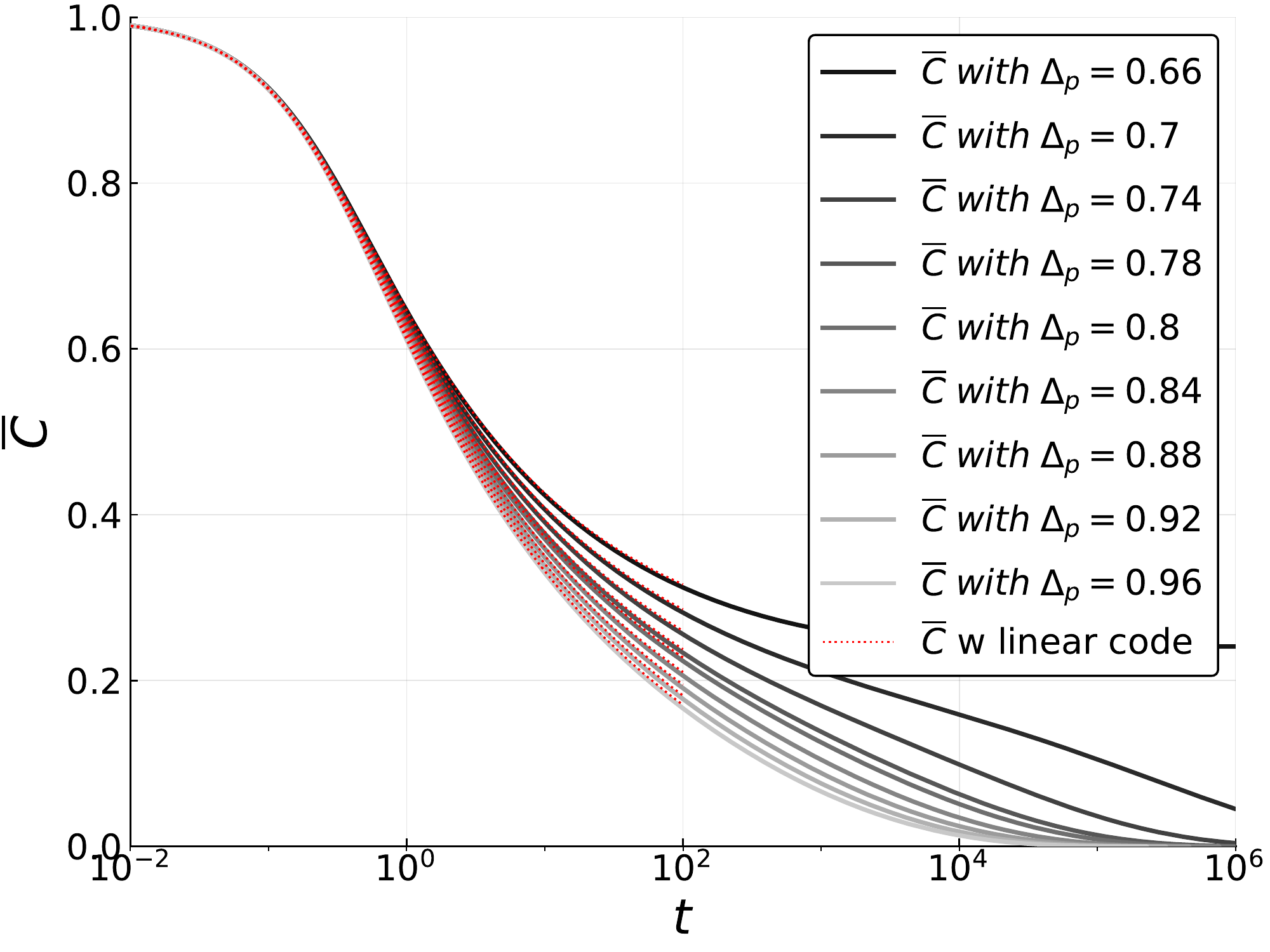}
		}
		\centering
		\subfigure[$p=3 \quad \Delta_2=1.10$]{
		\includegraphics[scale=.4]{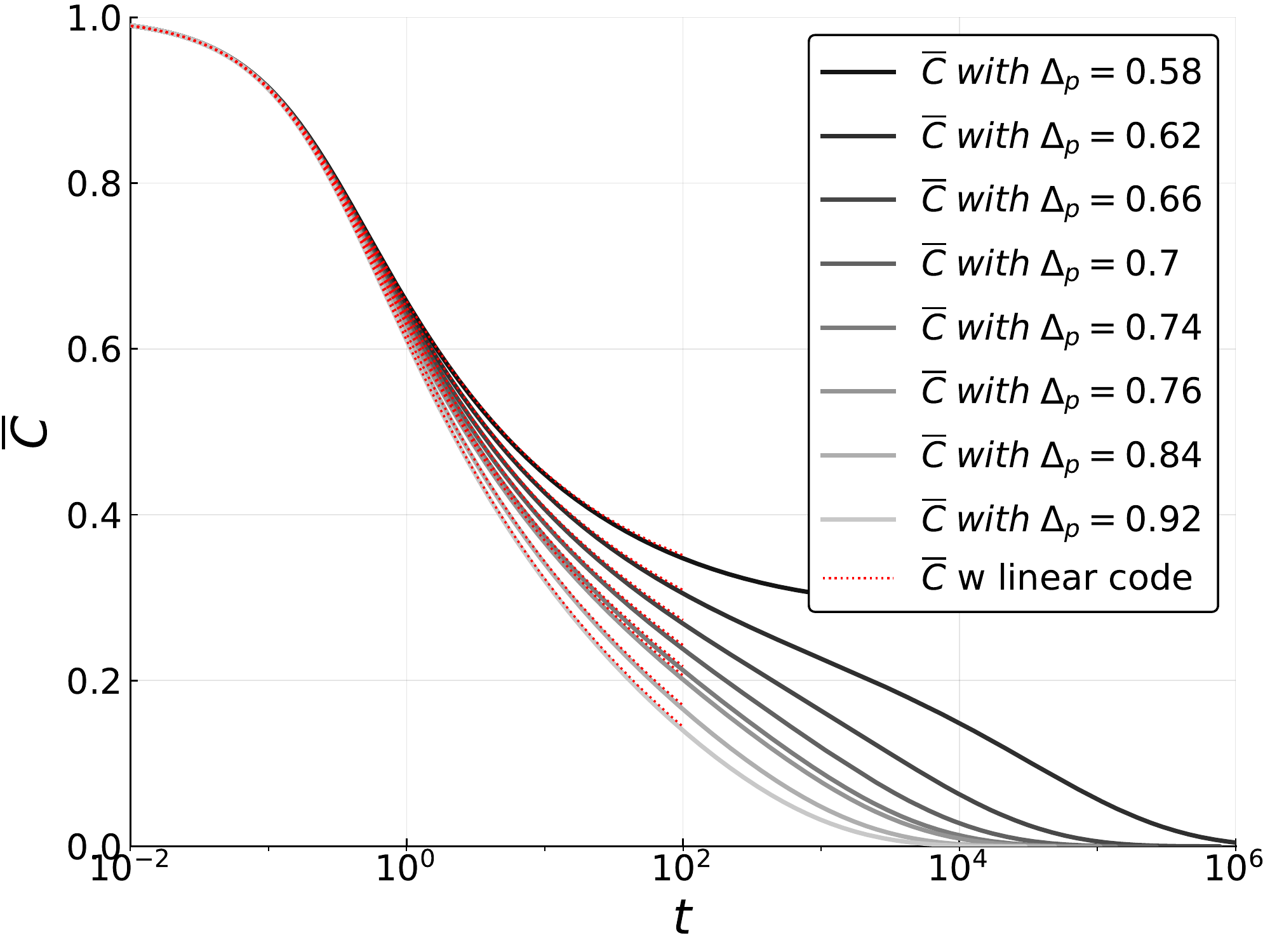}
		}
		\subfigure[$p=3 \quad \Delta_2=2.00$]{
		\includegraphics[scale=.4]{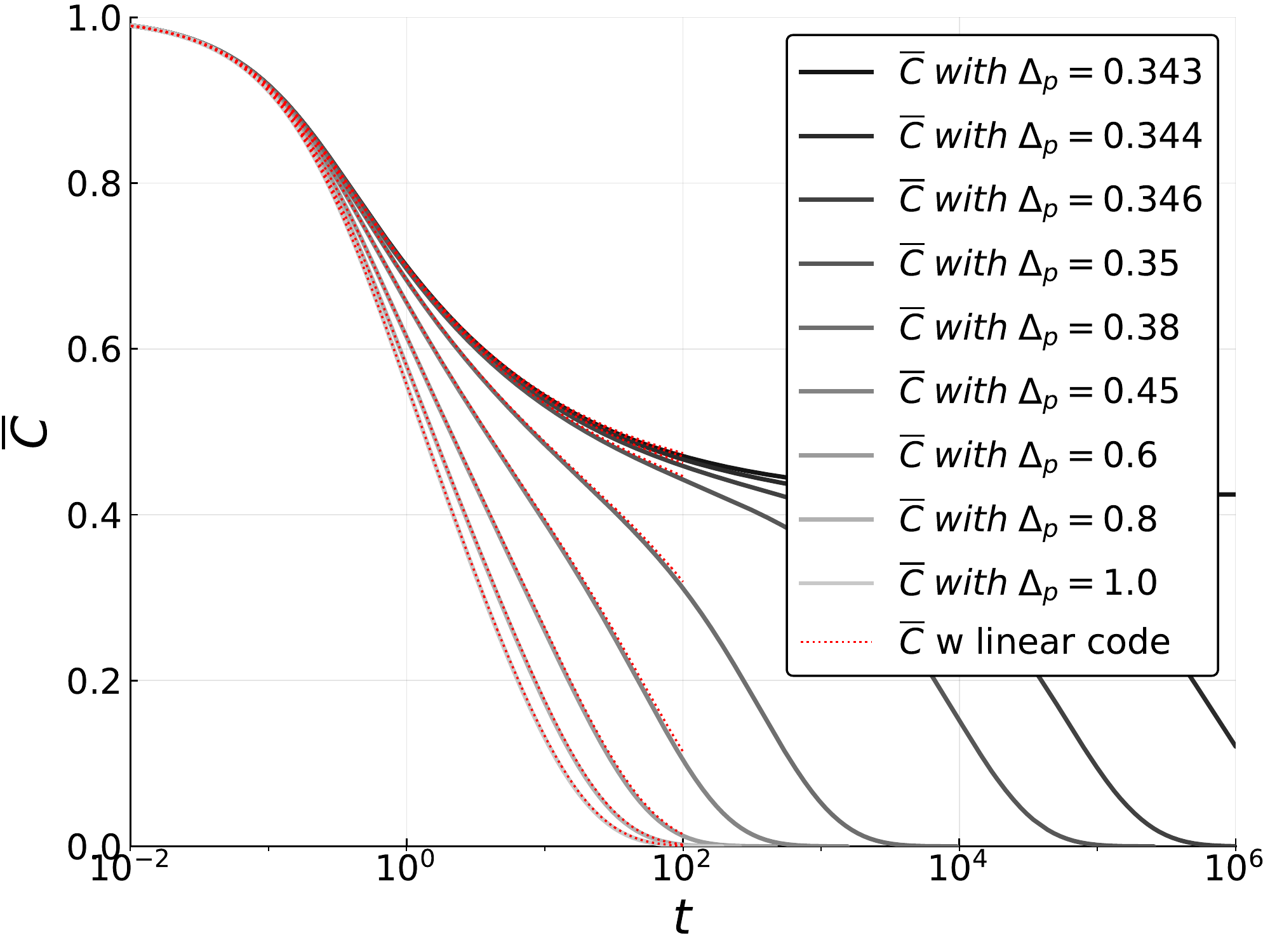}
		}
		\caption{Evolution of the correlation with the signal starting from the solution, $\Cmag_0=1.0$ at fixed $\Delta_2$ for different $\Delta_p$. The dotted red line overlapping with other lines, is the same quantity evaluated using the fixed grid algorithm up to time $100$. We have started the LSE from an informative initial condition.}
		\label{fig:overlap evolution comparison}
	\end{figure}

	\paragraph{Cross-checking using the fixed-grid algorithm.} For short times the dynamical equations were solved using the fixed-grid algorithm and compared with the outcome of the dynamic-grid algorithm, obtaining the same results, see Fig.~\ref{fig:overlap evolution comparison}. In the figure we used the fixed-grid with $t_{\max} = 100$ and the $\Delta t = 6.25*10^{-3}$.

	\paragraph{Same magnetization in the easy region. } In the impossible and easy regions, the overlap with the signal of both AMP and dynamic-grid integration, converges to the same value. In Fig.~\ref{fig:magnetization vs AMP} we show the overlap obtained with AMP, black dashed line, and the overlap achieved by the integration scheme at a given time. We can see that the overlap with the signal as obtained solving the LSE equations converges to the same value of the fixed point of AMP. Given a fixed $\Delta_2$ we can observe that the time to convergence increase very rapidly as we decrease $\Delta_p$. We fitted this increase of the \emph{relaxation time} to get the boundary of the Langevin hard region.

	\begin{figure}[H]
		\centering
		\subfigure[$p=3 \quad \Delta_2=0.40$]{
		\includegraphics[scale=.48]{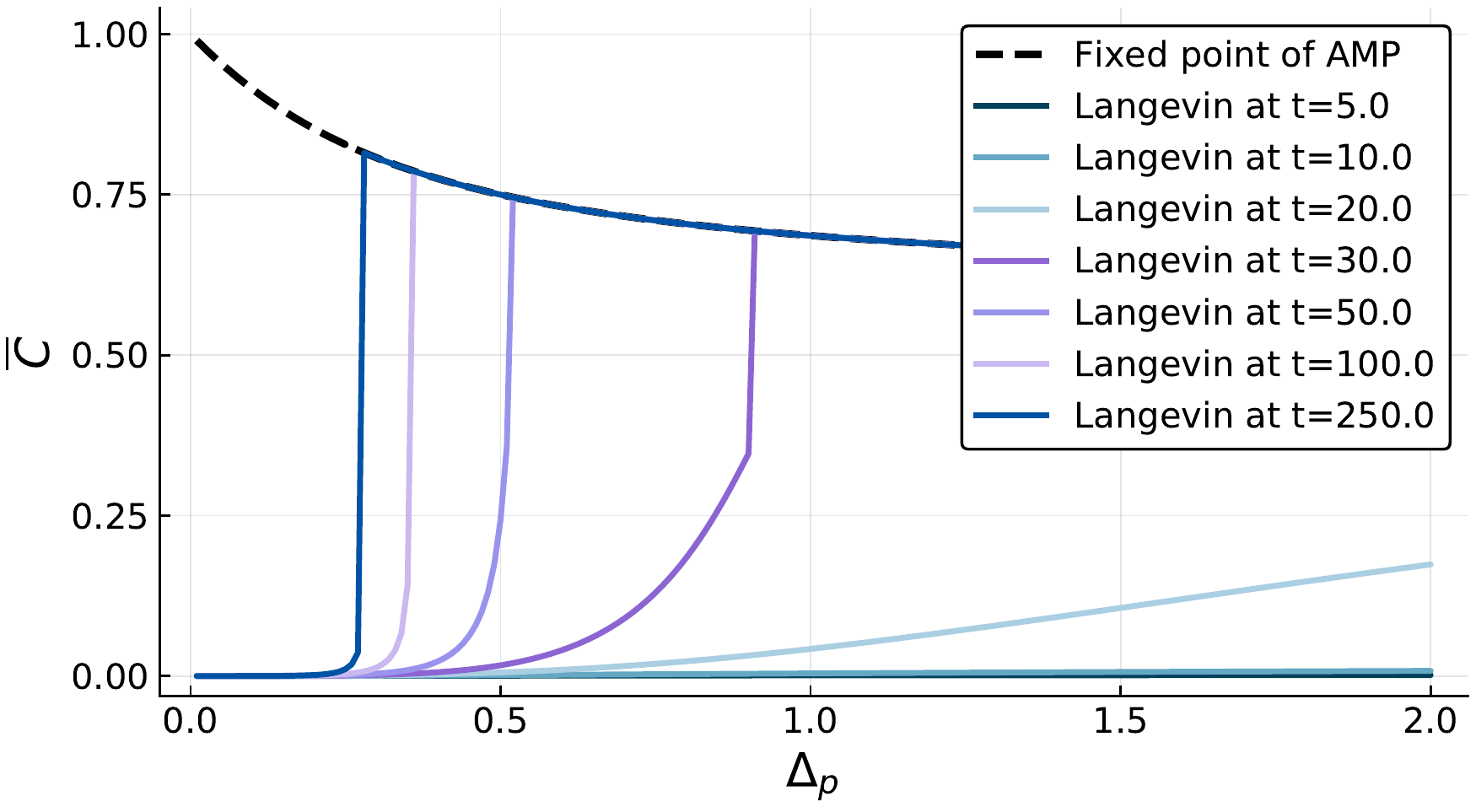}
		}
		\subfigure[$p=3 \quad \Delta_2=0.60$]{
		\includegraphics[scale=.48]{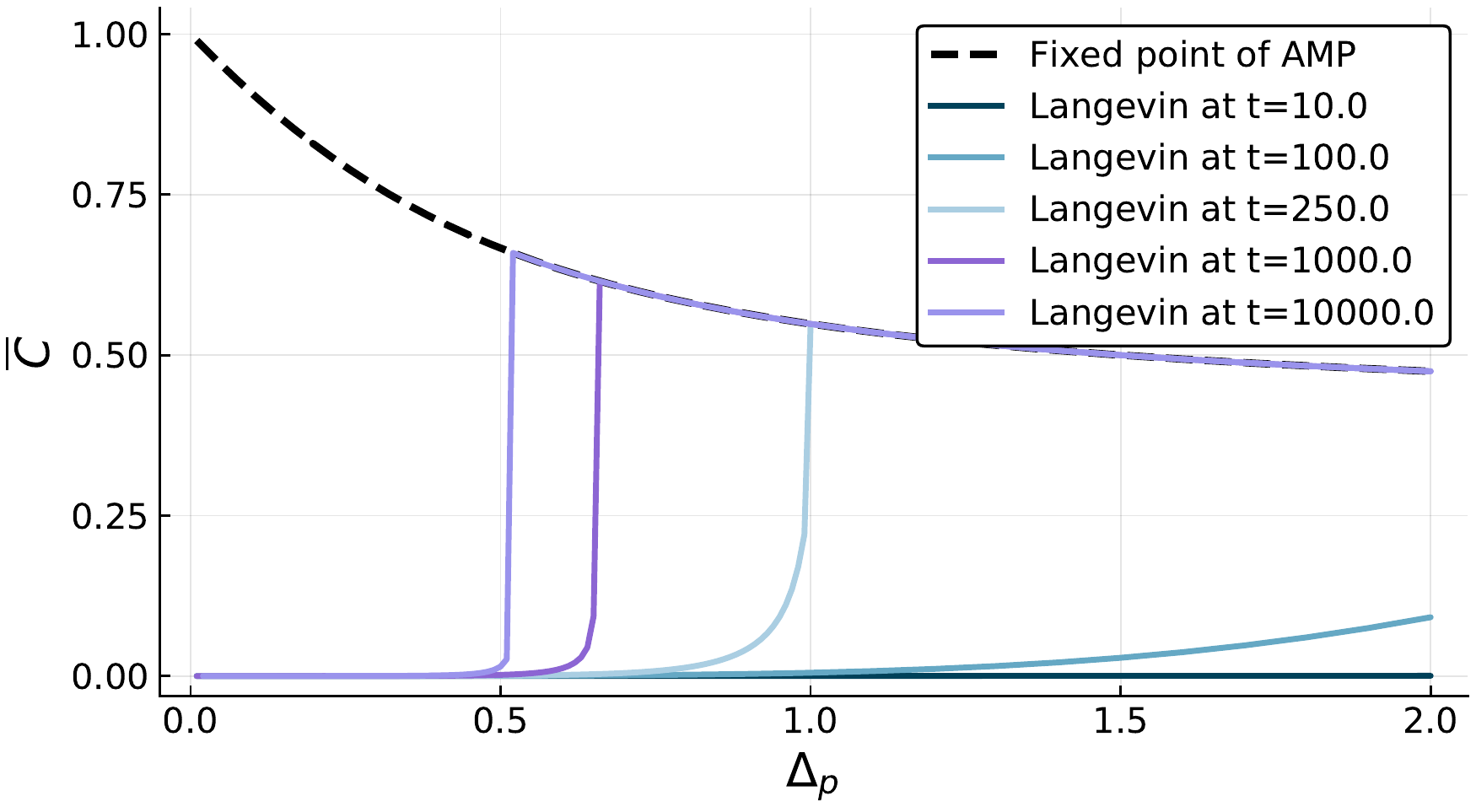}
		}
		\centering
		\subfigure[$p=3 \quad \Delta_2=0.70$]{
		\includegraphics[scale=.48]{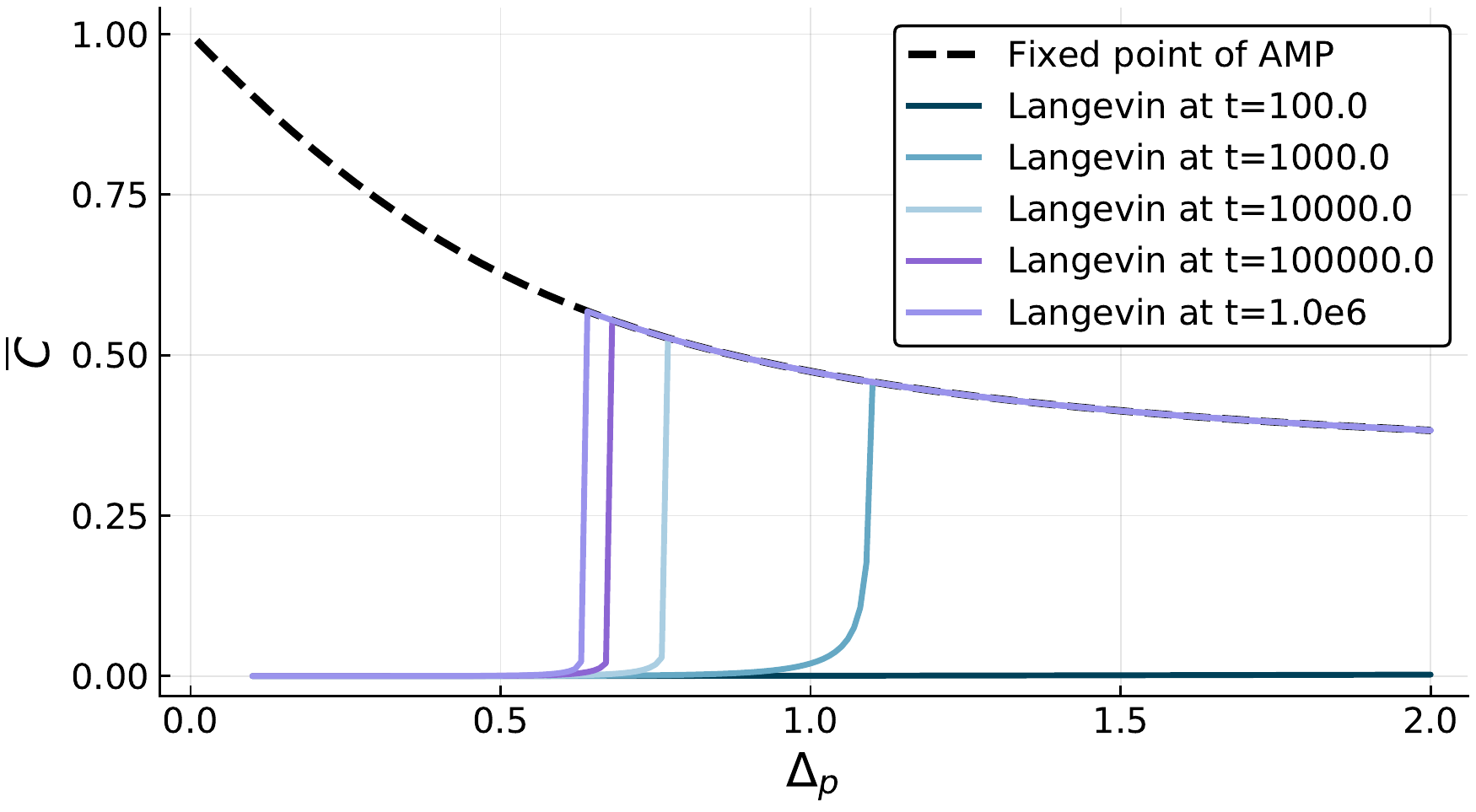}
		}
		\subfigure[$p=3 \quad \Delta_2=0.80$]{
		\includegraphics[scale=.48]{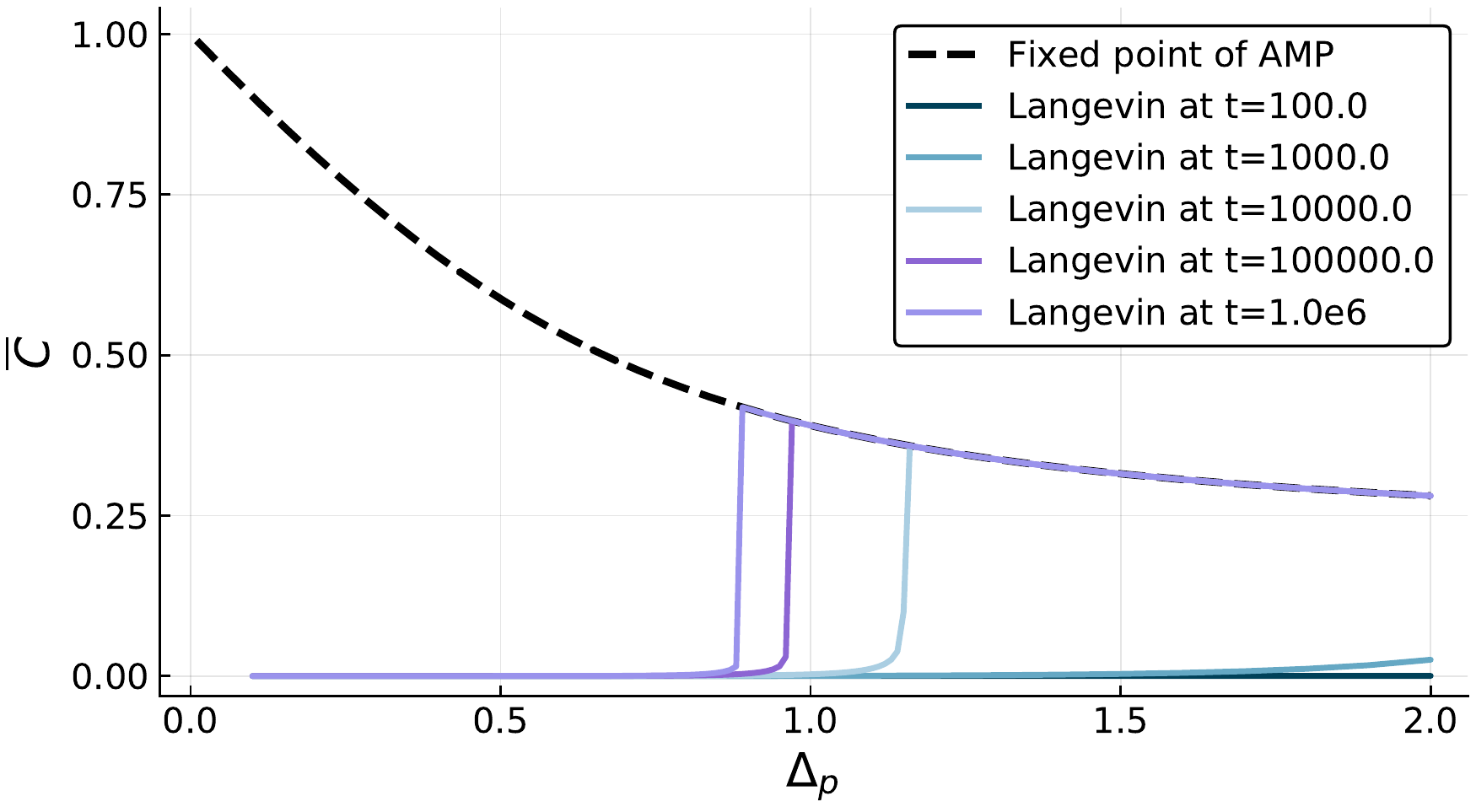}
		}
		\caption{Correlation with the signal of AMP (dotted lines) and Langevin (solid lines) at $k$th iteration and $t$ time respectively starting in both cases with an initial overlap of $10^{-4}$. The black dashed line is the asymptotic value predicted with AMP. In the easy region, provided enough running time, Langevin dynamics finds the same alignment as AMP. The figures show qualitatively the same behaviour for different values of $\Delta_2$.}
		\label{fig:magnetization vs AMP}
	\end{figure}

	\paragraph{Dynamical transition.}
	The dynamical transition where the finite magnetization fixed point disappears can be regarded as a clustering or dynamical glass transition.
	Indeed coming from the impossible phase, going towards the hard phase, at the dynamical transition the free energy landscape
	changes and the unique ergodic paramagnetic minimum of the impossible phase gets clustered into an exponential number of metastable glassy states (see Sec. \ref{sec:replica analysis}).
	Correspondingly the relaxation time of the Langevin algorithm diverges. Fitting this divergence with a power law we obtain an alternative estimation of the
	dynamical line.  In the right panel of Fig.~\ref{fig:phase diagram p=3_all} we plot with yellow points the dynamical transition line as extracted from the fit of the relaxation
	time of the Langevin algorithm extracted coming from the impossible phase and entering in the hard phase.

	\subsection{Extrapolation procedure}\label{SI:numerical_extrapolation}

	In order to determine the Langevin Hard region, given a fixed
        value of $\Delta_p$ ($\Delta_2$), we measure the relaxation
        time that it takes to relax to equilibrium. On approaching the
        Langevin hard region, this relaxation time increases and we
        extrapolate the growth to obtain the critical $\Delta_p^*$
        ($\Delta_2^*$ respectively) where the relaxation time appears
        to diverge. The extrapolation is done starting from $\Cmag_0=10^{-40}$ and assuming a power law divergence. Fig.~\ref{fig:phase diagram p=3_all} shows the results of this procedure for the cases $2+3$ and $2+4$. We remark that the divergence times increase as $p$ increases. Therefor in reason of the instability of the code for large times it becomes difficult to extrapolate the threshold accurately. In particular in the right panel of Fig.~\ref{fig:phase diagram p=3_all} when estimating the threshold for $2+4$, we consider horizontal sections and the points extrapolated for $\Delta_2$ close to the threshold $\Delta_2 = 1.0$ are very hard to estimate because of these instabilities.

	\begin{figure}
		\centering
		\includegraphics[scale=.5]{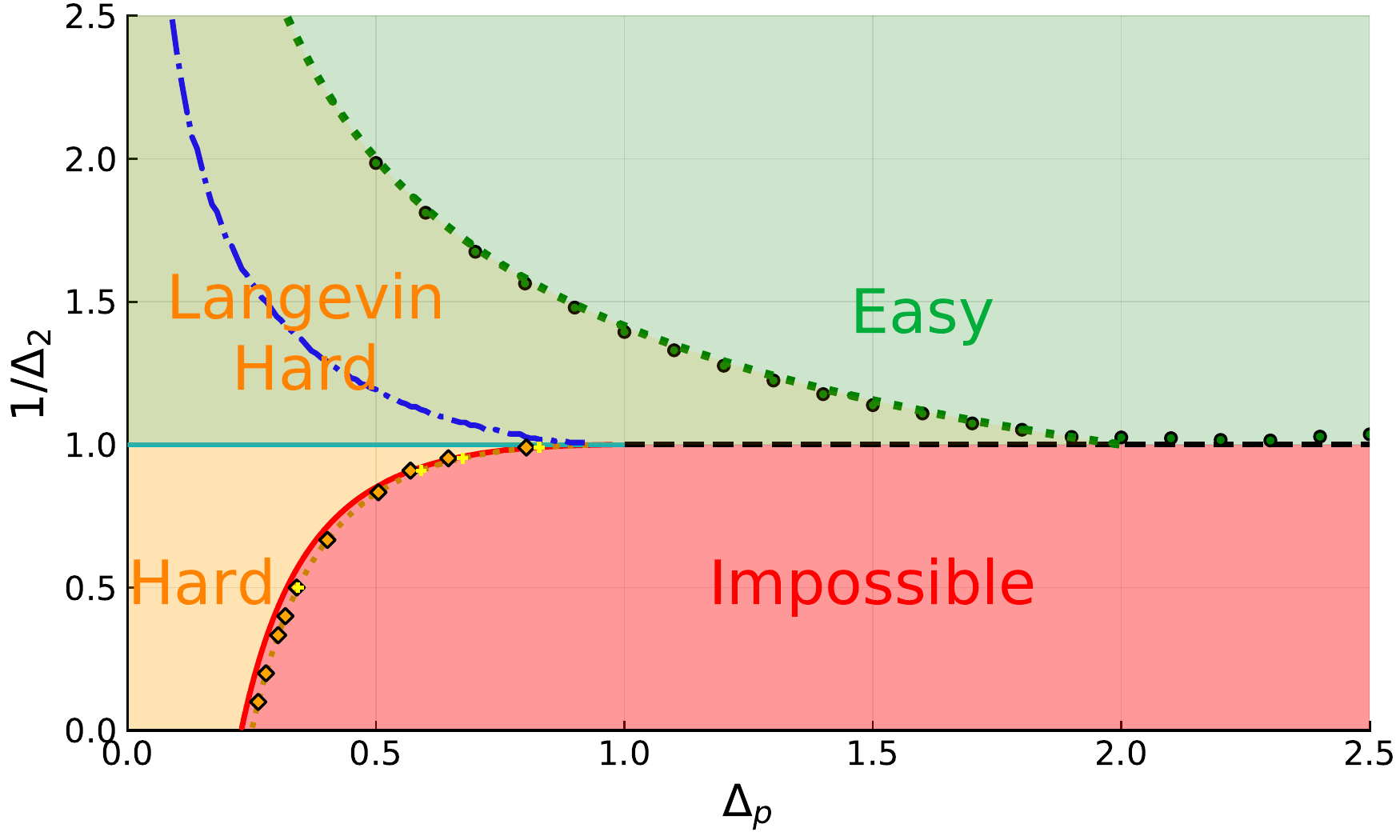}
		\includegraphics[scale=.5]{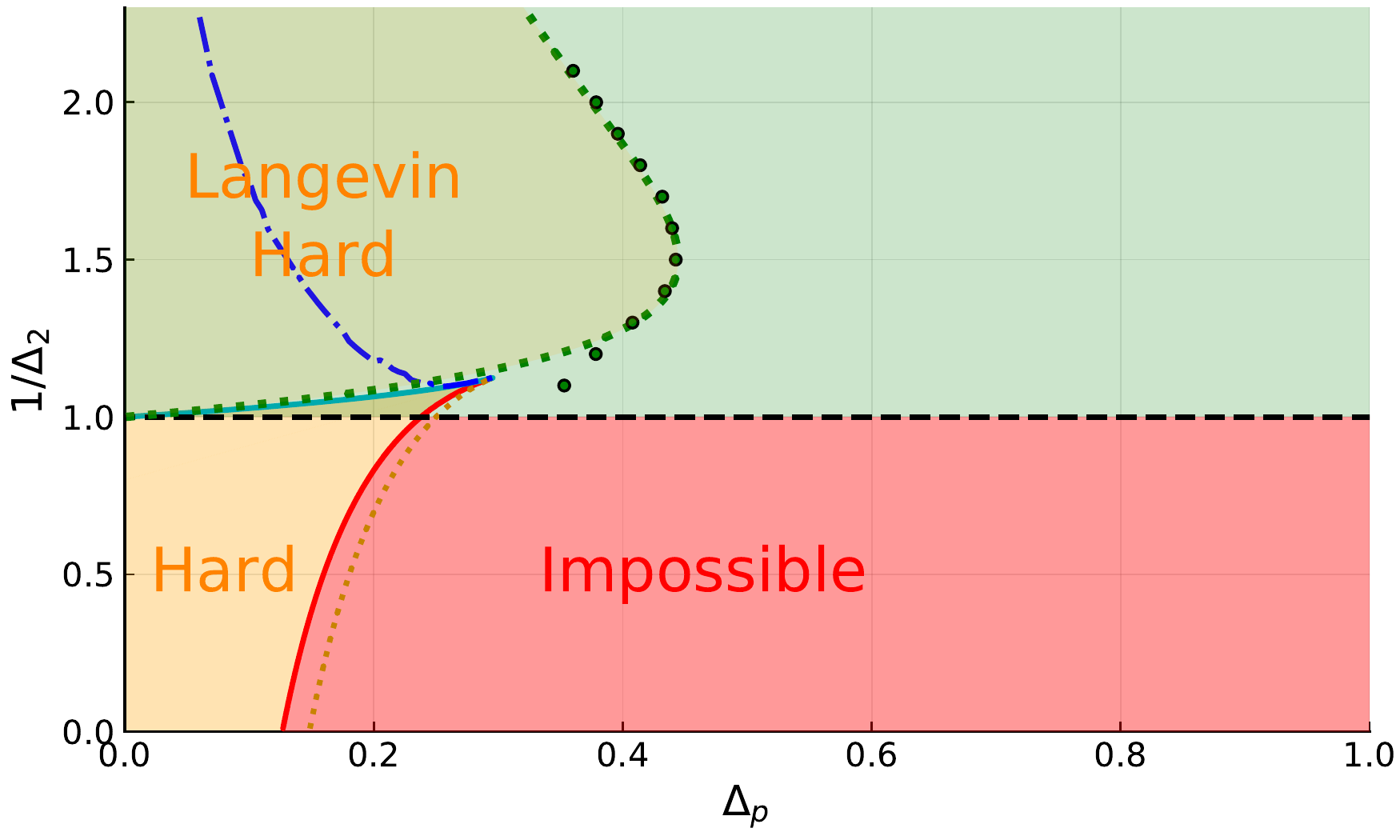}
		\caption{On the left: phase diagram of the spiked matrix-tensor model for
			$p=3$ as presented in the left panel of Fig.~\ref{fig:phase diagram p=3} with the
			additional boundary of the Langevin hard phase
                        (green circles and dotten green line).
			The data points (circles) have 
                        been obtained numerically by fitting the relaxation time
                        at fixed $\Delta_p$ and increasing
                        $\Delta_2$. The green dotted line are fixed
                        points of 
                        expression (\ref{eq:threshold}). 
			The blue dashed-dotted line
			marks a region above which we do not observe anymore a stable positive 1RSB complexity.
			Finally we plot with orange and yellow dots the dynamical transition line as extracted from the relaxation time of the Langevin algorithm coming from respectively the hard and impossible phase.
			On the right: phase diagram of the spiked matrix-tensor model for
			$p=4$ as presented in the right panel of
                        Fig.~\ref{fig:phase diagram p=3} with the
                        additional Langevin hard phase boundary. The
                        data points are obtained  fixing $\Delta_2$
                        and decreasing $\Delta_p$. Also
                        in this case we observe that Langevin hard
                        phase extends in the AMP easy
                        phase. Interestingly the Langevin hard phase
                        here folds and presents a re-entrant
                        behaviour, investigating the precise character
                        of this reentrance is hampered by vicinity of
                        the critical point and is left for future work.  The blue dashed line
			marks a region above which we do not observe anymore a stable positive 1RSB complexity.
		}    \label{fig:phase diagram p=3_all}
	\end{figure}

	\paragraph{Numerical checks on the extrapolation procedure.}

	To test
	the quality of the fits we use a similar numerical procedure to locate the
	spinodal of the informative solution, which is given by the points
	where the informative solution ceases to exist. This spinodal
	must be the same for both the AMP and the Langevin algorithm \cite{REVIEWFLOANDLENKA}.

	Since we aim at studying the spinodal of the informative solution, we initialize the LSE with $\Cmag_0=1$ and let it relax, measuring the time it takes to equilibrate at the value of $\Cmag$ given by the informative fixed point of AMP.
	We do this fixing $\Delta_2$ and changing $\Delta_p$.
	As we approach the critical $\Delta_{p,{\rm dyn}}$, the relaxation time will diverge and we can fit this divergence with a power law. The dynamic threshold extracted in this way is finally compared with the one obtained from AMP. In Fig.~\ref{fig:fitDeltapSol} we show how this scheme has been applied for $\Delta_2\in\{1.01,\;1.05,\;1.10,\;2.00\}$. As we get closer to the critical line $\Delta_2=1$ the relaxation time increases (and the height of the plateau decreases), making the fit harder.
	All in all, we observe a very good agreement between the points found with these extrapolation procedures and the prediction obtained with AMP, as shown in Fig.~\ref{fig:phase diagram p=3_all}.

	\subsection{Initial conditions}\label{SI:numerical_initial_condition}

	\begin{figure}
		\centering
		\includegraphics[scale=.45]{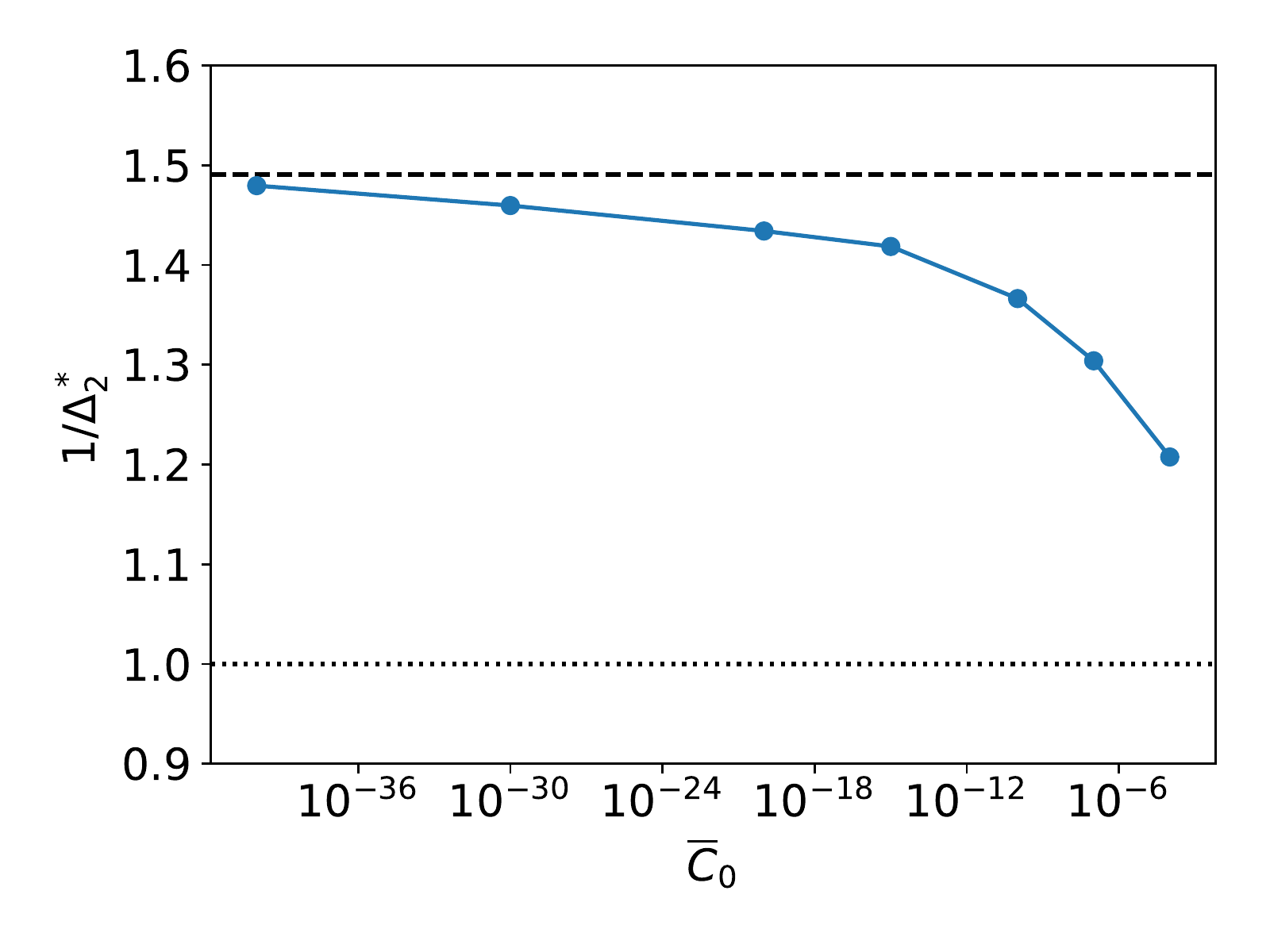}
		\caption{Estimated divergence point $\Delta_2^*$ at
                  fixed $\Delta_p=0.9$ with $p=3$ as a function of the
                  initial condition $\Cmag$, ranging from $10^{-40}$
                  to $10^{-4}$. The vertical axis is in linear scale
                  while the horizontal axis is in log-scale.
We observe that the dependence of estimated divergence points on the
initial condition is consistent with the asymptotic value
$1/\Delta_2^* = \sqrt{2/\Delta_p} \approx 1.491$ following from
eq.~(\ref{eq:threshold}), and depicted by the dashed line. 
The dotted line instead represents the AMP threshold for comparison.}
		\label{fig:relaxation_initial_condition}
	\end{figure}

	The LSE equations show a rather strong dependence on the initial condition $\Cmag_0$. A low initial magnetization will give a low initial momentum in the direction of the signal, as it the can be observed in the linear expansion of $\Cmag$ Eq.~\eqref{eq:numerical_linear_expantion}, and consequently the system will not be able to cross even very small barriers. The direct consequence is that reducing the initial magnetization the estimated threshold will get worse, in the sense that a larger signal-to-noise ratio will be required to find the solution. This finding can be observe in Fig.~\ref{fig:relaxation_initial_condition}, where different initial conditions are compared on the section $\Delta_p = 0.9$.
	
	Finally we remark how the different initial conditions affect the phase diagram. In Fig.~\ref{fig:phase_diagram_initial_condition} we compare the Langevin hard-easy threshold evaluated starting from different initial condition, showing that the region gets larger as the initial condition decrease up to convergence.
		
	\begin{figure}
		\centering
		\includegraphics[scale=.5]{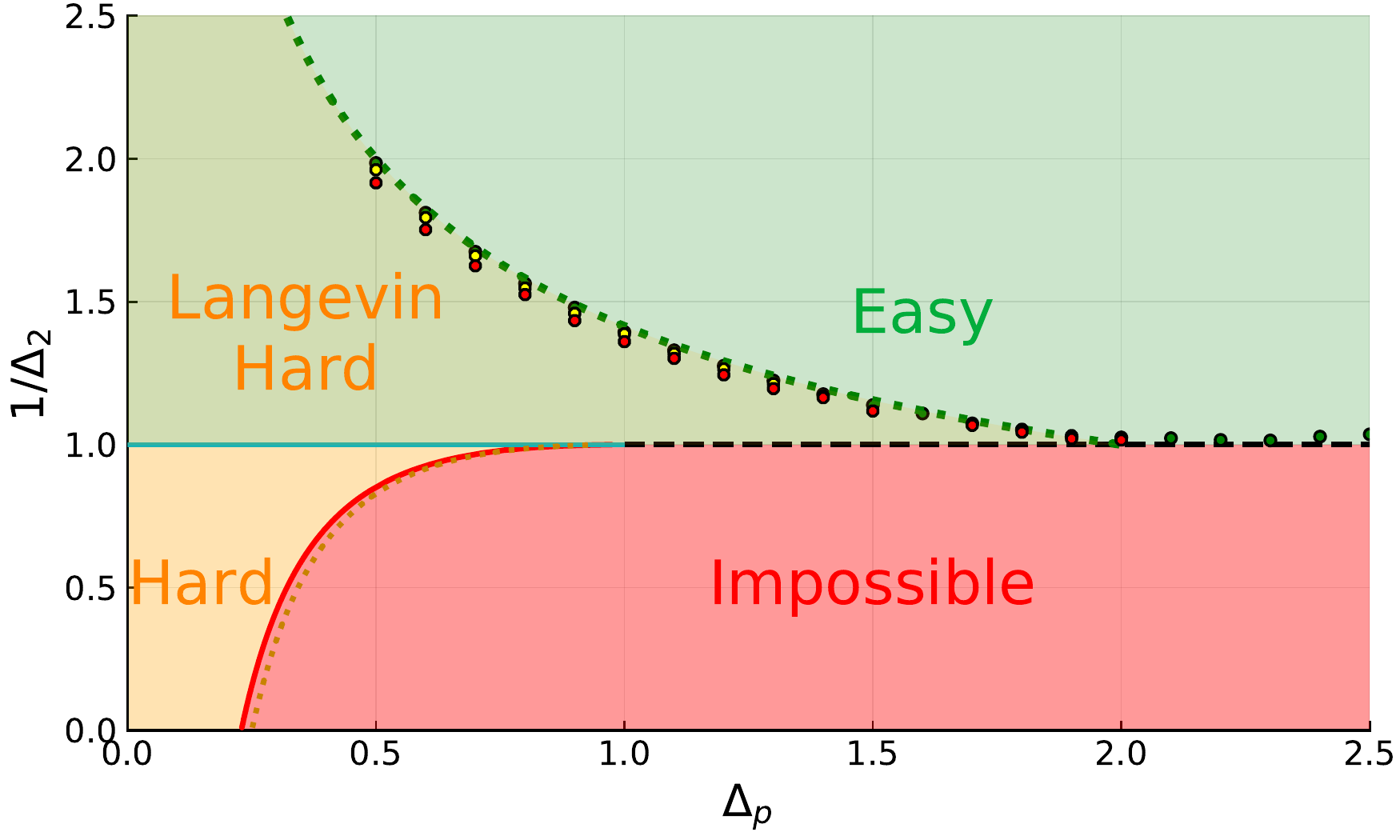}
		\caption{Phase diagram of the $p=3$ model where we compare the threshold of the Langevin hard region using different values of the initial conditions, respectively: green circles $\Cmag_0 = 10^{-40}$, yellow circles $\Cmag_0 = 10^{-30}$, red circles $\Cmag_0 = 10^{-20}$.}
		\label{fig:phase_diagram_initial_condition}
	\end{figure}
	
	\subsection{Annealing protocol}\label{SI:numerical_annealing}

	In this section we show that using specific protocols that
        separate the matrix and tensor part of the cost function
        (something we do not allow in the main part of this paper for
        the purpose of having a model as generic as possible) we are able to enter in the Langevin hard region.
	A generic annealing scheme would lower the noises of both the channels simultaneously, which will not be able to avoid the Langevin hard region. Instead we can use the following protocol
	\begin{equation}
		\begin{split}
			& T_2 \equiv 1 \, , \\
			& T_p \equiv T_p(t) = 1+\frac{C}{\Delta_p}e^{-\frac{t}{\tau_\text{ann}}} \ .
		\end{split}
	\end{equation}
	The constant $C$ allows to select at the initial time the desired effective
	$\Delta_{p,{\rm eff}}=\Delta_p+Ce^{-\frac{t}{\tau_\text{ann}}}$ far
	from (and much larger than) the original one. Instead $\tau_\text{ann}$
	chooses the speed of the annealing protocol. Fig.~\ref{fig:smartAnnealing}
	shows that using this protocol we are able to enter in Langevin Hard
	region even with $\Delta_2$ close to the AMP threshold. To this
	purpose we initiated the effective $\Delta_{p,{\rm eff}}$ close to $100$ (i.e. $C=100$), very far from the Langevin hard region, and we used different speeds for the annealing of $\Delta_p$ (different colors in the figures). In the figures we can observe that approaching the $\Delta_2 = 1$ we need slower and slower protocols (larger and larger $\tau_\text{ann}$). The reason for this behavior is due to the fact that approaching $\Delta_2 = 1$ with $\Delta_p=100$ a longer time is required to gain a non trivial overlap with the solution. Evidence of this growing timescale at $\Delta_p=100$ is given in Fig.~\ref{fig:relaxationTimeDeltap100} where we show the relaxation time for magnetizing to the solution varying $\Delta_2$. In particular, we can observe that at $\Delta_2 = 0.70$ the relaxation time is of the order of $100$ time units.

	\begin{figure}
		\centering
		\subfigure[$\Delta_2=1.01$: estimation $0.82857$ (AMP $0.82719$)]{
		\includegraphics[scale=.57]{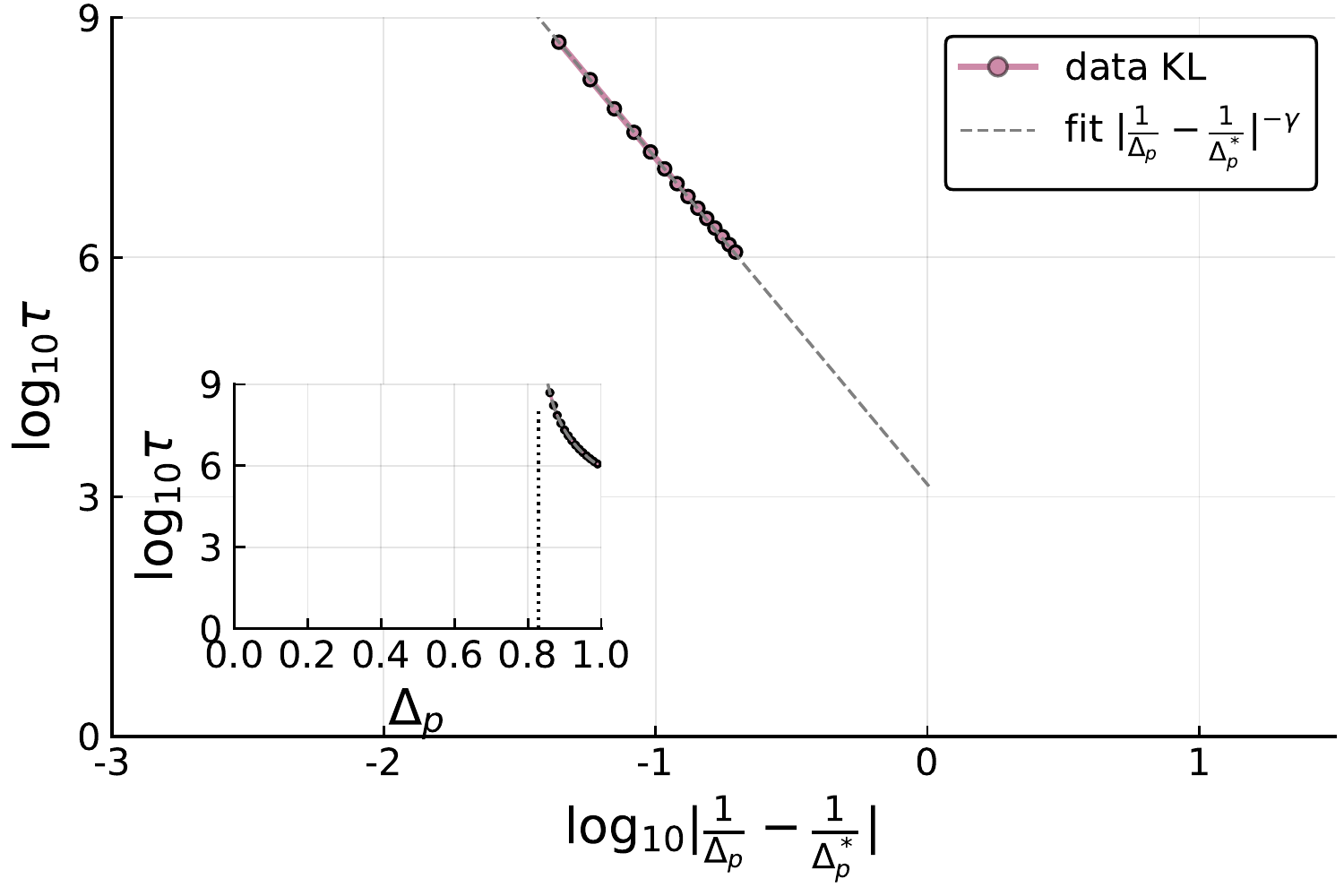}
		}
		\subfigure[$\Delta_2=1.05$: estimation  $0.67447$ (AMP $0.67382$)]{
		\includegraphics[scale=.57]{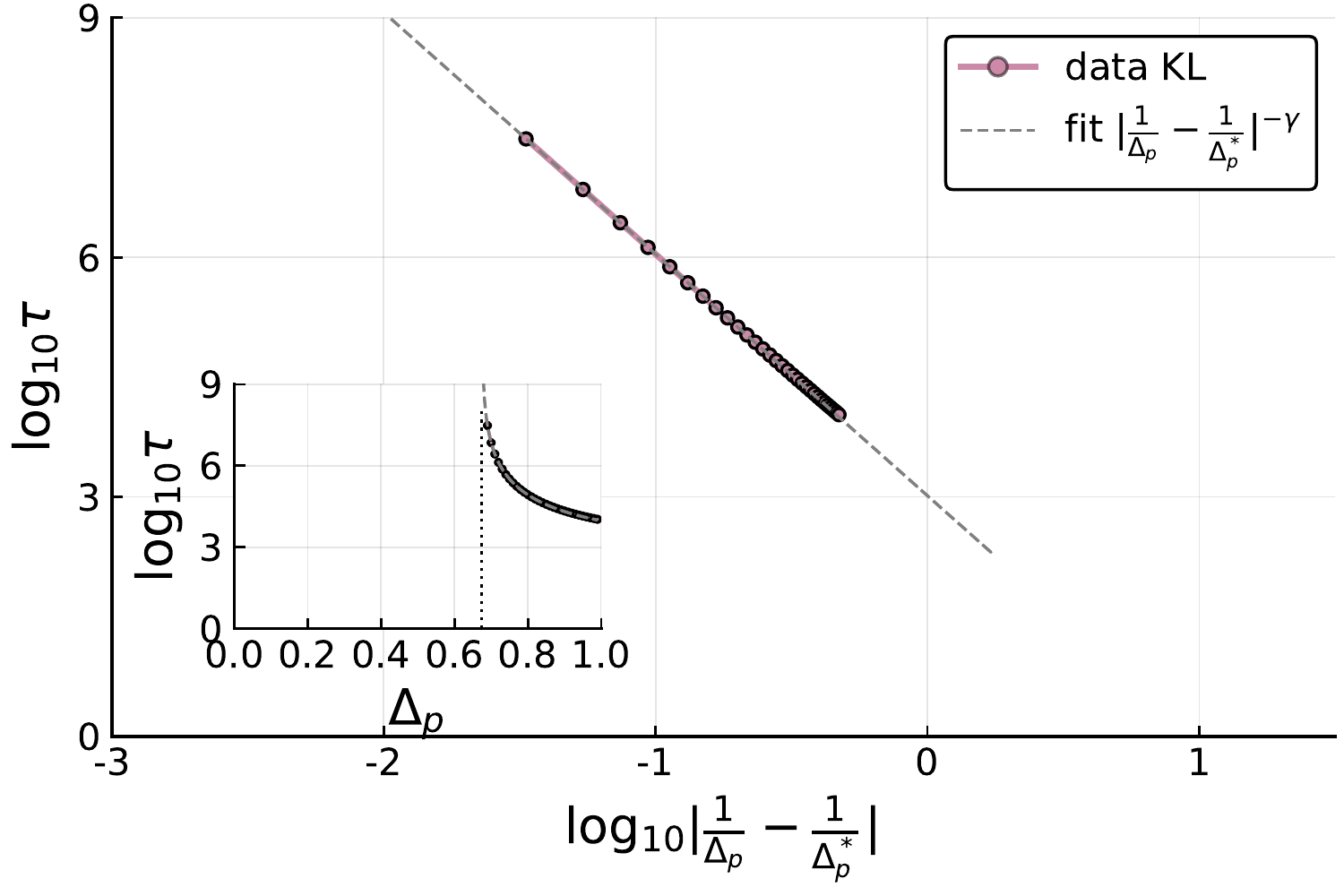}
		}
		\centering
		\subfigure[$\Delta_2=1.10$: estimation $0.59104$ (AMP $0.59034$)]{
		\includegraphics[scale=.57]{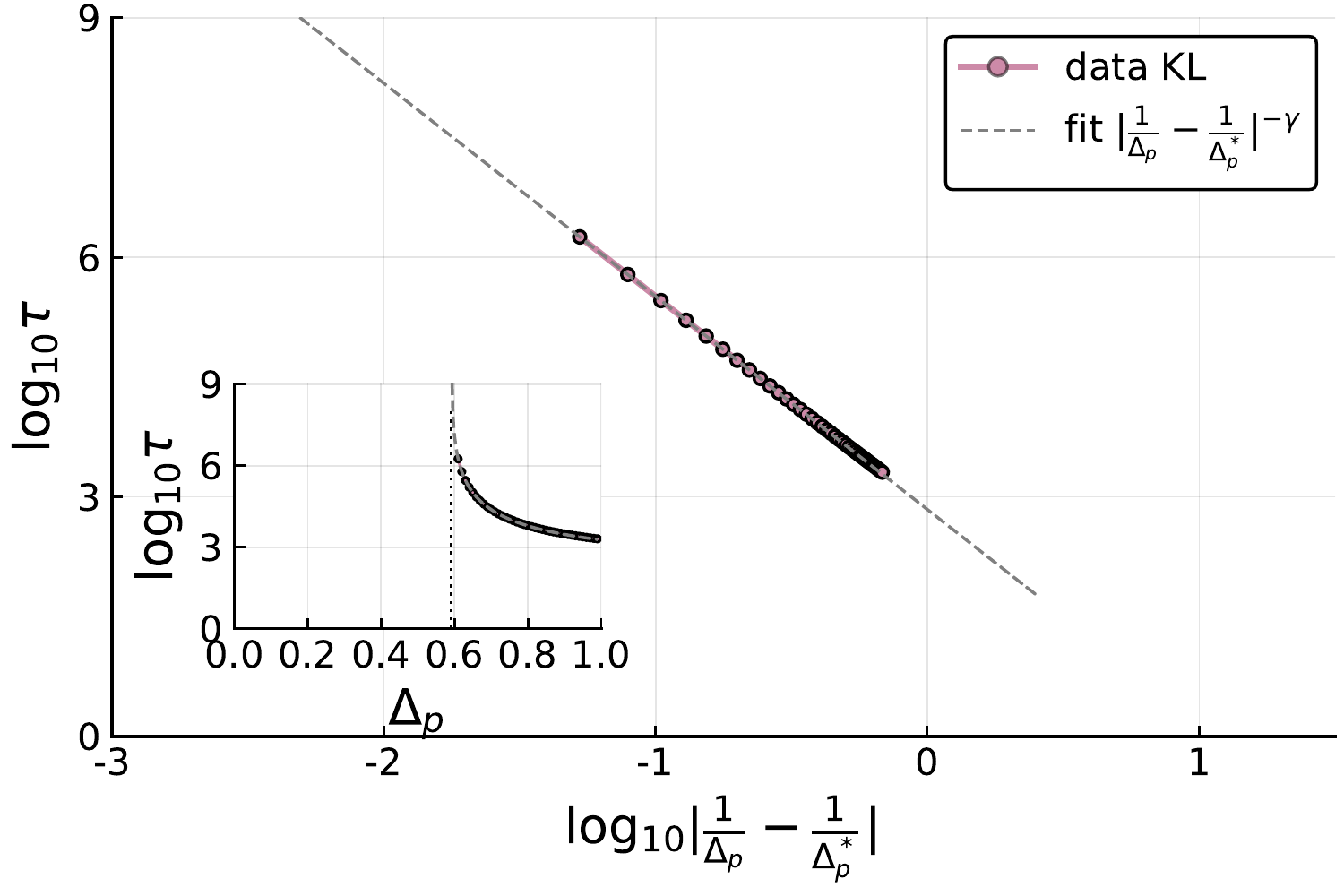}
		}
		\subfigure[$\Delta_2=2.00$: estimation  $0.34326$ (AMP $0.34314$)]{
		\includegraphics[scale=.57]{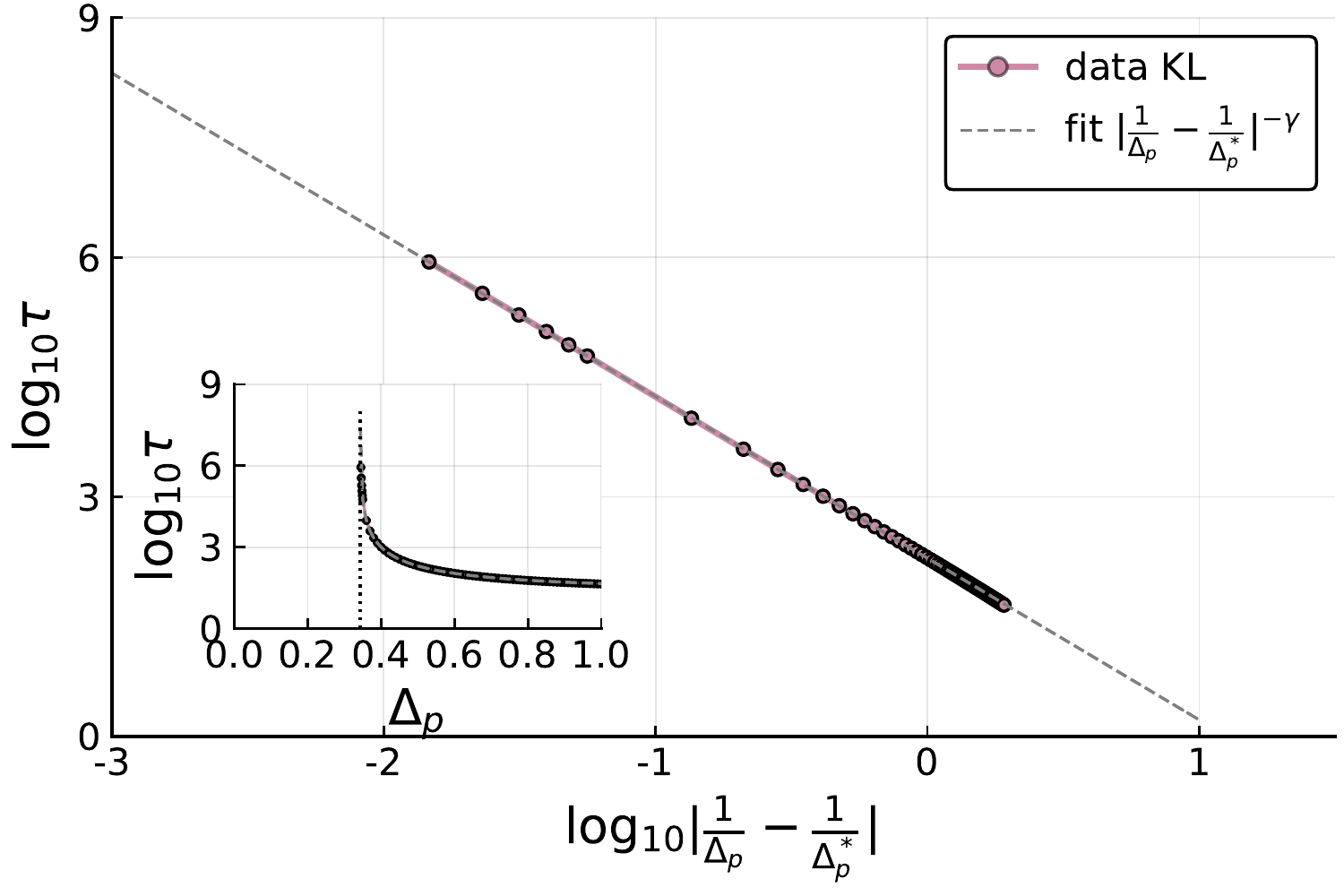}
		}
		\caption{Relaxation time obtained from the LSE starting from an informative initial condition $\Cmag_0=1$. The four cases refers to the $2+3$ model and are fitted with a power law and the relaxation time appears to diverge very close to point predicted by AMP (the AMP prediction is given in the captions).}
		\label{fig:fitDeltapSol}
	\end{figure}

	For the protocol to be successful it is therefore crucial that the
	annealing time $\tau_{ann}$ is large enough to give the possibility of
	magnetizing the solution before $\Delta_{p,{\rm eff}}$ has significantly decreased towards $\Delta_p$.
	According to this analysis, it is not surprising that in Fig.~\ref{fig:smartAnnealing} for $\Delta_2=0.90$ the proposed protocol seems not to be successful. For this value of the parameter $\Delta_2$, the time to find a solution even with $\Delta_p=100$ should be larger than $1000$ time units, which is much larger than the used $\tau_{ann}$ and anyway out of the time window of our numerical solution.
	However, with an annealing time large enough it would be in principle possible to recover exactly the same boundaries of the AMP easy region.

	\begin{figure}
		\centering
		\subfigure[$\Delta_2=0.50$, $\Delta_p=0.10$]{
		\includegraphics[scale=.49]{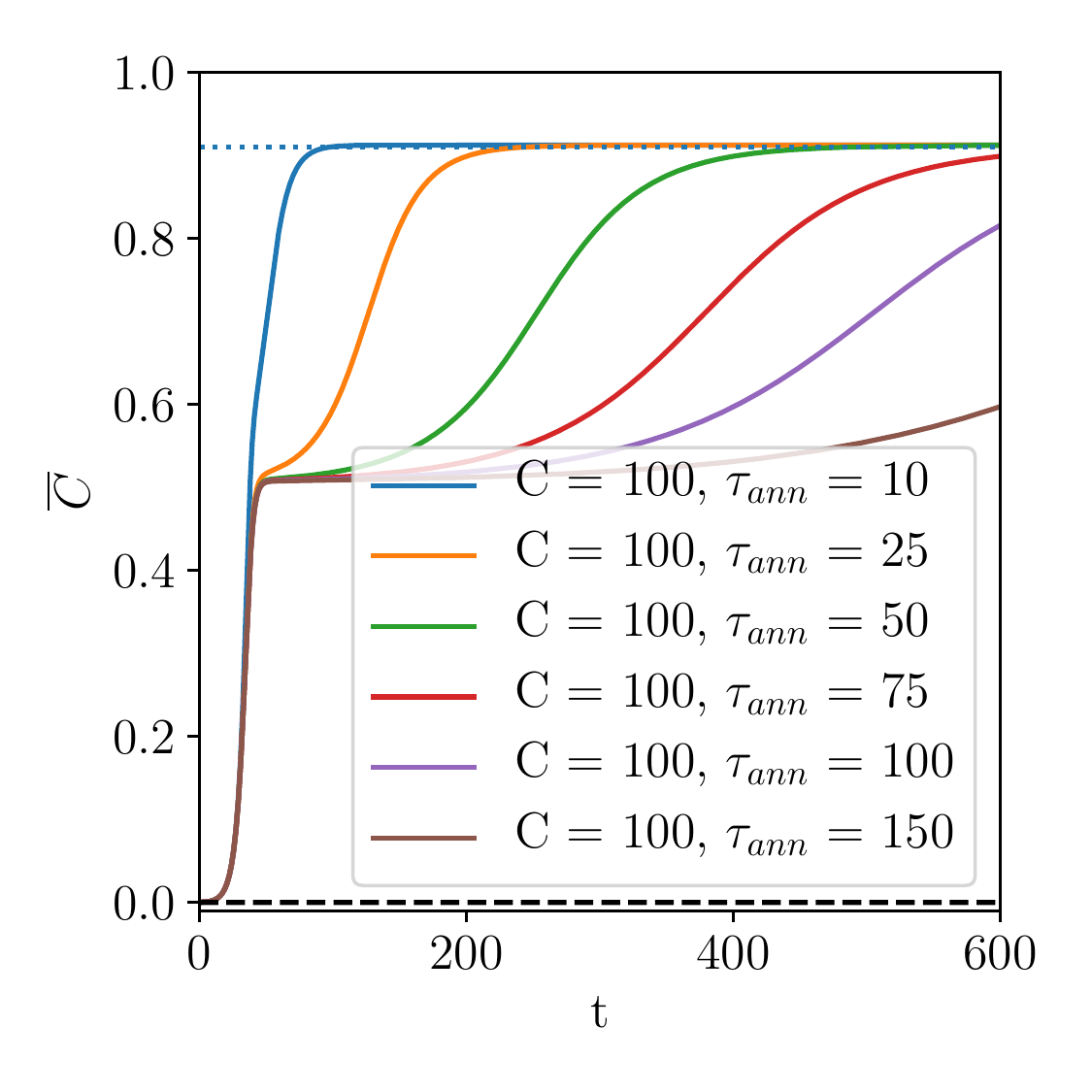}
		}
		\subfigure[$\Delta_2=0.50$, $\Delta_p=0.20$]{
		\includegraphics[scale=.49]{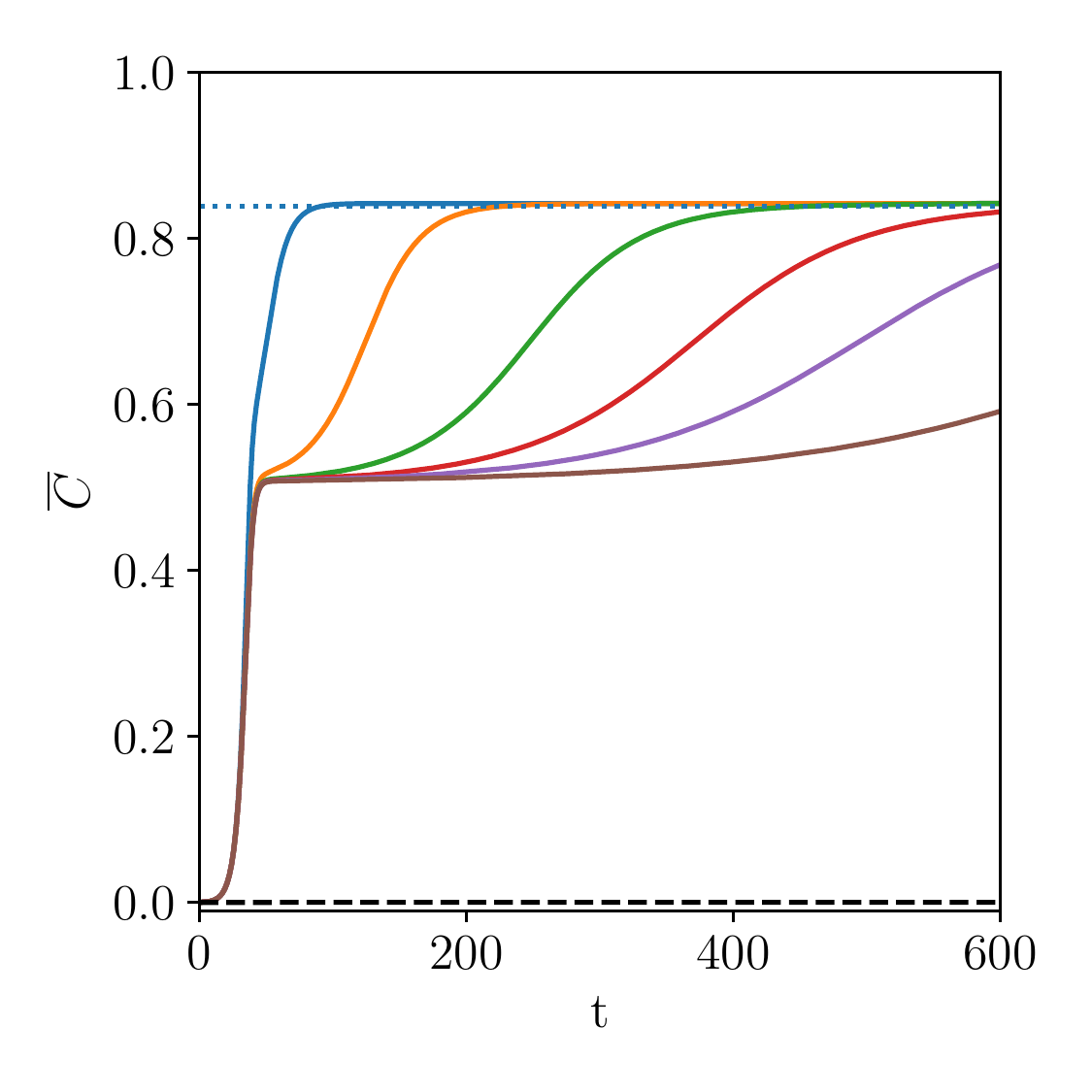}
		}
		\subfigure[$\Delta_2=0.50$, $\Delta_p=0.30$]{
		\includegraphics[scale=.49]{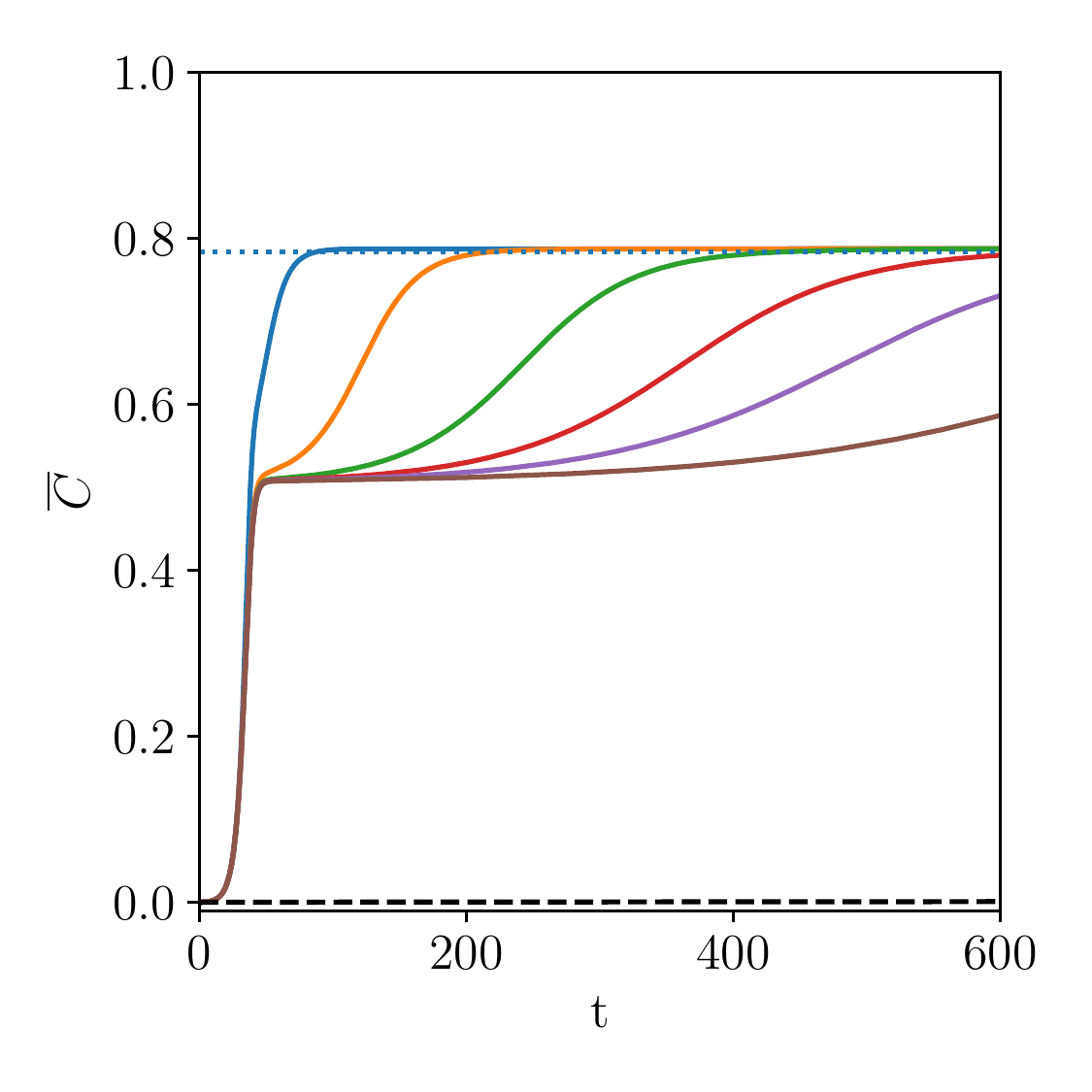}
		}
		\centering
		\subfigure[$\Delta_2=0.70$, $\Delta_p=0.10$]{
		\includegraphics[scale=.49]{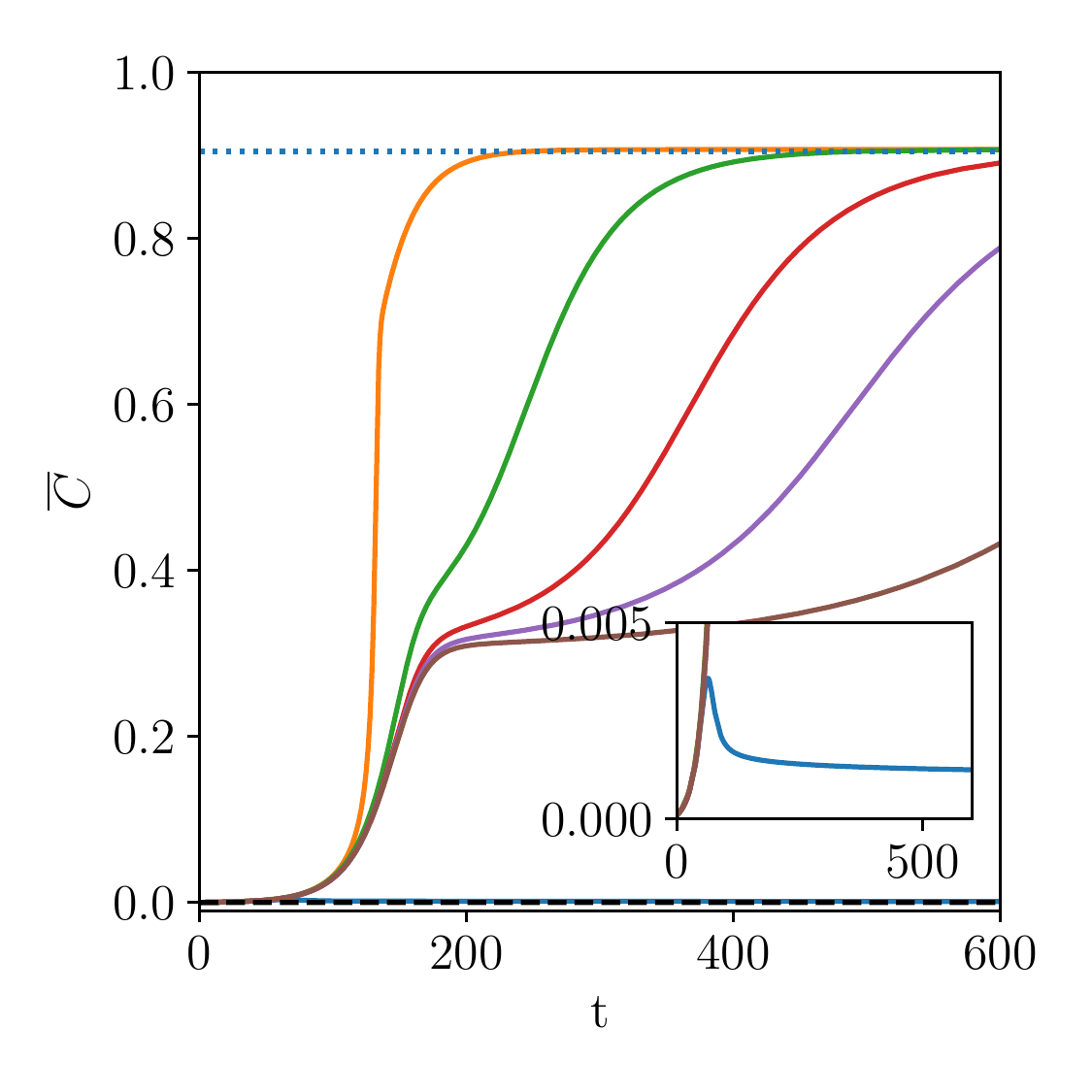}
		}
		\subfigure[$\Delta_2=0.70$, $\Delta_p=0.20$]{
		\includegraphics[scale=.49]{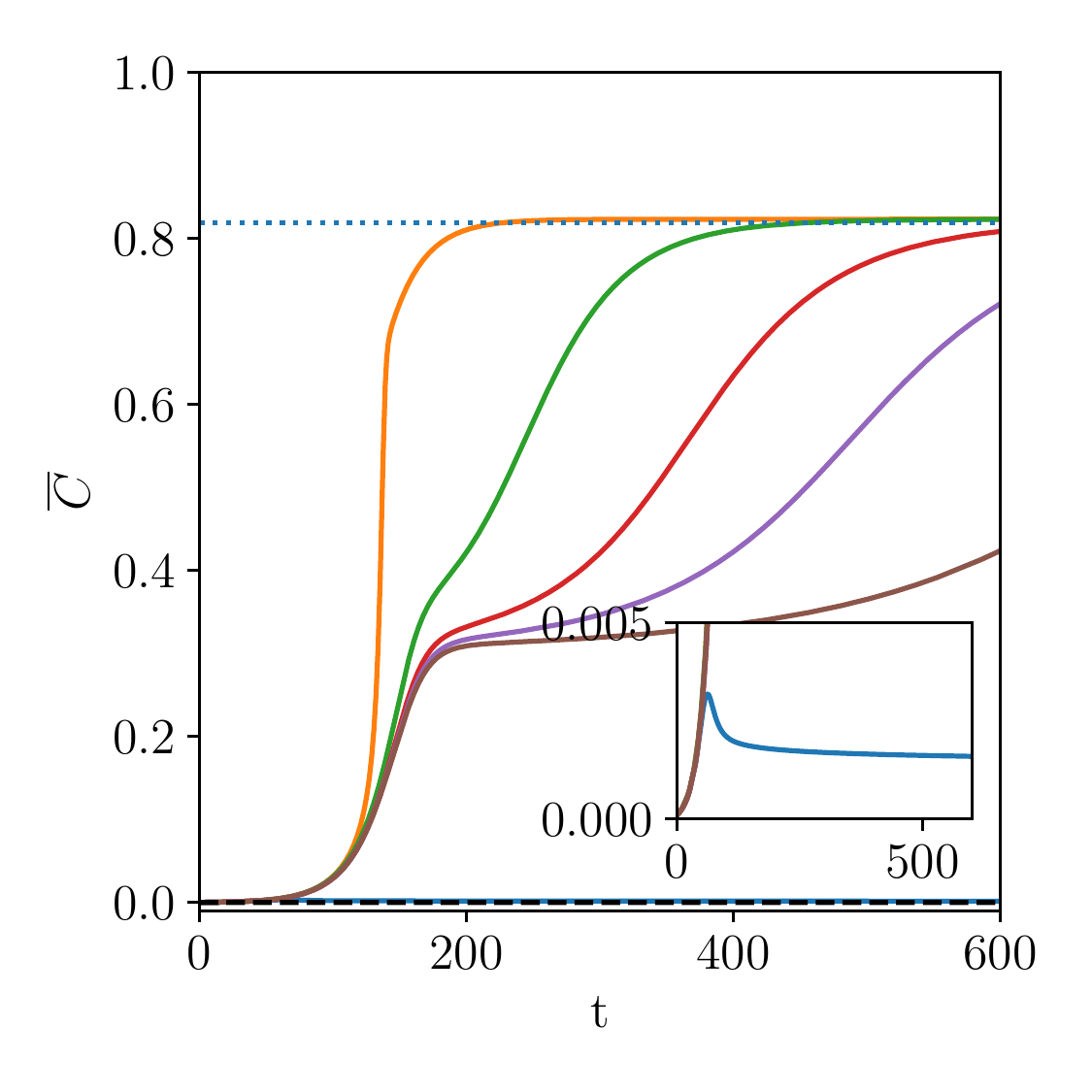}
		}
		\subfigure[$\Delta_2=0.70$, $\Delta_p=0.30$]{
		\includegraphics[scale=.49]{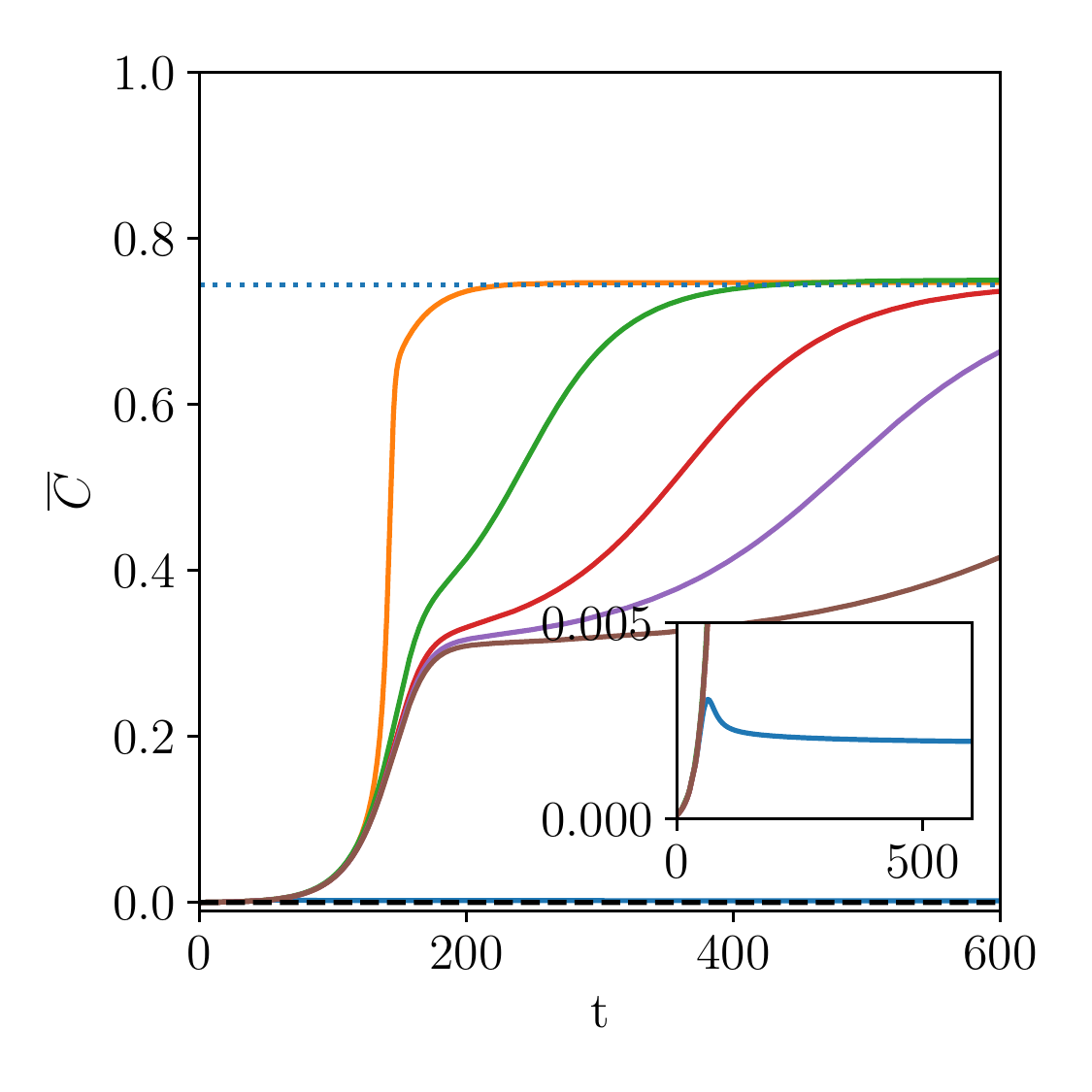}
		}
		\subfigure[$\Delta_2=0.90$, $\Delta_p=0.10$]{
		\includegraphics[scale=.49]{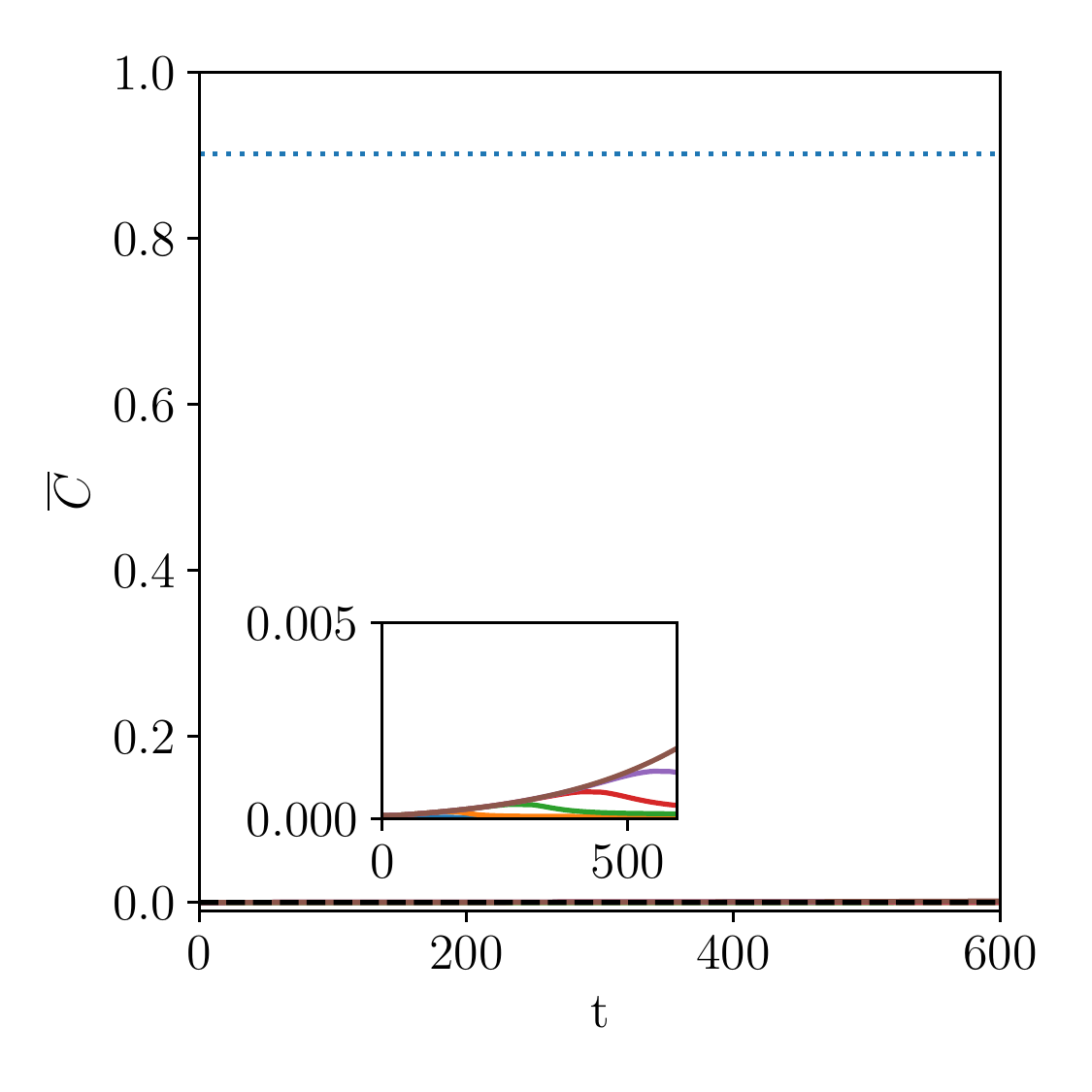}
		}
		\subfigure[$\Delta_2=0.90$, $\Delta_p=0.20$]{
		\includegraphics[scale=.49]{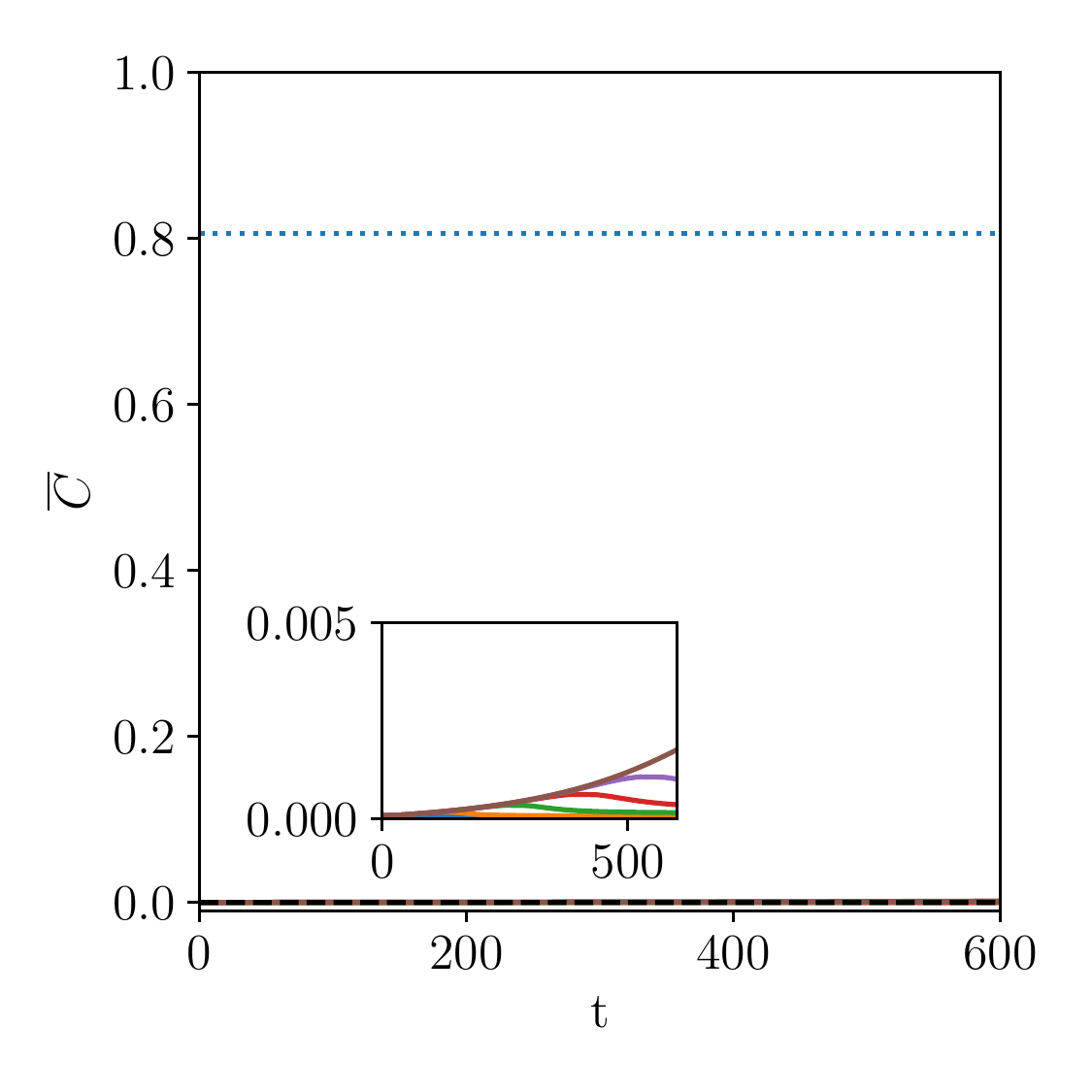}
		}
		\subfigure[$\Delta_2=0.90$, $\Delta_p=0.30$]{
		\includegraphics[scale=.49]{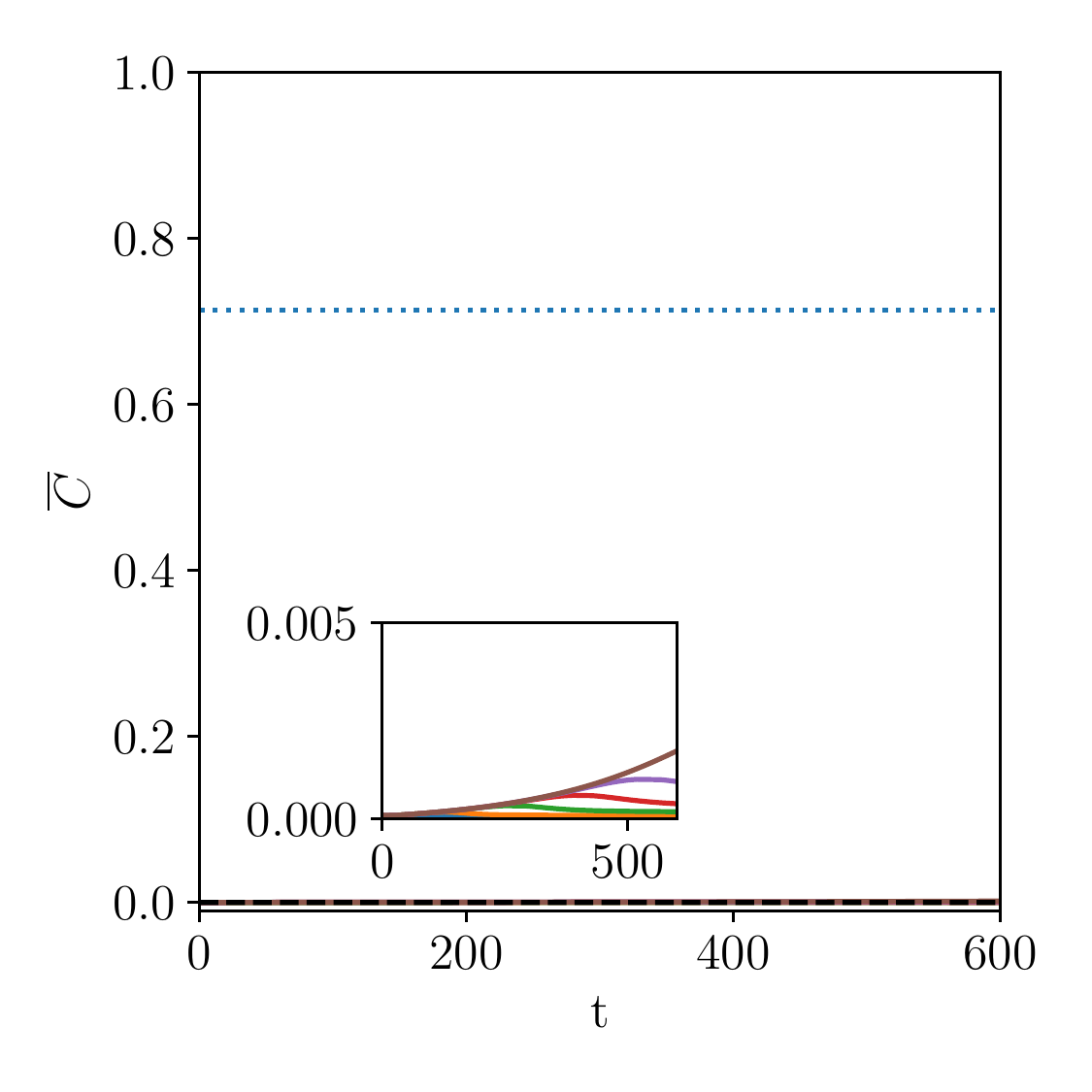}
		}
		\caption{The figures show the correlation with the
                  signal in time obtained using different annealing
                  protocols, whose details are reported in the legend
                  of the first figure. All the protocols have $C=100$
                  which means that all the dynamics start with close
                  effective $\Delta_p\sim100$,
                  $T_p(0)\Delta_p\simeq100$ and start with an initial
                  overlap $\Cmag_0=10^{-4}$. What changes among the
                  different lines is the relaxation speed, from the
                  fastest, drawn in blue, to the slowest, drown in
                  brown. They are compared with the asymptotic value
                  of AMP, dotted line, and the Langevin dynamics
                  without tensor-annealing, dashed line.}
		\label{fig:smartAnnealing}
	\end{figure}

	\begin{figure}
		\centering
		\subfigure[]{
		\includegraphics[scale=.4]{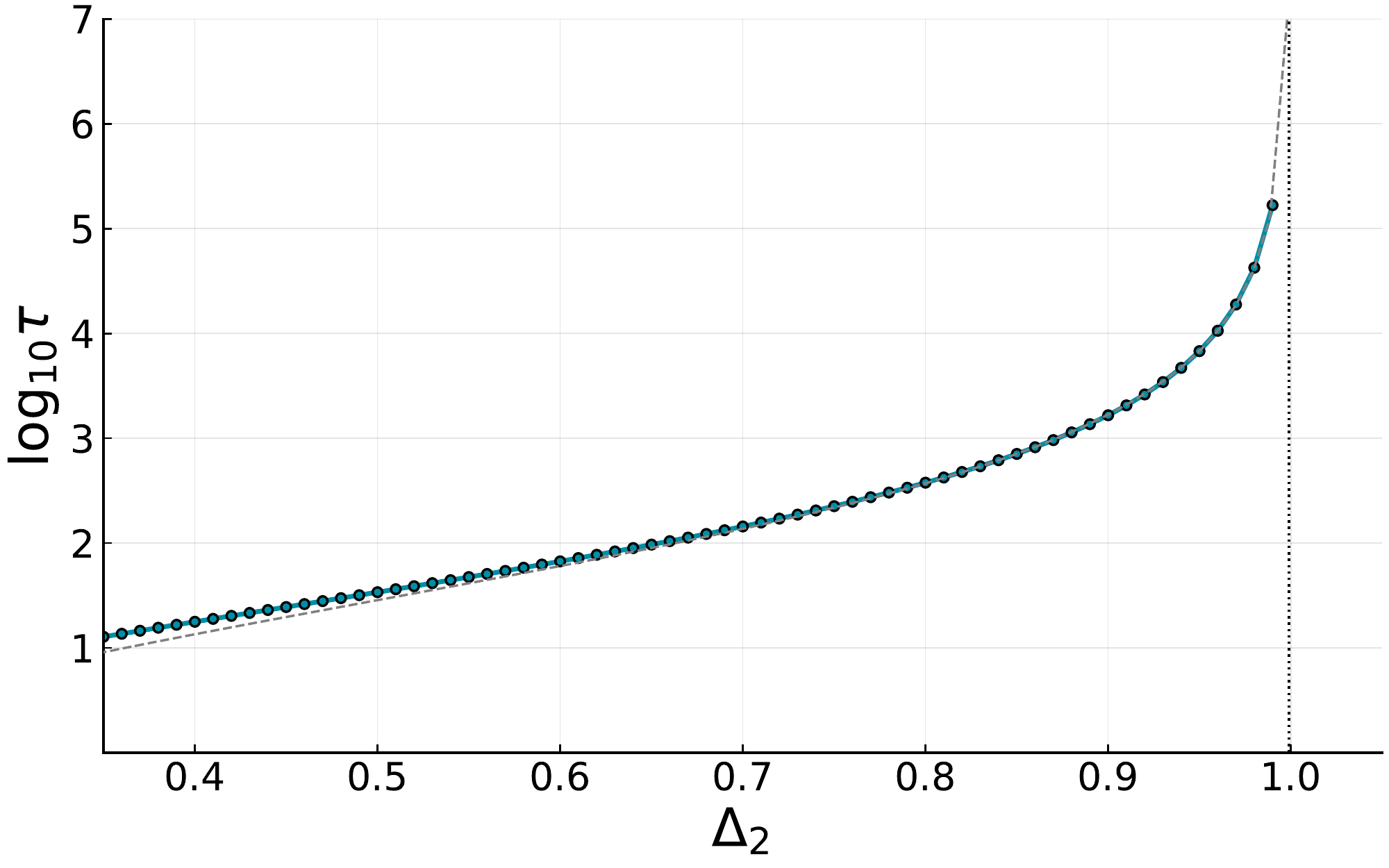}
		}
		\caption{Relaxation times of Langevin dynamics at $\Delta_p=100$ without using protocols starting from $\Cmag_0 = 10^{-4}$.}
		\label{fig:relaxationTimeDeltap100}
	\end{figure}

	\section{Glassy nature of the Langevin hard phase: the replica approach}\label{sec:replica analysis}

	In this section we study the landscape of the spiked matrix-tensor problem following the approach of \cite{antenucci2018glassy}.
	We underline here that we are interested in studying the free energy landscape problem rather than the energy landscape since
	the former is the relevant quantity for finite temperatures ($\beta=1$ in our case, as discussed in Sec.~\ref{sec:model}).
	The results of \cite{antenucci2018glassy} suggest that the AMP-hard phase
	and part of the AMP-easy phase are glassy. Therefore we could expect that low magnetization glassy states
	trap the Langevin algorithm and forbid the relaxation to the equilibrium configurations that surrounds the signal.
	This may happen also in a region where AMP instead is perfectly fine in producing configurations strongly correlated with the signal.
	In order to check this hypothesis we compute the
	logarithm of the number of glassy states, called the
	complexity by using the replica method \cite{monasson1995structural, antenucci2018glassy}.
	The goal of this analysis is to trace an additional
	line in the phase diagram that delimits the region where stable
	one step replica symmetry breaking (1RSB) metastable states
	exist. We conjecture that this provides a physical lower bound to the Langevin hard phase
	in the $(\Delta_p, 1/\Delta_2)$ phase diagram.

	\subsection{Computation of the complexity through the replica method}
       \label{sec:replicas_1RSB}
	The replica trick is based on the simple identity: $\mathbb{E}\log x =  \lim_{n\rightarrow0} \frac{\partial}{\partial n}\mathbb{E}x^n$.
	Using this observation we can compute the expected value of the free
	energy, $\Phi = - (\log Z)/N$, averaging the $Z^n$ and taking the limit
	$n\rightarrow0$. This is in general as difficult as the initial
	problem, however, if we consider only integer $n$ and extrapolate to
	$0$, the computation becomes much less involved due to the fact that for integer $n$ the average $\mathbb{E}x^n$ can be sometimes performed analytically.
	Indeed in this case the
	replicated partition function $Z^n$ can be regarded as the partition
	function of $n$ identical uncoupled systems or replicas. Averaging over the disorder we obtain a clean system of interacting replicas.
	The Hamiltonian of this system displays an emerging \emph{replica symmetry} since it is left unchanged by a permutation of replicas.
	This symmetry can be spontaneously broken in certain disordered models where frustration is sufficiently strong \cite{MPV87}.

	In mean field models characterized by fully connected factor graphs, the resulting Hamiltonian of interacting replicas
	depends on the configuration of the system only through a simple order parameter, the overlap $\tilde Q$ between them, which is a $n\times n$ matrix that describes the similarities
	of the configurations of different replicas in phase space.
	Furthermore the Hamiltonian is proportional to $N$ which means that in the thermodynamic limit $N\to \infty$,
	the model can be solved using the saddle point method.
	In this case one needs to consider a simple ansatz for the saddle point structure of the matrix $\tilde Q$ that allows to take the analytic continuation for $n\to 0$.
	The solution to this problem comes from spin glass theory and general details can be found in \cite{MPV87}.
	The saddle point solutions for $\tilde Q$ can be classified according to the replica symmetry breaking level
	going from the replica symmetric solution where replica symmetry is not spontaneously broken to various degree of spontaneous replica symmetry breaking
	(including full-replica symmetry breaking).
	Here we will not review this subject but the interested reader can find details in \cite{MPV87}.
	The model we are analyzing can be studied in full generality at any degree of RSB (see for example \cite{CL04, CL06, CL13} where the same models have been studied in absence of a signal).
	However here we will limit ourselves to consider saddle point solutions up to a 1RSB level.

	The complexity of the landscape can be directly related to replica symmetry breaking.
	A replica symmetric solution implies an ergodic free energy landscape characterized by a single pure state.
	When replica symmetry is broken instead, a large number of pure states arises and the phase space gets clustered in a hierarchical way \cite{MPV87}.
	Making a 1RSB approximation means to look for a situation in which the hierarchical organization contains just one level: the phase space gets clustered into an
	exponential number of pure states with no further internal structure.

	If we assume a 1RSB glassy landscape, we can compute the complexity of metastable states using a recipe due to Monasson \cite{monasson1995structural} (see also \cite{zamponi2010mean} for a pedagogical introduction).
	The argument goes as follows.

	Let us consider system with $x$ \emph{real} replicas infinitesimally coupled.
	If the free energy landscape is clustered into an exponential number of metastable states,
	the replicated partition function, namely the partition function of the system of $x$ real replicas, can be written as
	$$
		Z^x \simeq
		e^{N[\Sigma(f^*)- x \beta f^*]}
	$$
	where $f^*$ is the internal free energy of the dominant metastable states that is determined by the saddle point condition
	$\frac{d\Sigma}{df}(f^*)= \beta x$ and $\beta$ the inverse temperature. Note that since we are interested in the Bayes optimal case, this corresponds to set $\beta=1$. In the analysis we will consider a generic $\beta$ before taking the limits in order to derive the averaged energy by taking its derivative. 
	The function $\Sigma(f)$ is the complexity of metastable states having internal entropy $f$.
	Therefore, using the free parameter $x$ we can reconstruct the form of $\Sigma(f)$ from the replicated free energy.
	In order to compute the replicated free energy we need to apply the
	replica trick on the replicated system, $\overline{\log Z^x} =
		\lim_{n\rightarrow0} \frac{\partial}{\partial n}\overline{(Z^x)^n}$. Calling the
	replicated free energy $\Phi = -\frac1{N}\overline{\log Z^x}$, we get
	the complexity as $\Sigma = x \frac{\partial\Phi}{\partial x}-\Phi$.

	We can now specify the computation to our case where the partition function is the normalization of the posterior measure.
	With simple manipulations of the equations \cite{CC05}, the partition function can be expressed as the integral over the overlap matrix
	\begin{equation}
		\overline{(Z^x)^n} = \overline{Z_x^n} \propto \int \prod_{ab}dQ_{ab} e^{\beta NnxS(Q)} \simeq \limsup_Q e^{\beta NnxS(Q)}\,;
	\end{equation}
	where the overlap $Q$ is a $(nx+1)\times (nx+1)$ matrix
	\begin{equation*}
		Q =
		\left(
		\begin{array}{c|c}
				1      & m \cdots m                                  \\ \hline
				m      & \raisebox{-15pt}{{\huge\mbox{$\tilde{Q}$}}} \\[-4ex]
				\vdots &                                             \\[-0.5ex]
				m      &
			\end{array}
		\right)
	\end{equation*}
	that contains a special row and column that encodes the overlap between different replicas with the signal
	and therefore the corresponding overlap is the \emph{magnetization} $m$.

	The 1RSB structure for the matrix $\tilde Q$ can be obtained by defining the following $nx\times nx$ matrices: the identity matrix $\mathbb{I}_{ij} = \delta_{ij}$, the full matrix $\mathbb{J}^{(0)}_{nx,ij} = 1$, and $\mathbb{J}_{nx}^{(1)} = \text{diag}(J_{x}^{(0)},\dots,J_{x}^{(0)})$ a block diagonal matrix where the diagonal blocks $J_{x}^{(0)}$ have size $x\times x$ and are matrices full of 1. In this case the 1RSB ansatz for $\tilde Q$ reads
	\begin{equation*}
		\tilde{Q}=(1-q_M)\mathbb{I}_{nx}+(q_M-q_m)\mathbb{J}_{nx}^{(1)}+q_m\mathbb{J}_{nx}^{(0)}\,.
	\end{equation*}
	Using this ansatz we can compute $S(Q)$ that is given by
	\begin{equation}\label{eq:1RSB action}
		\begin{split}
			\beta S(Q) &= \frac1{nx}\left[\frac12\log\text{det}\;Q + \frac{\beta^2}{2p\Delta_p}\sum_{a,b=1}^n Q_{ab}^p + \frac{\beta^2}{4\Delta_2}\sum_{a,b=1}^n Q_{ab}^2 + \frac{\beta}{p\Delta_p}\sum_{a=1}^n Q_{0a}^p + \frac{\beta}{2\Delta_2}\sum_{a=1}^n Q_{0a}^2\right]=\\
			&= \frac12\log(1-q_M)+\frac1{2x}\log\frac{1-q_M+x(q_M-q_m)}{1-q_M}+\frac12\frac{q_m-m^2}{1-q_M+x(q_M-q_m)}+\\
			&+ \frac{\beta^2}{2p\Delta_p}\left(1-q_M^p+x(q_M^p-q_m^p)+\frac2\beta m^p\right)+\frac{\beta^2}{4\Delta_2}\left(1-q_M^2+x(q_M^2-q_m^2)+\frac2\beta m^2\right)\,.
		\end{split}
	\end{equation}
	From Eq.~\eqref{eq:1RSB action} we obtain the saddle point equations
	\begin{equation}\label{eq:Saddle Point complexity}
		\begin{split}
			& 0 = 2\frac{\partial S}{\partial q_M} = (x-1)\left[\frac1x\left(\frac1{1-q_M+x(q_M-q_m)}-\frac1{1-q_M}\right)-\frac{q_m-m^2}{[1-q_M+x(q_M-q_m)]^2}+\beta^2\left(\frac{q_M^{p-1}}{\Delta_p}+\frac{q_M}{\Delta_2}\right)\right]\,; \\
			& 0 = 2\frac{\partial S}{\partial q_m} = x \left[\frac{q_m-m^2}{[1-q_M+x(q_M-q_m)]^2}-\beta^2\left(\frac{q_m^{p-1}}{\Delta_p}+\frac{q_m}{\Delta_2}\right)\right]\,; \\
			& 0 = \frac{\partial S}{\partial m} = \frac{-m}{1-q_M+x(q_M-q_m)}+\beta^2\left(\frac{m^{p-1}}{\Delta_p}+\frac{m}{\Delta_2}\right)\,.
		\end{split}
	\end{equation}

The above 1RSB fixed point equations can be used to derive the de
Almeida-Thouless instability of the RS solution towards 1RSB. This
stability condition, sometimes called {\it the replicon} also
determines the overlap of the marginal threshold states. The stability
analysis is done by expansion of eqs.~(\ref{eq:Saddle Point
  complexity}) in a small parameters $q_M - q_m = \varepsilon \ll 1$
and investigating whether under iterations such a small difference
grows of decreases. This leads directly to the threshold condition on the
overlap
\begin{equation}
  \frac{1}{ \beta^2 (1 - q^{\rm th})^2 }= (p-1)\frac{(q^{\rm th})^{(p-2)}}{\Delta_p}
   + \frac{1}{\Delta_2}   \label{eq:replicon} \, . 
\end{equation}
This condition is then used in the derivation of the Langevin
threshold (\ref{eq:threshold}) in the main text. 

	From Eq.~\eqref{eq:1RSB action} we obtain also the averaged energy
	\begin{equation}\label{eq:energy_1RSB}
		E = \frac{\partial \overline{\log Z^x}}{\partial \beta}\Big|_{\beta=1} = \frac{1-q_M^p+x(q_M^p-q_m^p)+m^p}{p\Delta_p}+\frac{1-q_M^2+x(q_M^2-q_m^2)+m^2}{2\Delta_2}\,.
	\end{equation}
	In particular the threshold states are characterized by $q_M = q^\text{\rm th}$, fixed by Eq.~\eqref{eq:replicon}, $q_m=0$ and $m=0$. Imposing these values, we can use the saddle point equation for $q_M$, Eq.~\ref{eq:Saddle Point complexity}, to fix the Parisi parameters $x$, 
\begin{equation}
	x(q^\text{\rm th}) = \frac1{(1-q^\text{\rm th})\left[\frac{(q^\text{\rm th})^{p-1}}{\Delta_p} + \frac{q^\text{\rm th}}{\Delta_2}\right]} - \frac1{q^\text{\rm th}} + 1\;. 
\end{equation}
These pieces together give Eq.~\eqref{eq:energy_th} showed in the main text.	
	
	Having obtained the energy we can consider $\beta=1$ fixed for the rest of the analysis. We can observe that starting from
	this expression we can derive the RS free energy, \eqref{eq:free
		entropy generic}, $q_M =q_m$ or equivalently in the limit
	$x\rightarrow1$.
	The low magnetization solution to these equations gives the complexity of the metastable branch of the posterior measure which is given by
	\begin{equation}
		\begin{split}
			-\Sigma(x;Q^*) & = -\frac12\log\frac{1-q_M+x(q_M-q_m)}{1-q_M}+\frac{x}2\frac{q_M-q_m}{1-q_M+x(q_M-q_m)}-\frac{x^2}2\frac{(q_m-m^2)(q_M-q_m)}{[1-q_M+x(q_M-q_m)]^2}+\frac{x^2}2\frac{q_M^p-q_m^p}{p\Delta_p}+\\
			& +\frac{x^2}2\frac{q_M^2-q_m^2}{2\Delta_2}\,.
		\end{split}
	\end{equation}
	The free parameter $x$ allows us to tune the free energy of the states
	of which we compute the complexity.
	Thus we can characterize the part of the phase diagram where
        an exponential number of states is present.

	To complete the 1RSB analysis we compute the stability of the
        1RSB saddle point solution for $Q$. This is done analogously
        to the derivation of the \emph{replicon} condition
        (\ref{eq:replicon}), analyzing stability of the 1RSB towards
        further replica symmetry breaking.
	Following \cite{crisanti1992sphericalp, CL13} we obtain two replicon eigenvalues given by
	\begin{align}
		\label{eq:replicon I}
		\lambda_I    & =
		1-(1-q_M+x(q_M-q_m))^2\left[(p-1)\frac{q_m^{p-2}}{\Delta_p}+\frac1{\Delta_2}\right]\,
		,                \\
		\label{eq:replicon II}
		\lambda_{II} & =
		1-(1-q_M)^2\left[(p-1)\frac{q_M^{p-2}}{\Delta_p}+\frac1{\Delta_2}\right]
		\, .
	\end{align}

	\begin{figure}
		\centering
		\includegraphics[scale=.5,clip]{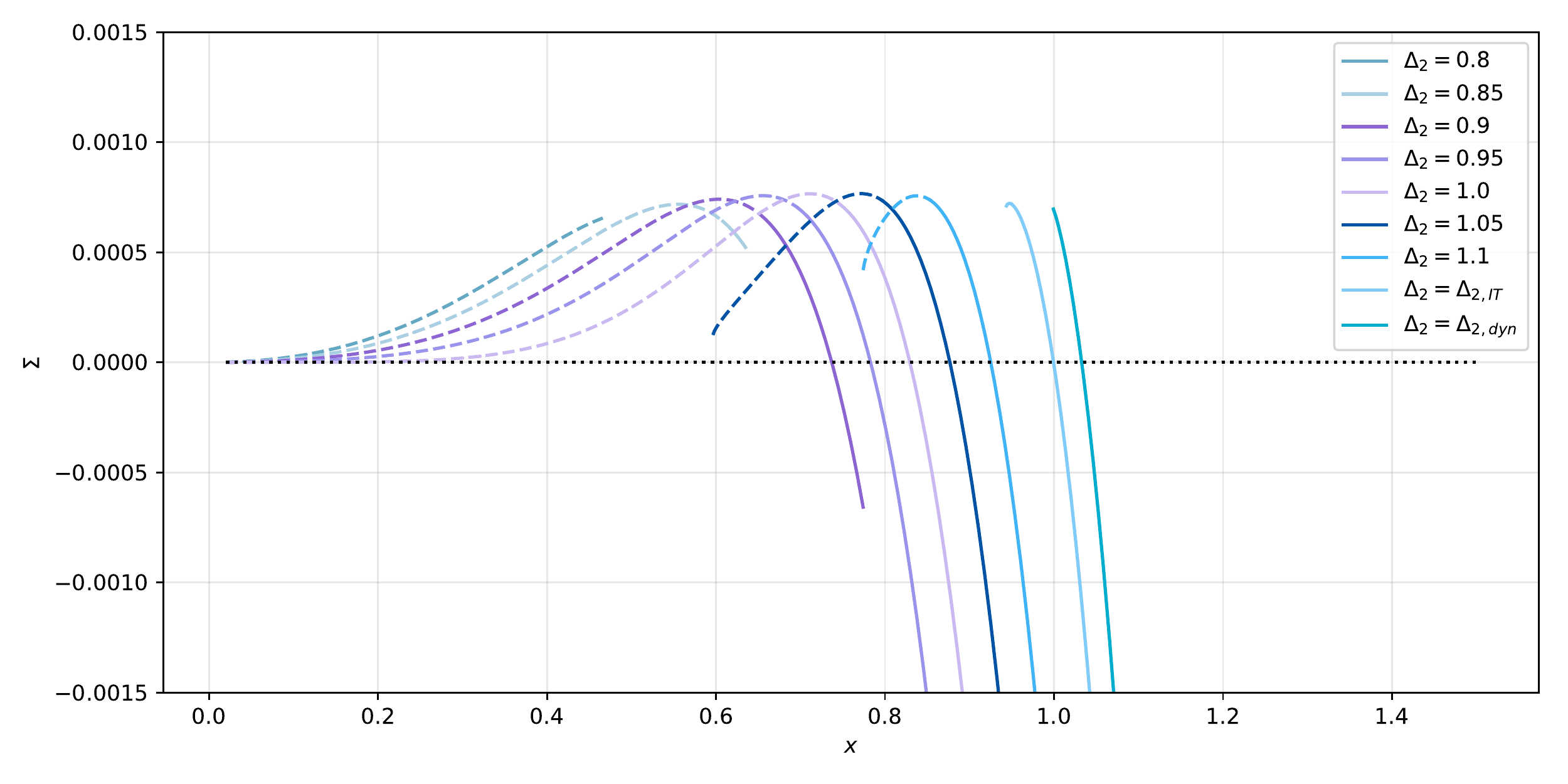}
		\caption{Complexity with as a function of the Parisi parameter $x$ for $p=3$ on the line $\Delta_p=0.5$. The solid line characterizes the stable part of the complexity while the dashed line the unstable one.}
		\label{fig:complexity p3 Deltap0.50}
	\end{figure}

	\begin{figure}
		\centering
		\subfigure[$p=3$; $\Delta_p = 0.50$; $\Delta_2 = 0.85$]{
		\includegraphics[scale=.4]{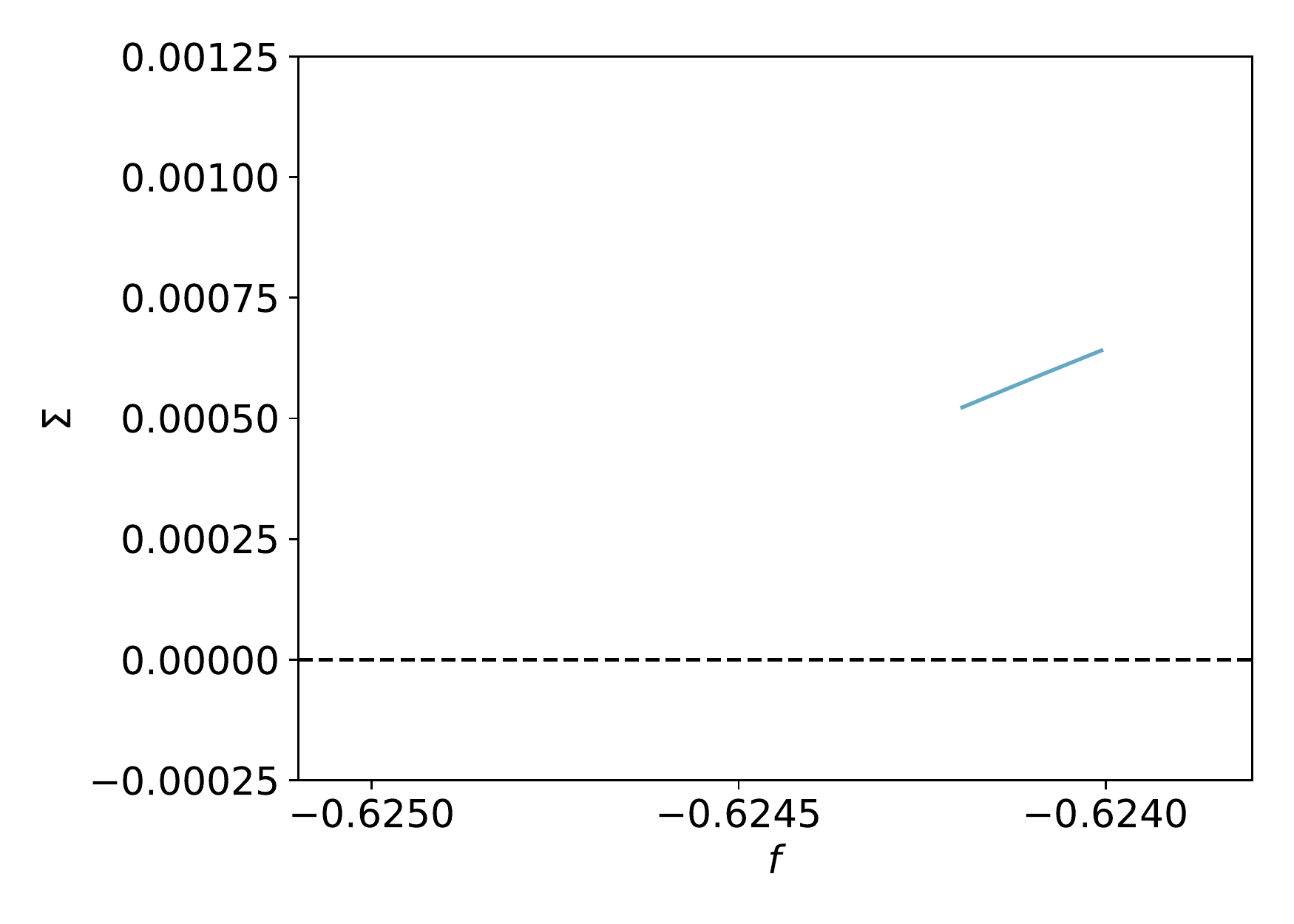}
		}
		\subfigure[$p=3$; $\Delta_p = 0.50$; $\Delta_2 = 0.90$]{
		\includegraphics[scale=.4]{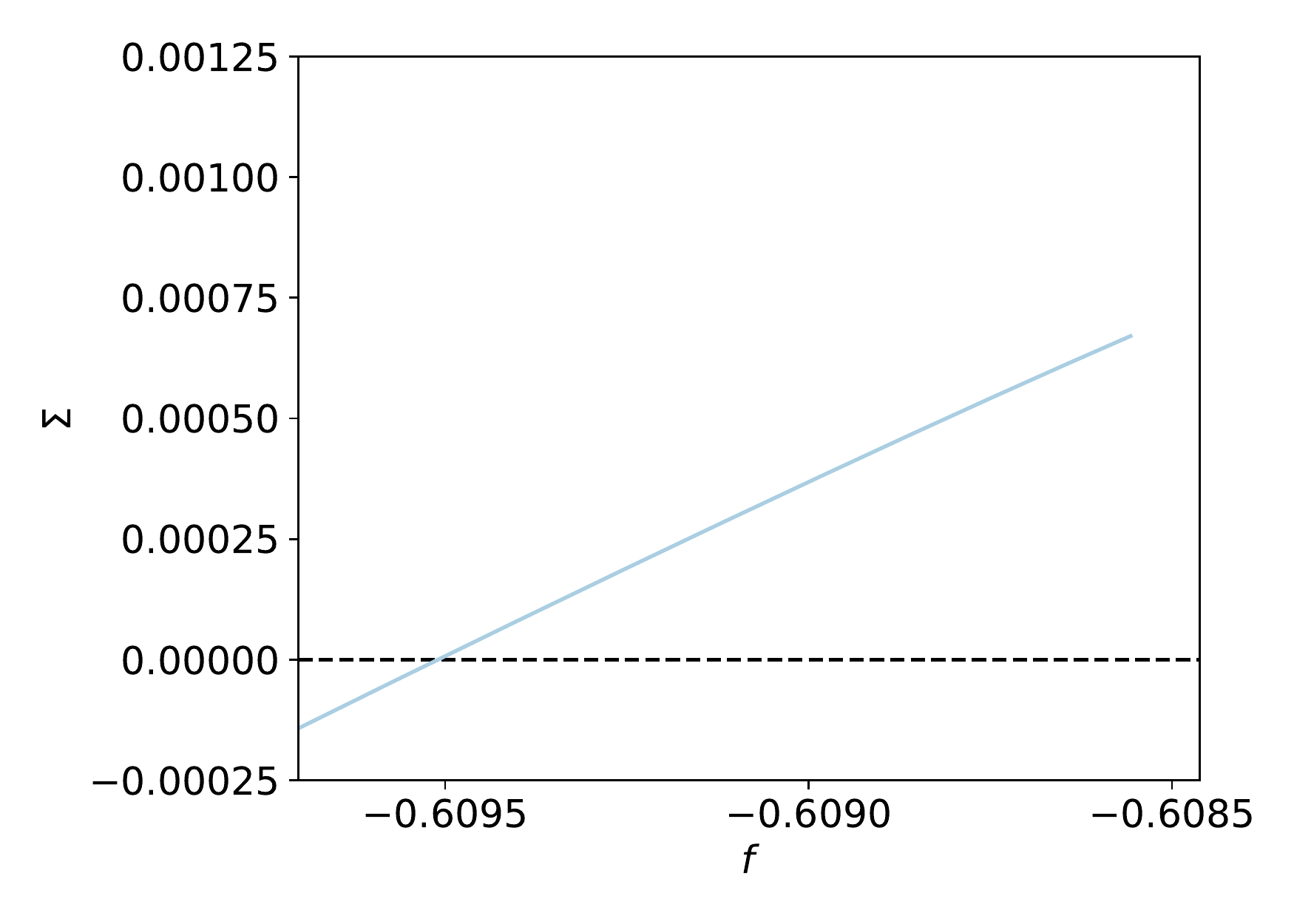}
		}\\
		\subfigure[$p=3$; $\Delta_p = 0.50$; $\Delta_2 = 0.95$]{
		\includegraphics[scale=.4]{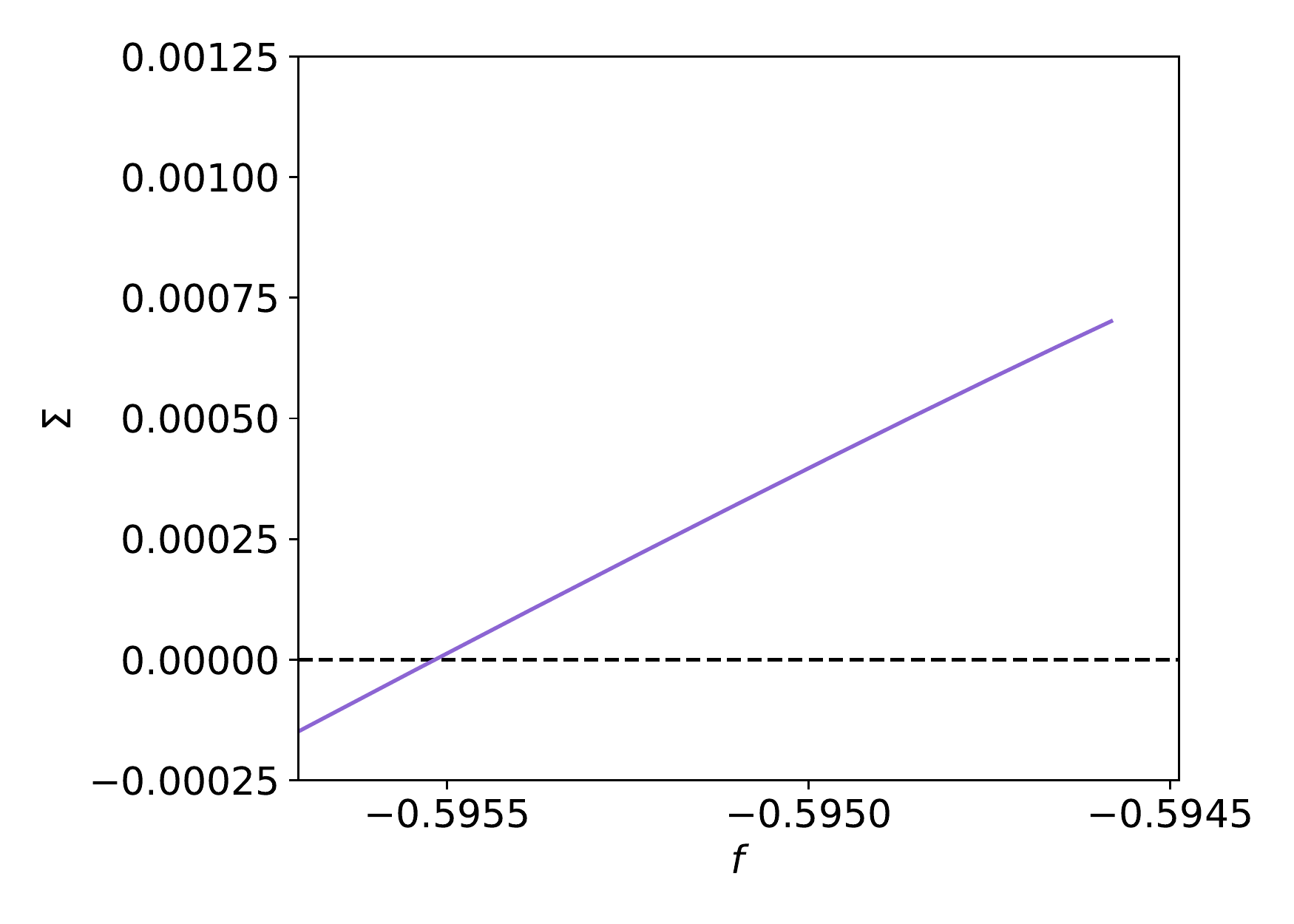}
		}
		\subfigure[$p=3$; $\Delta_p = 0.50$; $\Delta_2 = 1.00$]{
		\includegraphics[scale=.4]{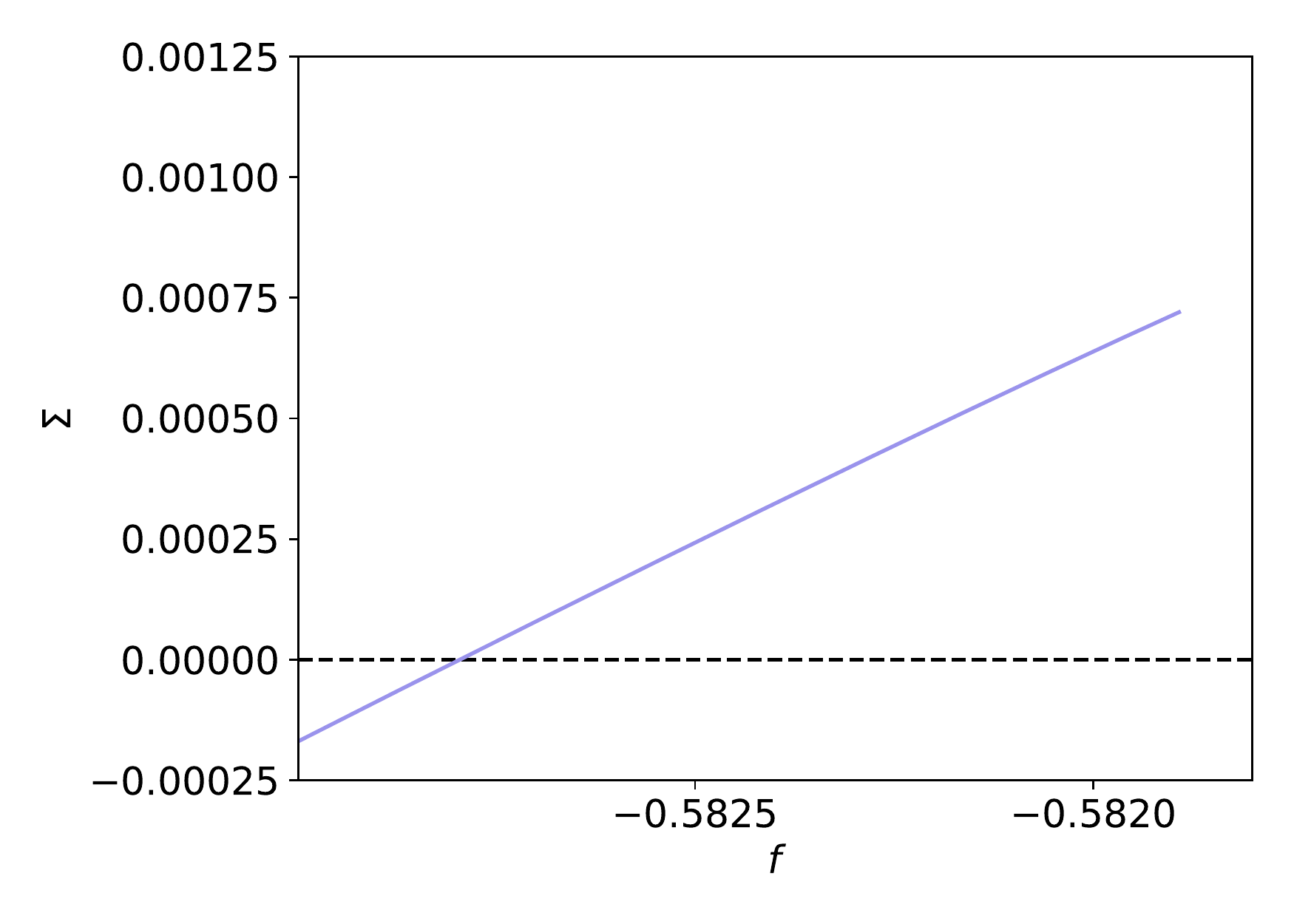}
		}
		\caption{The stable part of the 1RSB complexity as
			a function of the free energy for $p=3$ and
			$\Delta_p=0.5$.}
		\label{fig:complexity f* p3 Deltap0.50}
	\end{figure}

	We can analyze what happens to the landscape when we fix
        $\Delta_p<1$ and we start from a large value of
        $\Delta_2<\Delta_{2,{\rm dyn}}(\Delta_p)$ and we decrease $\Delta_2$.
	In this case for sufficiently high $\Delta_2$ and large enough $\Delta_p$ the system is in a paramagnetic phase and no glassy states are present.
	At the dynamical transition line instead we find a positive complexity as plotted in Fig.~\ref{fig:complexity p3 Deltap0.50}. At this point the equilibrium states that dominate the posterior measure are the
	so called \emph{threshold} states for which the complexity is maximal. For those states the eigenvalue $\lambda_{II}=0$ which confirms that these states are marginally stable \cite{cugliandolo1993analytical}.
	Decreasing $\Delta_2$ one crosses the information theoretic phase transition where the relevant metastable states that dominate the posterior measure have zero complexity.
	This corresponds to a freezing/condensation/Kauzmann transition. Below the information theoretic phase transition the thermodynamics of the posterior measure is dominated
	by the state containing the signal. However one can neglect the high magnetization solution of the 1RSB equations to get the properties of the metastable branch and computing the
	complexity of states that have zero overlap with the signal.
	The complexity curves as a function of the Parisi parameter $x$ for decreasing values of $\Delta_2$ are plotted in Fig.~\ref{fig:complexity p3 Deltap0.50} for fixed $\Delta_p=0.5$ and several $\Delta_2$.
	The curves contain a stable 1RSB part and an unstable one where $\lambda_{II}$ is negative.
	The 1RSB line shown in Figs.~\ref{fig:phase diagram p=3_all} is obtained by looking at when the states with positive complexity and $\lambda_{II}=0$ disappear.
	This means that it gives the point where the 1RSB marginally stable states disappear and therefore it is expected to be a lower bound for the disappearance of glassiness in the phase diagram.
	The important outcome of this analysis is that for $\Delta_2<1$ but not sufficiently small, namely in part of the AMP-easy phase, the replica analysis predicts the existence of 1RSB marginally stable glassy states that may trap the Langevin algorithm from relaxing towards the signal \cite{antenucci2018glassy}
	and therefore supports the existence of the Langevin hard
        phase. This approach, however, does not predict
          quantitatively correctly the extent on the Langevin-hard
          phase for reasons that remain obscure and should be
          investigated further.  

	Finally in Fig.~\ref{fig:complexity f* p3 Deltap0.50} we plot the complexity as a function of the internal free energy of the metastable states for some values of $\Delta_2$ and $\Delta_p$.

	\subsection{Breakdown of the fluctuation-dissipation theorem in the Langevin hard phase}
	When the Langevin algorithm is able to reach equilibrium, being it the signal or the paramagnetic state, it should satisfy the Fluctuation-Dissipation
	Theorem (FDT) according to which the response function is related to the correlation function through $R(t,t') = -\frac{\partial C(t,t')}{\partial t}$.
	Furthermore, time translational invariance (TTI) should arise
        implying that both correlation and response functions should
        be functions of only the time difference meaning that
        $R(t,t')=R(t-t')$ and $C(t,t')=C(t-t')$, note that all one time quantities are constant in equilibrium.
	When the dynamics is run in the glass phase, metastable states may forbid equilibration.
	In this case time translational invariance is never reached at
        long times, it is supposed to be reached only on exponential timescales in the system size, and the dynamics displays aging violating at the same time the FDT relation.
	The analysis of the asymptotic aging dynamics has been cracked by Cugliandolo and Kurchan in \cite{cugliandolo1993analytical, CK95} (see also \cite{Cu03} for a pedagogical review) in the simplest spin glass model (see also \cite{CK94SK} for a much more complex situation) where no signal is present.
	The outcome of this work is that when the dynamics started from a random initial conditions is run in the glass phase, it drives the system to surf on the \emph{threshold} states.
	In the model analyzed in \cite{cugliandolo1993analytical} these states correspond to the 1RSB marginally stable glassy states that maximize the complexity.
	In this section we analyze the Cugliandolo-Kurchan scenario by contrasting the numerical solution of the dynamical equations with the replica analysis of the complexity.
	According to \cite{cugliandolo1993analytical}, the long time
        Langevin dynamics, but still for times that are not
        exponentially large in the system size $N$, can be characterized by two time regimes.
	For short times differences $t-t'\sim {\cal O}(1)$ and $t'\to \infty$, the system obeys the FDT theorem and TTI; this regime can be understood as a first fast local equilibration in the nearest metastable state available.
	On a longer timescale $t-t'\to \infty$ and $t/t'<\infty$, the dynamics surfs on threshold states and FDT and TTI are both violated.
	In this time window, both the response and correlation
        functions become functions of $\lambda=h(t)/h(t')$ being
        $h(t)$ an arbitrary reparametrization of the time variable. The function $h(t)$ must be a monotonously increasing function. The asymptotic reparametrization invariance is a key property of the dynamical equations \cite{cugliandolo1993analytical}. By defining ${\cal C(\lambda)} = C(t,t')$ and ${\cal R(\lambda)} = tR(t,t') $ the Cugliandolo-Kurchan solution implies that in this aging regime the FDT relation can be generalized to
	\begin{equation}
		\label{eq:generalized FDT}
		\mathcal{R}\left(\lambda\right) = x\    \mathcal{C}'\left(\lambda\right)
	\end{equation}
	with $x$ an \emph{effective} FDT ratio that controls how much the FDT is violated.
	In the scenario of \cite{cugliandolo1993analytical}, the value of $x$ coincides with the 1RSB Parisi parameter that corresponds to threshold states computed within the replica approach.
	In order to test this picture we follow Cugliandolo and Kurchan \cite{CK97} and we plot the integrated response ${\cal {F}}(t,t') = -\int_{t'}^t R(t,t'')dt''$ as a function of $C(t,t')$ in a parametric way.
	This is done in Fig.~\ref{fig:generalized FDT fit}.

	If FDT holds at all timescales, one should see a straight line with slope $-1$.
	Instead what we see in the Langevin hard phase is that for large values of $t'$ the curves approach asymptotically for $t'\gg 1$ two straight lines.
	For high values of $\cal C$, meaning for short time differences $t-t'\sim {\cal O}(1)$, the slope of the straight line is $-1$ which means that ${\cal F}=1-{\cal C}$ as implied by the short time FDT relation.
	On longer timescales FDT is violated, confirming the glassiness of the Langevin hard phase.
	By doing a linear fit we can use the data plotted in Fig.~\ref{fig:generalized FDT fit} to estimate the FDT ratio $x$ appearing in Eq.~(\ref{eq:generalized FDT}). This can be compared with the Parisi parameter $x$ for which we have marginally stable 1RSB states.
	We find an overall very good agreement (data coming from the fit is reported in the caption of Fig.~\ref{fig:generalized FDT fit}).
	The small discrepancy between the two values of $x$ can be either due to the numerical accuracy in solving the dynamical equations as well as the possibility that the 1RSB threshold is not exactly the one that characterizes the long time dynamics. Further investigations are needed to clarify this point.
	Finally, according to \cite{cugliandolo1993analytical} the value of $\cal C$ at which the two straight line cross should coincide with the value of $q_M$ computed for the threshold states within the 1RSB solution.
	Again we find a very good agreement.

	\begin{figure}
		\centering
		\subfigure{
		\includegraphics[scale=.6]{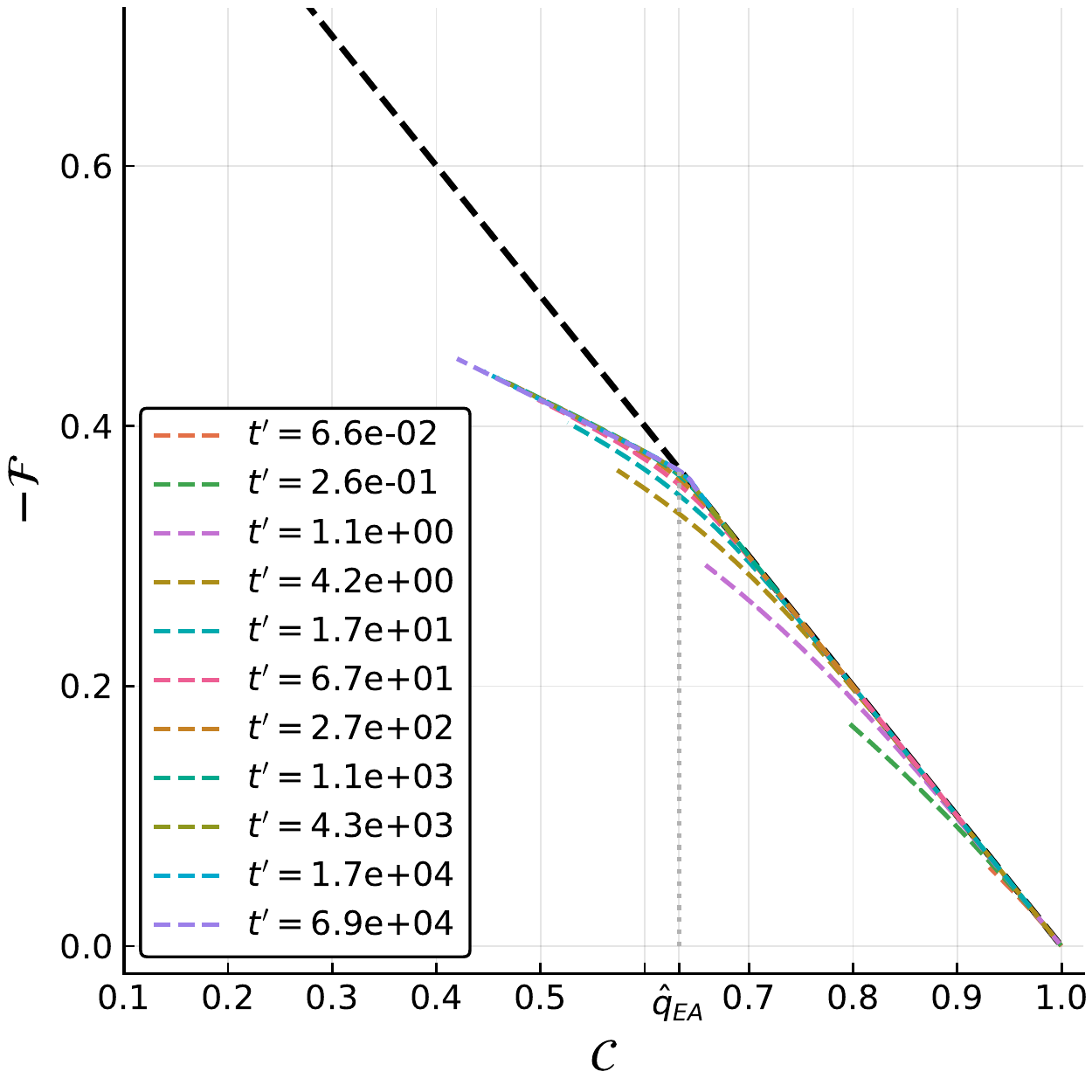}
		\includegraphics[scale=.6]{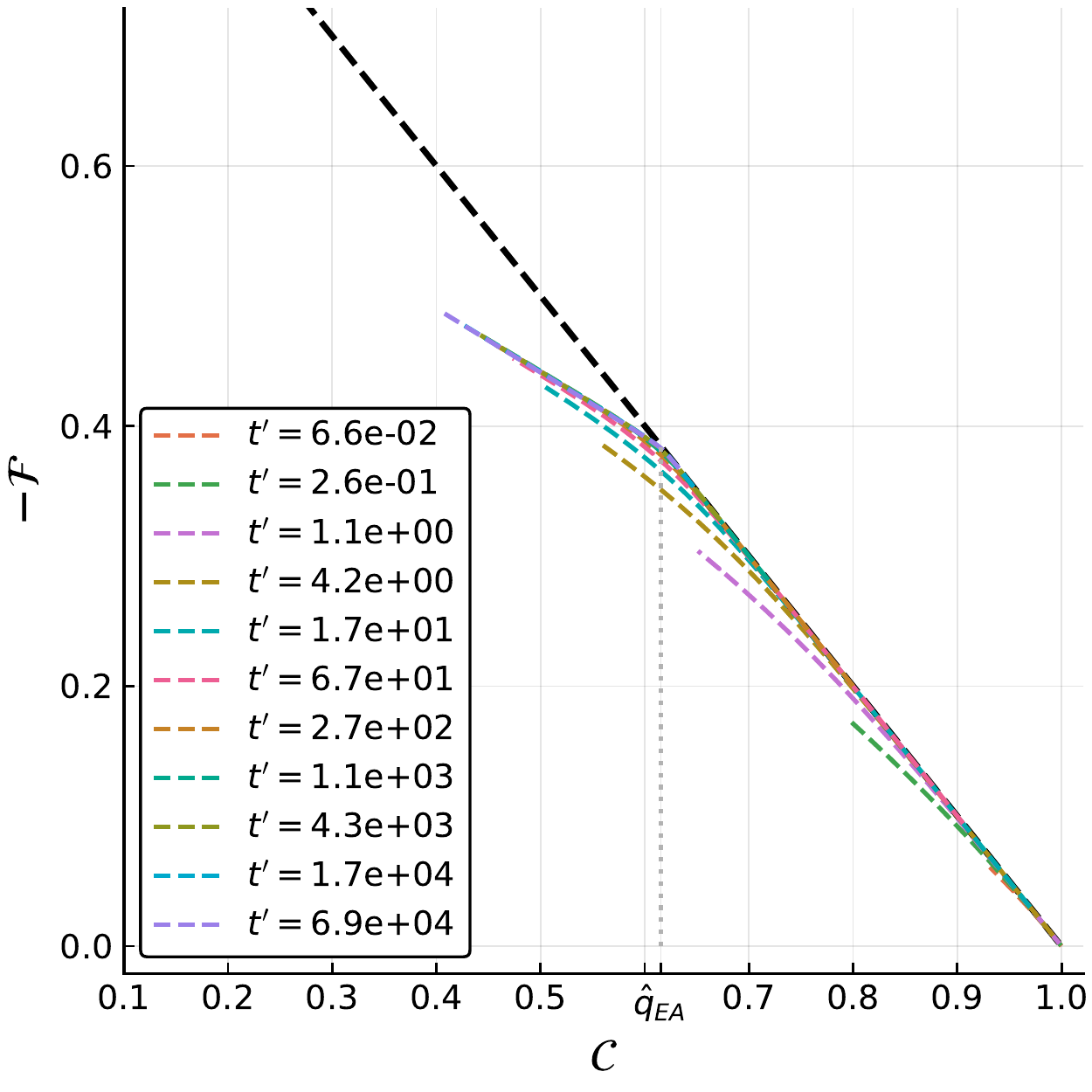}
		}
		\caption{Left panel: parametric plot of integrate response function with
			respect to correlation function for $p=3$, $\Delta_2 = 0.8$ and
			$\Delta_p = 0.2$. The different lines represent different
			waiting time, $t'$. The black dashed line correspond to the FDT prediction
			$x=1$. The vertical dotted line is the point where we
			observe a kink, which we denote by
			${\cal C}=\hat q_{EA}$ and should be equal to the saddle point value of $q_M$ as extracted from the 1RSB threshold states in the replica computation \cite{cugliandolo1993analytical}:
			$\hat q_{EA} = 0.633$ and $q_M = 0.638$. For $\mathcal{C}$
			smaller than $q_{EA}$ the FDT is violated and is replaced by a generalized version as in Eq.~(\ref{eq:generalized FDT}). We can
			obtain the value of the FDT ratio from a fit of the slope of the asymptotic curves for ${\cal C}<\hat q_{EA}$.
			We obtain $\hat x = 0.397$ which should be compared with the Parisi
			parameter that corresponds to 1RSB marginally stable states obtained from the replica computation that is $x = 0.408$.
			Right panel:  parametric plot of the integrated response as a function of the correlation
			for $p=3$ and $\Delta_2=1.4$ and $\Delta_p=0.2$. In this case the value of the FDT ratio
			extracted from fitting the data is $\hat x=0.397$ to be compared with the value of the Parisi parameter for the 1RSB threshold states that is $x=0.408$.
			At the same time data gives $\hat q_{EA}=0.633$ while the replica computation gives $q_M =0.638$.
		}
		\label{fig:generalized FDT fit}
	\end{figure}


\section{Free-energy Hessian, BBP transition and Langevin threshold}
\label{App:TAP}
In the following we present the derivation and the analysis of the Langevin threshold based on the study of the free energy Hessian.
The starting point of the analysis is the so-called TAP free energy,
i.e. the free energy as a function of the local magnetizations. The
TAP free-energy was introduced in the early days of spin-glass theory \cite{TAP,MPV87} and is now receiving a lot of attention in the mathematical community, see e.g. \cite{chen2018tap}. A straightforward generalization of 
the results of \cite{crisantisommers} allows one to obtain the TAP free energy for the model considered in this work, i.e. 
for the Hamiltonian (\ref{Hamiltonian}):
\begin{equation}
F(\{m_i\})=-\frac{\sqrt{(p-1)!}}{\Delta_pN^{(p-1)/2}}\sum_{i_1<\dots<i_p}T_{i_1\dots i_p}m_{i_1}\dots m_{i_p} \nonumber\\
-\frac{1}{\Delta_2\sqrt{N}}\sum_{i<j}Y_{ij}m_im_j+f(q) N \, . 
\end{equation}  
where we have set the temperature to one, $q=\sum_i m_i^2/N$ and $f(q)$ reads:
\[ f(q) = 
-\frac 1 2
\log(1-q)-\frac{1}{2p\Delta_p}\left[1+(p-1)q^{p}-pq^{p-1}\right]-\frac{1}{4\Delta_2}\left[1+q^{2}-2q\right]
\, .
\]
The so-called TAP states are local minima of $F(\{m_i\})$. We are interested in the free energy Hessian evaluated at the TAP states having zero overlap with the signal: 
\begin{equation}
\frac{\partial^2 F}{\partial m_i \partial m_j}=  G_{ij}+\delta_{ij}f'(q)-\frac{1}{\Delta_2}\frac{x_i^*x_j^*}{N}
+f''(q)\frac{m_im_j}{N} \, , \label{hessian_app}
\end{equation} 
where the matrix $G_{ij}$ is defined as 
\[
G_{ij}= -\frac{1}{\Delta_2\sqrt{N}}\xi_{ij} 
-\frac{\sqrt{(p-1)!}}{\Delta_p
  N^{(p-1)/2}(p-2)!}\sum_{i_1,\dots,i_{p-2}}\xi_{iji_1\dots i_{p-2}}
m_{i_1}\dots m_{i_{p-2}} \, .
\]
As shown originally in the spin-glass literature \cite{BM}, and recently put on a firmer basis by the Kac-Rice method \cite{auffinger2013random,ros2018complex}, the matrix $G_{ij}$ is statistically equivalent to a random matrix belonging to the Gaussian Orthogonal
Ensemble (GOE). In our case, the corresponding GOE matrix has elements which are i.i.d. Gaussian random variables with mean zero and variance $\sigma^2_F/N$, 
where 
\[
\sigma^2_F(q)=\frac{(p-1)q^{p-2}}{\Delta_p}+\frac{1}{\Delta_2}\,.
\] Neglecting for the moment the last two terms in eq. (\ref{hessian}),
one finds that the free-energy Hessian is the sum of a GOE matrix and the identity multiplied by $f'(q)$. The corresponding 
density of eigenvalues is therefore the Wigner semicircle with support $[-2\sigma_F(q)+f'(q),2\sigma_F(q)+f'(q)]$. 
This result is valid for any TAP state. The threshold states, which are the ones trapping the Langevin dynamics, are 
characterized by a vanishing fraction of zero modes, i.e. the left edge of the support of the Wigner semi-circle is zero. 
Their overlap is therefore fixed by the equation: 
\begin{equation}
2\sigma_F(q_{\rm th})=f'(q_{\rm th})\qquad  \rightarrow \qquad
\frac{1}{1-q_{\rm th}}=\sqrt{\frac{(p-1)q_{\rm th}^{p-2}}{\Delta_p}+\frac{1}{\Delta_2}}\,
. \label{qth}
\end{equation}
Let's consider now the role of the last two terms in eq. (\ref{hessian}). Both are rank-one perturbations and hence can lead 
to a BBP transition \cite{baik2005phase}, i.e. an eigenvalue that pops out of the Wigner semi-circle with an eigenvector having a finite overlap in the direction of the perturbation. It can be easily checked that $f''(q_{\rm th})\ge 0$; therefore the last term cannot lead to any negative eigenvalue and does not play any role in determining the stability of the threshold states.  It is the other 
term which is responsible for the instability in the direction of the signal. In fact, it is the contribution due to the spike; it becomes larger when the signal to noise ratio, $1/\Delta_2$, increases.  

The condition for the BBP transition for a GOE matrix having elements with variance $\sigma_F^2/N$ and which is perturbed 
by a rank one perturbation of strength $1/\Delta_2$ is $\frac{1}{\Delta_2}=\sigma_F$. This is the equation 
for the Langevin threshold:
\[
\frac{1}{\Delta_2}=\sqrt{\frac{(p-1)q_{\rm th}^{p-2}}{\Delta_p}+\frac{1}{\Delta_2}} \ ,
\]
Together with (\ref{qth}), this leads to the 
equation (\ref{eq:threshold}) presented in the main text and implies $q_{\rm th}=1-\Delta_2^*$ at the Langevin threshold.

\end{widetext}

\end{document}